\documentclass[10pt]{article} % For LaTeX2e
\usepackage[accepted]{tmlr}
\usepackage{amsmath,amsfonts,bm}

\def\eqref#1{equation~\ref{#1}}
\def\1{\bm{1}}

\DeclareMathAlphabet{\mathsfit}{\encodingdefault}{\sfdefault}{m}{sl}
\SetMathAlphabet{\mathsfit}{bold}{\encodingdefault}{\sfdefault}{bx}{n}

\DeclareMathOperator*{\argmax}{arg\,max}
\DeclareMathOperator*{\argmin}{arg\,min}

\usepackage{hyperref}
\usepackage{url}
\usepackage{dependencies}

\title{Freeze, Prompt, and Adapt: A Framework for Source-free Unsupervised GNN Prompting}

\author{\name Peyman Baghershahi \email pbaghe2@uic.edu \\
      \addr Department of Computer Science\\
      University of Illinois Chicago
      \AuthorAND \\
      \name Sourav Medya \email medya@uic.edu \\
      \addr Department of Computer Science\\
      University of Illinois Chicago}

\def\month{06}  % Insert correct month for camera-ready version
\def\year{2026} % Insert correct year for camera-ready version
\def\openreview{\url{https://openreview.net/forum?id=9KKgIQwCLO}} % Insert correct link to OpenReview for camera-ready version

\begin{document}

\maketitle

\begin{abstract}
Prompt tuning has become a key mechanism for adapting pre-trained Graph Neural Networks (\gnns) to new downstream tasks. However, existing approaches are predominantly supervised, relying on labeled data to optimize the prompting parameters and typically fine-tuning a task-specific prediction head---practices that undermine the promise of parameter-efficient adaptation.
We propose \textit{Unsupervised Graph Prompting Problem (\problemname)}, a challenging new setting where the pre-trained \gnn is kept entirely frozen, labels on the target domain are unavailable, the source data is inaccessible, and the target distribution exhibits covariate shift.
%To challenge this reliance, we introduce a rigorous problem setup: adapting a fully frozen pre-trained \gnn to a target dataset under a covariate shift, using only unlabeled data. 
To address this, we propose \namemodel, the first fully unsupervised \gnn prompting framework. \namemodel leverages consistency regularization and pseudo-labeling to train a prompting function, complemented with diversity and domain regularization to mitigate class imbalance and distribution mismatch.
%Our approach incorporates two regularization techniques to align the prompted graph distribution with the source data and to mitigate prediction bias. 
Our extensive experiments demonstrate that \namemodel consistently outperforms state-of-the-art supervised prompting methods with access to labeled data, demonstrating the viability of unsupervised prompting as a practical adaptation paradigm for \gnns. Code: \url{https://github.com/pbaghershahi/UGPrompt}.
\end{abstract}

\section{Introduction}

%%%% Points

Prompt tuning \citep{prefixtuning2021, powerofscale2021} has driven many recent developments for Large Language Models (LLMs). These methods optimize external prompting parameters to guide a model's responses. The goal is to avoid fine-tuning the model's vast number of internal parameters for adaptation to new downstream tasks, keeping the core pre-trained knowledge intact \citep{beyondfinetuning2025, parameterefficient2024, promptsurvey2023}. However, prompting becomes more challenging for graphs because: First, there is a wide range of graph tasks; thus, unlike general language tasks \citep{bert2019, gpt12018}, \textit{task unification} is difficult in graphs. This hinders the collection of large-scale data, e.g., from the internet \citep{gpt22019}, creating bottlenecks in pre-training \textit{large} Graph Neural Networks (\gnns) for general-purpose use. Second, because of the limited reasoning capabilities of \gnns \citep{gfm2024, position2024}, it is non-trivial to constitute instructive graph prompts in human (natural) language \citep{talkgraph2024}. Thus, \gnn prompting should be understood as a parameter-efficient adaptation paradigm inspired by LLM prompting, rather than as evidence that current \gnns possess LLM-like zero-/few-shot reasoning abilities.

A few recent studies have adopted prompting for \gnns to align the objectives of pre-training on source data and fine-tuning on target data, mostly following the ``pre-train, prompt, fine-tune'' pipeline \citep{gppt2022}. These works design unified tasks that allow optimizing a \gnn with semantically similar objectives on the pretext and downstream tasks \citep{prodigy2023, graphprompt2023, gpf2023, allinone2023, multigprompt2024, dagprompt2025}. However, the current \gnn prompting paradigm suffers from two key limitations that hinder its efficiency relative to LLMs. First, existing methods rely heavily on labeled data — costly to obtain — to achieve competitive performance. Second, they require training new projection heads for each downstream task, a form of \textit{lightweight fine-tuning} \citep{prefixtuning2021}. These dependencies on additional parameters and labeled data (especially in scenarios where the original source data is inaccessible, e.g., due to privacy) prevent \gnns from being used as truly frozen models. This gap motivates our work to establish a more practical and efficient prompting paradigm for \gnns.

% A few recent studies have adopted prompting for \gnns to align the objectives of pre-training on source data and fine-tuning on target data, mostly following the ``pre-train, prompt, fine-tune'' pipeline \citep{gppt2022}. These works design unified tasks that allow optimizing a \gnn with semantically similar objectives on the pretext and downstream tasks \citep{prodigy2023, graphprompt2023, gpf2023, allinone2023, multigprompt2024, dagprompt2025}. Following this line of work, we use the term \textit{prompting} to refer to adapting a pre-trained \gnn by optimizing a small prompting module while keeping the base model fixed. However, the current \gnn prompting paradigm suffers from two key limitations that hinder its efficiency compared to its LLM counterparts. First, the existing methods rely heavily on labeled data---which is costly to obtain---to achieve competitive performance. Second, they require training new projection heads for each downstream task, a form of \textit{lightweight fine-tuning} \citep{prefixtuning2021}. These dependencies on additional parameters and labeled data (especially in scenarios where the original source data is inaccessible, e.g., due to privacy) prevent \gnns from being used as truly frozen models. This gap motivates our work to establish a more practical and efficient prompting paradigm for \gnns.

To directly address these limitations, we first introduce the Unsupervised Graph Prompting Problem (\problemname), a rigorous novel problem formulation. The \problemname setup evaluates a method under four key conditions: the \gnn's parameters are frozen, there is a covariate shift in the target data distribution, no target labels are available for adaptation, and the source data is inaccessible. 
While this setup shares similarities with Unsupervised Source-Free Domain Adaptation (SFDA) \citep{surveysda2024}, it differs in a crucial aspect: \problemname requires the entire source-trained \gnn to be frozen, whereas SFDA methods \citep{soga2024, graphcta2024} rely on fine-tuning the model's parameters. This setting places our work within the paradigm of parameter-efficient prompting, rather than full model adaptation.
% While this setup is related to Unsupervised Source-Free Domain Adaptation (SFDA) \citep{surveysda2024}, \problemname corresponds to a more restrictive prompt-based case: the entire pre-trained GNN is frozen, whereas SFDA methods usually rely on fine-tuning the model's parameters \citepp{soga2024, graphcta2024}. This restriction places our work within parameter-efficient prompting rather than full model adaptation.

Within this challenging setup, we propose \namemodel, the first fully unsupervised \gnn prompting framework. \namemodel trains a prompting function using consistency regularization and confident pseudo-labeling, enabling the frozen \gnn to adapt its knowledge to the new target distribution. To ensure robustness, we introduce two additional regularization techniques: one to counteract prediction bias from class imbalance and another to make the prompted graphs close to the original data distribution. Our extensive experiments show that \namemodel, despite being fully unsupervised, consistently outperforms state-of-the-art (SOTA) prompting methods that have the advantage of full access to labeled data. Our major contributions are summarized as follows.

\begin{itemize}
    \item \textbf{Problem formulation. }We propose \problemname, a novel problem setup that isolates the true effectiveness of a prompting function by disallowing any updates to the source-trained \gnn's parameters. 
    \item  \textbf{Novel unsupervised methodology. }We propose \namemodel, the first fully unsupervised \gnn prompting method that leverages consistency regularization and pseudo-labeling to adapt a frozen \gnn to new data distributions.
    \item \textbf{Empirical analysis. }We demonstrate that \namemodel substantially outperforms supervised SOTA methods on node and graph classification tasks, validating the effectiveness of unsupervised adaptation in this novel and more practical setting. 
\end{itemize}

% Our major contributions are summarized as follows. 1) We propose a challenging setup for benchmarking \gnn prompting methods, which isolates the effectiveness of prompting functions by disallowing updates on the \gnn's parameters. 2) We propose a fully unsupervised \gnn prompting method, \namemodel, that uses consistency regularization and pseudo-labeling for training and better generalization while the \gnns' parameters are frozen. 3) \namemodel outperforms the SoTA \gnn prompting methods on node and graph classification across multiple datasets under different settings without supervision while the competitors rely on labels. 

\section{Background \& Problem Formulation}

Previous studies on adapting prompt tuning for \gnns use lightweight fine-tuning \citep{prefixtuning2021, powerofscale2021} with supervision, which has been addressed widely in different domains such as computer vision \citep{infonce2018, simclr2020, transfervision2021} and NLP \citep{bert2019, transfernlp2019}. Specifically, recent works have followed the \textit{``pre-train, prompt, fine-tune''} setup \citep{gppt2022, graphprompt2023, allinone2023, gpf2023}. In this section, we first introduce the setting and discuss its limitations, and then present our proposed problem setting, which addresses these limitations.

\subsection{Pre-train, Prompt, Fine-tuning}

This pipeline aims to bridge the generalization gap between pre-trained \gnns and downstream tasks that differ semantically from the pretext tasks. Unlike the traditional supervised and ``pre-training, fine-tuning'' methods, this pipeline employs a task unification step before pre-training and fine-tuning. This is essential because it aligns the pretext and downstream objectives to optimize a pre-trained model on new datasets. The steps are as follows.

\textit{(1) Pre-train. }Formally, given a set of tuples $\mathcal{S} = \{(\task_i, \data_i)\}^{N_s}_{i=1}$, with $N_s$ samples, from pretext task $\task_i$ and dataset $\data_i$, first all the tasks are unified to task $\task_u$ and the corresponding changes apply for their associated datasets to make a new set $\mathcal{S}_u = \{(\task_u^i, \data_u^i)\}^{N_s}_{i=1}$. Then a \gnn encoder $g(.;\theta_g)$ is pre-trained on $\mathcal{S}_u$ using an unsupervised approach such as contrastive learning \citep{graphcl2020, simgrace20222}. For downstream tasks, by adding a projection head $h(.; \theta_h)$ after the encoder, a model $\psi = h \circ g$ is formed, of which the encoder parameters $\theta_g$ are frozen, and only the head parameters $\theta_h$ will be trained. 
\textit{(2) Prompting. }At this stage, a prompting function $f(., \theta_f)$ is employed to construct a prediction model $\varphi$. The prompting function is either a prefix module, i.e., $\varphi = h \circ g \circ f$, or a postfix one, i.e.,  $\varphi = f \circ h \circ g$. \textit{(3) Fine-tuning. } In the final step, the set of parameters $\{\theta_h, \theta_f\}$ of $\varphi$ are optimized for every unified downstream task $\task_u$ with the new labeled samples.

\textit{Limitations. }A deeper look at the \textit{``pre-train, prompt, fine-tune''} pipeline reveals that although the current methods impose less trainable parameters compared to full fine-tuning (fine-tuning both pre-trained feature encoder and decoder), they involve partial (lightweight) fine-tuning \citep{parameterefficient2024, prefixtuning2021, powerofscale2021}, as they also train the \gnn's decoder (projection head) along with the new parameters of the prompting function. Subsequently, labels become essential for this partial fine-tuning, and they are unable to leverage unlabeled data from large datasets when collecting labeled data is challenging \citep{gpt22019}. Also, fine-tuning a large pre-trained model on large datasets may introduce noisy information when the labeled downstream datasets are small \citep{statisticalearning2004, understandml2014}. Therefore, it reduces model generalization across diverse applications \citep{gpt32020}.

\subsection{Our Problem Setting}
To address the above limitations, we first need a suitable problem setting that offers insights into how well a prompting method performs when there is lack of labeled data. It is also crucial to evaluate if the method generalizes across tasks without fine-tuning the base \gnn model parameters.

\paragraph{Unsupervised Graph Prompting Problem (UGPP).} \textit{Suppose a \gnn model $\varphi(.; \theta_g, \theta_h) = h(.; \theta_h) \circ g(.; \theta_g)$ is given, where $g$ and $h$ are its encoder and decoder. Also, $\varphi$ is trained for task $\task$ on a labeled source dataset $\data_s = \{(x_s^i, y_s^i) : x_s^i \sim \prob^s_X,\ y_s^i \sim \prob^s_{Y|X}\}_{i=1}^{Ns}$, where $x_s^i$ is a sample (e.g., a graph or node), and $y_s^i$ is its associated label. The problem is to train a prompting module $f(.; \theta_f)$ on an unlabeled target dataset $\data_t = \{x_t^j: x_t^j \sim \prob^t_X\}_{j=1}^{N_t}$, s.t. $\prob^s_X \neq \prob^t_X$, to enhance the performance of $\varphi$ for task $\task$ on $\data_t$, assuming that $\theta_g$ and $\theta_h$ are fixed, $\data_s$ is unobservable, and $\prob^t_{Y|X}$ remains invariant across domains, i.e., $\prob^t_{Y|X} = \prob^s_{Y|X}$}.

This problem, \textsc{Ugpp}, focuses on prompting without labeled data. A prompting method that performs well in this setting has three advantages: \textit{First,} it is model- and task-agnostic, i.e., it works for conventional \gnns (e.g., GCN, GAT) and tasks (e.g., node/graph classification). \textit{Second,} it is unsupervised, allowing the use of many datasets to improve generalization. \textit{Third,} this setting does not depend on the source data. This is particularly beneficial when the source data is inaccessible, e.g., due to privacy issues. Throughout the paper, we use ``source-trained'' to refer to the model trained on the labeled source domain for task $\task$, and reserve ``pre-trained'' for the general prompting literature.

% It is noteworthy that our setting fundamentally differs from conventional Unsupervised SFDA settings. While SFDA typically involves fine-tuning the parameters of the source-trained model, our setting mandates that the pre-trained GNN remains entirely frozen. All adaptation is achieved exclusively by optimizing the external prompt parameters. Please find additional discussions on the \problemname definition in Appendix \ref{apdx:ugpp}. Nevertheless, we compare our proposed method (described next) against recent graph SFDA methods in Appendix \ref{apdx:sfda_comparison}.

It is noteworthy that \problemname can be viewed as a restrictive prompt-based instance of Unsupervised SFDA. Conventional SFDA methods typically fine-tune the source-trained model, whereas \problemname keeps the source-trained \gnn, including both the encoder and the prediction head, entirely frozen. Therefore, all adaptation is isolated in the external prompting function. This restriction allows us to evaluate the effect of prompting without entangling it with the base model's latent fine-tuning, and it also makes the setting model-agnostic. Please find additional discussions on the \problemname definition in Appendix \ref{apdx:ugpp} We also compare our proposed method against recent graph SFDA methods in Appendix \ref{apdx:sfda_comparison}
\begin{figure*}[t]
 \vspace{-4mm}
  \centering
    \begin{tikzpicture}
      \node[anchor=south west,inner sep=0] (image) at (0,0) {\includegraphics[trim=0 105 0 90, clip, scale=0.6]{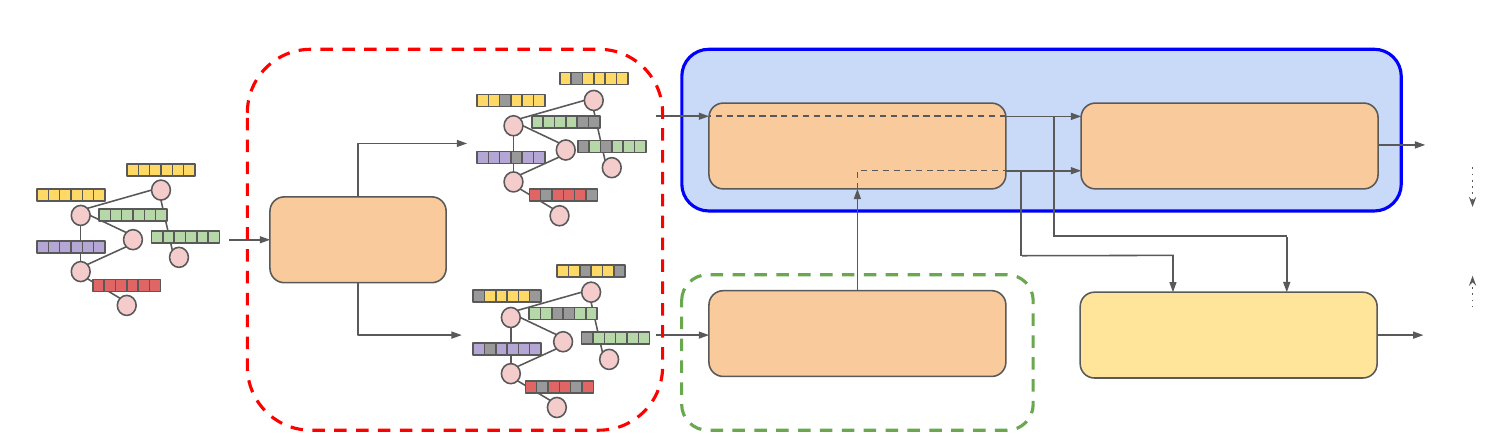}};
      \begin{scope}[x={(image.south east)},y={(image.north west)}]
        \node[anchor=center] at (0.09,0.22) {\scriptsize $\graph$};
        \node[anchor=center] at (0.09,0.7) {\scriptsize \textcolor{darkBlue}{Unlabeled Data}};
        \node[anchor=center] at (0.24,0.455) {\scriptsize Augmentation};
        \node[anchor=center] at (0.19,0.8) {\scriptsize A)};
        \node[anchor=center] at (0.47,0.82) {\scriptsize C)};
        \node[anchor=center] at (0.47,0.085) {\scriptsize B)};
        \node[anchor=center] at (0.69,0.835) {\scriptsize \textcolor{darkBlue}{GNN (Frozen)}};
        \node[anchor=center] at (0.69,0.76) {\scriptsize $z^{\graph}_a$};
        \node[anchor=center] at (0.685,0.64) {\scriptsize $z^{\graph}_p$};
        \node[anchor=center] at (1.005,0.67) {\scriptsize $\loss_c + \lambda_1 \loss_{div}$};
        \node[anchor=center] at (0.99,0.24) {\scriptsize $\lambda_2 \loss_{adv}$};
        \node[anchor=center] at (0.98,0.49) {\scriptsize \textcolor{darkBlue}{Unsupervised}};
        \node[anchor=center] at (0.98,0.42) {\scriptsize \textcolor{darkBlue}{Training}};
        \node[anchor=center] at (0.575,0.66) {\scriptsize Encoder ($g$)};
        \node[anchor=center] at (0.82,0.66) {\scriptsize Projection Head ($h$)};
        \node[anchor=center] at (0.82,0.24) {\scriptsize Discriminator};
        \node[anchor=center] at (0.575,0.265) {\scriptsize Prompting Function ($f$)};
        \node[anchor=center] at (0.575,0.2) {\scriptsize (Trainable)};
        % \node[anchor=center] at (0.375,0.9) {\scriptsize $\graph_w$};
        % \node[anchor=center] at (0.375,0.00) {\scriptsize $\graph_a^u$};
        \node[anchor=center] at (0.295,0.75) {\scriptsize $\graph_w$};
        \node[anchor=center] at (0.295,0.15) {\scriptsize $\graph_s$};
        \node[anchor=center] at (0.555,0.43) {\scriptsize $\graph_p$};
        % \node[anchor=center] at (0.75,0.43) {\large //};
      \end{scope}
    \end{tikzpicture}
     \vspace{-1mm}
  \caption{Overview of \namemodel. A) A non-parametric algorithm generates a weak augmentation $\graph_w$ and a strong augmentation $\graph_s$ from an unlabeled graph $\graph$. B) The learnable prompting function $f$ generates a prompted graph $\graph_p$ from $\graph_s$. C) The base \gnn with frozen parameters scores $\graph_p$ and $\graph_w$. A discriminator taking input from the latent representation ($z_a^{\graph}$ for $\graph_w$ and $z_p^{\graph}$ for $\graph_p$) of the \gnn's encoder regularizes the model to adapt to the input distribution.}
  \label{fig:framework}
    \vspace{-1mm}
\end{figure*}
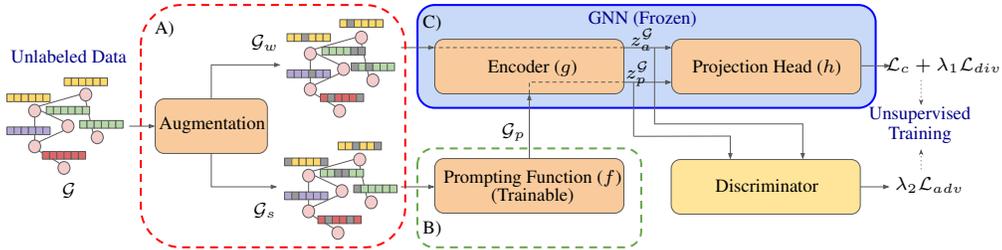

\section{Our Method: \namemodel}
%\todo{need a NICE pipeline figure}\\
Here, we introduce \namemodel, an unsupervised \gnn prompting method, to address \problemname.\\
%\sm{self-note: checked till here}
\textbf{Motivation.} We aim to design an unsupervised prompting framework that helps a source-trained \gnn make robust predictions while its parameters are frozen. To achieve this, we take advantage of pseudo-labeling \citep{pseudolabel2013, selftraining2020} and consistency regularization \citep{stochasticregu2016, temporalensemble2017}. To train a prompting function in a fully unsupervised manner, \namemodel first obtains randomly augmented graphs from the target dataset. Then we employ consistency regularization \citep{flexmatch2021, freematch2023}, where certain predictions of the base \gnn are filtered by a confidence threshold and exploited as pseudo-labels for optimization. Indeed, knowledge from the source dataset is used to reduce prediction entropy and handle distribution shifts by leveraging \gnn's confident predictions for unlabeled target data. The prompting module enhances target input samples by adding key information, making them more similar to the source samples. Empirical evidence to support this claim is in the Appendix \ref{apdx:shift_visualization}. We describe our framework (Figure \ref{fig:framework}) for graph-level tasks. Note that it also generalizes to node- and edge-level tasks.

\textbf{Overview of \namemodel.}Our framework to address \problemname has a training and an inference step. The training step involves two components: \textit{consistency-based prompting} and \textit{prompt regularization}. \textit{Consistency-based prompting} starts by augmenting each input graph twice, with one of the augmentations modified by a prompting function. Next, the \gnn scores both samples. The objective is to train the prompting function such that the \gnn produces ``consistent prediction scores'' for both samples of this pair with a certain confidence. In \textit{prompt regularization} we introduce two regularization techniques: one to tackle biased predictions caused by a class imbalance in the data and another to prevent generating out-of-distribution (OOD) prompted graphs. During inference, a test graph is fed to the prompting function without augmentation, and its output goes to the \gnn. We discuss a task unification step that generalizes to other graph tasks, e.g., node classification, in Appendix \ref{apdx:exp_details}.

\subsection{The Consistency-based Prompting}\label{sec:consistency_prompting}
Our prompting method is designed to reduce the discrepancy of the base \gnn predictions over random augmentations of the same input graphs. We achieve this without labels by first augmenting the data with an algorithmic step, then generating pseudo-labels from the unlabeled augmentations when their assigned scores by the \gnn meet a certain confidence threshold. We provide the details of these two stages below.

\textbf{\textit{Algorithmic augmentation.}} 
Consistency regularization techniques \citep{flexmatch2021} train \textit{strong augmentations} of samples using the pseudo-labels derived from their \textit{weak augmentations}. Since our focus is on optimizing the prompting module rather than fine-tuning the base \gnn, we adopt this technique as follows. We use a random non-parameterized augmentation algorithm (we use random feature masking in our experiments). More specifically, we mask a group of features with a certain probability. We augment an input graph $\graph$ to create a weak augmentation $\graph_w$ with masking probability $p_w$ and also a strong augmentation $\graph_s$ with probability $p_s$, where $p_s > p_w$. We pass $\graph_s$ through a learnable prompting function $f(.; \theta_f)$ to obtain a \textit{prompted graph} $\graph_p=f(\graph_s; \theta_f)$. We keep $\graph_w$ unchanged and call it a \textit{non-prompted} augmentation graph.

\textbf{\textit{Learnable prompting. }}
We use a prefix prompting module to transform input samples for the base \gnn. Our approach is generic enough to allow the integration of different prompting functions. For our experiments, we choose a function $f$ that enriches node feature vectors, as this design aligns with our random feature masking augmentation technique. This way of adding learnable parameters is used in GPF-Plus \citep{gpf2023}. We discuss the choice of input-level prompt tuning and its relation to other parameter-efficient adaptation methods such as LoRA \citep{lora2022}, in Appendix~\ref{apdx:prompting_function}.

Specifically, we learn a prompting function $f$ with parameter set $\theta_f=\{t^*_j: t^*_j \in \realset^d\}_{j=1}^{n_t}$. For input graph $\graph$ of $N$ nodes with nodes features set $X = \{x_i: x_i \in \realset^d\}_{i=1}^N$, the function $f$ makes prompted graph $\graph_p$ with feature set $X_p = \{x_i+t_i: x_i \in X, \ x_i, t_i\in \realset^d\}_{i=1}^{N}$ such that $t_i = \sum_{j=1}^{n_t}\alpha_{i,j}t^*_j$ and $\alpha_{i,j}=\frac{\exp(x_i^Tt^*_j)}{\sum_{l=1}^{n_t}\exp(x_i^Tt^*_l)}$ .

\textbf{\textit{Consistency-based Objective.}}
We optimize $\theta_f$ to minimize the discrepancy between the \gnn's prediction scores for the non-prompted augmented graph $\graph_w$ and the prompted graph $\graph_p$. This would lower the entropy of the \gnn's scores for the target unlabeled data. Intuitively, a well-trained frozen \gnn model $\varphi$ makes accurate predictions for samples close to the source distribution. Therefore, as training proceeds and $\varphi$ scores different random perturbations of the same samples, we utilize its confident predictions as pseudo-labels for optimization, as this helps $f$ to capture the distribution shift and make the  predictions robust.
We achieve this by passing $\graph_p$ and $\graph_w$ to $\varphi$ for prediction as:
\vspace{-4mm}

\begin{align}\label{eq:predscores}
    \tilde{\mathbf{p}}^{\graph}_{\varphi} &= \delta(h(z^{\graph}_a; \theta_h)) \hspace{1in} \hat{\mathbf{p}}^{\graph}_{\varphi} = \delta(h(z^{\graph}_p; \theta_h))
\end{align}

where $z^{\graph}_a = g(\graph_w; \theta_g)$ and $z^{\graph}_p = g(\graph_p; \theta_g)$, $\delta(.)$ denotes the softmax function, $\tilde{\mathbf{p}}^{\graph}_{\varphi}, \hat{\mathbf{p}}^{\graph}_{\varphi} \in \realset^C$, and $C$ is the number of classes. Pseudo-labels are made as $\tilde{y}^{\graph}_{\varphi} =\argmax \tilde{\mathbf{p}}^{\graph}_{\varphi}$, and finally, the consistency loss is:
\vspace{-1mm}
\begin{align}\label{eq:consistencyloss}
    \loss_c = \frac{1}{|\mathcal{B}|}\sum_{\graph \in \mathcal{B}}\mathbb{1}(\max(\tilde{\mathbf{p}}^{\graph}_{\varphi}) > \tau)CE(\tilde{y}^{\graph}_{\varphi}, \hat{\mathbf{p}}^{\graph}_{\varphi})
\end{align}

where CE(.,.) is the cross-entropy function, $\mathcal{B}=\{\graph_i\}_{i=1}^{|\mathcal{B}|}$ is a sample batch, and $\tau$ is a confidence threshold. $\tau$ excludes low-certainty predictions (samples less aligned with the source data distribution) and can be fixed or class-dynamic (see Appendix \ref{apdx:exp_details}). 

Unlike standard semi-supervised adaptation, \namemodel's use of consistency regularization is novel in that we apply it in a stricter frozen-\gnn prompting setting, allowing only learning an external prompt that serves as a strong augmentation to align target samples with the source-trained model.

\subsection{Prompt Regularization}\label{sec:prompt_regularization}

\textbf{Diversity.} Due to the class imbalance, only reducing the consistency loss may cause biased predictions and trivial solutions such that every sample is assigned to the same class to reduce the overall entropy. To mitigate this, inspired by \citep{doweuda2020}, we regularize the model to maximize the entropy of the scores' expected value over a batch, and encourage diverse predictions. 
\vspace{-1mm}
\begin{align}\label{eq:diversityloss}
    \loss_{div} = -H(\hat{\mathbf{q}}) = \mathbf{1}^\top(\hat{\mathbf{q}}\odot\log\hat{\mathbf{q}}); \ \hat{\mathbf{q}} = \frac{1}{|\mathcal{B}|}\sum_{\graph \in \mathcal{B}} \hat{\mathbf{p}}^{\graph}_{\varphi}
\end{align}

Here $H(.)$ is the entropy function, and $\odot$ is the Hadamard product. Employing consistency regularization with an adjusted confidence threshold ($\tau$) and integrating a diversity loss ($\mathcal{L}_{div}$) to prevent class collapse helps our framework address pseudo-label dependency and potential calibration risks.

% \textbf{Domain Adaptation.}
% While the prompting function $f$ minimizes the discrepancy of predictions for the same sample, it may also create OOD prompted graphs. However, even without access to the source data samples, the knowledge learned from these samples is preserved in the frozen parameters $\theta_{\varphi}$. 

\textbf{Domain Adaptation. }
While the prompting function $f$ minimizes the discrepancy of predictions for the same sample, it may also create OOD prompted graphs. Since the source data are unavailable, we cannot directly align prompted representations with the source distribution. Instead, we can exploit the knowledge learned from source samples that is preserved in the frozen parameters $\theta_{\varphi}$ to mitigate the OOD issue. To achieve this, we train an adversarial discriminator $d(.; \theta_d)$---e.g., a simple feedforward network with trainable parameters $\theta_d$---to distinguish a prompted graph $\graph_p$ from a non-prompted augmented graph $\graph_w$. Formally, we optimize the discriminator as follows:

\vspace{-1mm}
\begin{align}\label{eq:discloss}
    \theta_d^\star = \underset{\theta_d}{\argmin} &- \frac{1}{2|\mathcal{B}|}\sum_{\graph \in \mathcal{B}} [\log \sigma(d(z_a^{\graph}; \theta_d)) + \log (1-\sigma(d(z_p^{\graph}, \theta_d)))]
\end{align}

where $\sigma(.)$ is the sigmoid function. We normalize the sum by $\frac{1}{2|\mathcal{B}|}$ since every graph has two samples.
% Note that, $\loss_d$ is only used to optimize the discriminator and does not appear in the final objective.
Nevertheless, we regularize $f$ with the following objective to make $g$'s representations for prompted graphs closer to the non-prompted augmented graphs.

\vspace{-2mm}

\begin{align}\label{eq:domainloss}
    \loss_{adv} = - \frac{1}{|\mathcal{B}|}\sum_{\graph \in \mathcal{B}} \log \sigma(d(z_p^{\graph}; \theta_d^\star))
\end{align}

The adversarial discriminator regularizes the prompted representations to stay close to their non-prompted weak augmentations, preventing the prompt from moving them to arbitrary regions of the latent space. Thus, $L_{\mathrm{adv}}$ should be interpreted as a target-side stability regularizer, while source compatibility is mainly induced by the frozen model and the confidence-filtered consistency objective. We provide an MMD-based diagnostic supporting this interpretation in Appendix~\ref{app:mmd_confident_samples}.

\subsection{Final Objective \& Complexity Analysis}\label{sec:objective_complexity_analysis}
Our unsupervised objective approach involves three parts. Eq. \ref{eq:consistencyloss} encourages consistency across \gnn predictions to handle distribution shift and exploit the \gnn's learned knowledge from source data. Eq. \ref{eq:diversityloss} and Eq. \ref{eq:domainloss} handle class imbalance and avoid generating OOD prompted graphs, respectively. The final objective to optimize $\theta_f$ becomes:

\vspace{-4mm}

\begin{align}\label{eq:totalloss}
    \theta_f^\star = \underset{\theta_f}{\argmin} \ \loss; \qquad \loss = \loss_c + \lambda_1 \loss_{div} + \lambda_2 \loss_{adv}
\end{align}

where $\lambda_1$, $\lambda_2$ are hyper-parameters. 
% The impact of each objective is evaluated in the ablation studies.
We bring empirical evidence supporting the logic behind adding each of the regularization objectives terms in Appendix \ref{apdx:regeffect}.

%Since all objective functions are on the model predictions and consisted of probabilities, they are in the same scale. We evaluate the impact of these hyper-parameters on the model performance in the ablations studies.

During inference, the augmentation step is skipped, allowing the prompting module to produce a prompted graph directly from an input graph to align it with the trained model's knowledge. This prompted graph is passed to the trained \gnn for prediction. The pseudocode of \namemodel for both the training and inference is presented in Algorithms \ref{alg:ugprompttraining} and \ref{alg:ugpromptinference} of the Appendix.

%\subsection{Complexity Analysis}
The time complexity of a regular \gnn (e.g. GCN), is $O(NLd^2 + L|E|d)$, where $N$,$E$, $L$, and $d$ are the number of nodes, edges, \gnn layers, and the dimensionality of node embeddings respectively. A common graph augmentation algorithm, like feature masking, requires $O(Nd)$ operations. The complexity of the prompting method used in our experiments is $Ndn_t$, where $n_t$ is the number of trainable prompting vectors. Thus, the overall complexity is $O(NLd^2 + L|E|d + Ndn_t)$. 
% We compare our training and test times against the baselines in Appendix \ref{apdx:computation_time} and show that \namemodel is, on average, more efficient at test time. 

We compare wall-clock training and inference time against the baselines in Appendix~\ref{apdx:prompting_function} (Table~\ref{tab:time_comparison}). \namemodel has slightly higher training cost due to consistency learning and the discriminator, but this overhead is only used during training. At inference time, the augmentation pipeline and discriminator are removed, and the model only applies the lightweight prompting function before the frozen \gnn. Therefore, \namemodel remains close to the BaseModel inference cost while adding only a small number of trainable prompt parameters and a lightweight discriminator during training.

%$L$ is the number of \gnn layers, and $d$ is the dimensionality of node embeddings. A common graph augmentation algorithm like feature masking requires $O(Nd)$ operations. The complexity of the prompting method used in our experiments (GPF-Plus \citep{gpf2023}) is $NBd_f$, s.t. $B=n_p$ is the number of trainable prompting vectors. Thus, the overall complexity is $O(NLd^2 + L|E|d + NBd)$.
%\clearpage
\section{Experiments}

% \vspace{-1mm}
% \subsection{Experimental Setup}
% \vspace{-1mm}

\textbf{Datasets and Code. }We experiment on six standard datasets for graph and node classification. We use ENZYMES \citep{enzymesdb2004}, PROTEINS \citep{proteinfunction2005}, DHFR \citep{dhfr2003}, BBBP and BACE \citep{moleculenet2017} datasets for graph classification, which have continuous or discrete features. For node classification, we use Cora, CiteSeer, PubMed \citep{planetoid2016}, Flickr \citep{graphsaint2020}, Cornell, Texas, Wisconsin \citep{geomgcn2020}. Please find more details in Appendix \ref{apdx:datasets}. 
% The code is available at \url{https://anonymous.4open.science/r/UGPrompt-6C3E}.

\noindent
\textit{Distribution Shift in Datasets.}
Our problem definition (\problemname) requires evaluating a source-trained \gnn on target datasets that exhibit a covariate shift from the source data. To implement this, our main experiments induce shifts based on fundamental graph properties. For graph classification, we generate datasets with varying edge homophily ratios \citep{beyondhomophily2020}, a property known to intrinsically affect \gnn information aggregation \citep{mpnn2017}. For node classification, we use PageRank \citep{evalrobust2023} to create a popularity-based shift, which provides a challenging evaluation scenario. Further details on the generation of these distributions are in Appendix \ref{apdx:dist_shift}, and additional experiments on other shifts (e.g., graph density, clustering coefficient) are in Appendix \ref{apdx:exp_dist_shift}.

\begin{table*}[t]
    \scriptsize
    %\vspace{-2mm}
    \caption{Graph classification results on target datasets (for GCN and GAT base models) show our unsupervised \namemodel largely outperforming competitors that use 25\% labeled data.}
    \centering
    \vspace{-1mm}
    \resizebox{0.9\textwidth}{!}{
        \begin{tabular}{L{0.2in}L{0.65in}C{0.2in}*{10}{C{0.25in}}}
            \toprule
            \multirow{2}{6em}{\textbf{Base\\GNN}} & \multirow{2}{6em}{\textbf{Method}} & \multirow{2}{*}{\hspace{-2mm}\textbf{\%Label}} &  \multicolumn{2}{c}{\textbf{ENZYMES}}  &  \multicolumn{2}{c}{\textbf{PROTEINS}}  &  \multicolumn{2}{c}{\textbf{DHFR}} & \multicolumn{2}{c}{\textbf{BBBP}} & \multicolumn{2}{c}{\textbf{BACE}} \\
            \cmidrule(lr){4-5} \cmidrule(lr){6-7} \cmidrule(lr){8-9} \cmidrule(lr){10-11} \cmidrule(lr){12-13}
            & & & \textbf{F1} & \textbf{IMP} & \textbf{F1} & \textbf{IMP} & \textbf{F1} & \textbf{IMP} & \textbf{F1} & \textbf{IMP} & \textbf{F1} & \textbf{IMP} \\
            \midrule
            \multirow{7}{*}{GCN} & BaseModel & 0 & 47.7\std{5.7} & - & 51.8\std{7.0} & - & 75.5\std{6.3} & - & 87.3\std{1.2} & - & 58.5\std{3.3} & - \\
            \cmidrule(lr){2-13}
            & Fine-Tuning & \multirow{5}{*}{25} & 46.4\std{0.1} & -2.7 & 47.5\std{0.7} & -8.3 & 76.8\std{0.1} & 1.7 & \textbf{88.3}\std{1.2} & \textbf{1.1} & 64.0\std{0.7} & 1.1 \\
            & GraphPrompt & & 38.1\std{1.4} & -20.1 & 50.8\std{1.4} & -1.9 & 71.7\std{1.1} & -5.0 & 81.9\std{1.2} & -6.2 & \underline{64.3}\std{0.7} & \underline{9.9} \\
            & GraphPrompt+ & & 23.5\std{2.5} & -50.7 & 44.7\std{5.1} & -13.7 & 64.9\std{6.2} & -14.0 & 82.0\std{1.2} & -6.1 & 61.0\std{3.1} & 8.4 \\
            & All-In-One & & 45.8\std{1.9} & -4.0 & 38.1\std{13.4} & -26.4 & \textbf{79.1}\std{0.6} & \textbf{4.6} & 85.7\std{1.2} & -1.8 & 52.5\std{6.1} & -10.3 \\
            & GPF-Plus & & \underline{48.3}\std{1.8} & \underline{1.3} & \underline{53.8}\std{2.6} & \underline{3.9} & 76.9\std{0.3} & 1.9 & \underline{88.0}\std{1.2} & \underline{0.8} & 63.4\std{1.2} & 8.4 \\
            \cmidrule(lr){2-13}
            & \namemodel & \textbf{0} & \textbf{49.1}\std{0.6} & \textbf{2.9} & \textbf{56.0}\std{1.5} & \textbf{8.1} & \underline{77.0}\std{2.4} & \underline{2.0} & \textbf{88.3}\std{0.2} & \textbf{1.1} & \textbf{64.6}\std{1.9} & \textbf{10.4} \\
            \midrule
            \multirow{7}{*}{GAT} & BaseModel & 0 & \underline{44.1}\std{6.4} & - & 51.5\std{8.1} & - & \underline{77.3}\std{3.5} & - & 87.7\std{1.2} & - & 46.4\std{1.2} & - \\
            \cmidrule(lr){2-13}
            & Fine-Tuning & \multirow{5}{*}{25} & 42.9\std{0.0} & -2.7 & 50.2\std{0.8} & -2.5 & 76.5\std{0.0} & -1.0 & \textbf{88.4}\std{0.1} & \textbf{0.8} & 46.4\std{0.7} & -2.7 \\
            & GraphPrompt & & 29.3\std{1.0} & -33.6 & 49.5\std{1.1} & -3.9 & 72.8\std{0.8} & -5.8 & 83.4\std{0.3} & -4.9 & 57.9\std{0.5} & 0.0 \\
            & GraphPrompt+ & & 25.7\std{4.6} & -41.7 & 48.7\std{7.7} & -5.4 & 57.2\std{9.1} & -26.0 & 82.4\std{2.0} & -6.0 & 64.2\std{3.7} & 10.9 \\
            & All-In-One & & 39.5\std{2.5} & -10.4 & 31.5\std{14.8} & -38.8 & 76.0\std{2.0} & -1.7 & 86.5\std{0.4} & -1.4 & 57.9\std{3.9} & -9.8 \\
            & GPF-Plus & & 42.9\std{2.0} & -2.7 & \underline{54.9}\std{1.9} & \underline{3.7} & \underline{77.3}\std{0.9} & 0.0 & \underline{88.3}\std{0.3} & \underline{0.7} & \underline{65.8}\std{0.9} & \underline{13.6} \\
            \cmidrule(lr){2-13}
            & \namemodel & \textbf{0} & \textbf{45.9}\std{2.2} & \textbf{4.1} & \textbf{56.4}\std{2.0} & \textbf{9.5} & \textbf{78.2}\std{0.9} & \textbf{1.2} & 87.8\std{0.2} & 0.1 & \textbf{66.1}\std{0.2} & \textbf{14.2} \\
            \bottomrule
        \end{tabular}}
        \vspace{-4mm}
    \label{tab:graphgcn25}
\end{table*}

\textbf{Evaluation Setting.} All datasets are split in half to create source and target sets with shifted distributions. We train a base \gnn on the source set and evaluate it on the target set. Since baseline prompting methods rely on supervised training, we allow only the baselines to access labeled data for training in four setups of 25\%, 50\%, 75\%, and 100\% (full supervision). Please find experiments on 50\%, 75\% and 100\% labeled data in Appendix \ref{apdx:fully_supervised}. We report the F1-score and the relative improvement in F1-score (IMP\%) compared to the BaseModel baseline.

\textbf{Baselines.} We consider several types of baselines.\\
\textit{(1) BaseModel.} The base \gnn without prompting and fine-tuning, which is expected to be outperformed by prompting methods. We use GCN \citep{gcn2017} and GAT \citep{gat2018} as the base \gnn. More experiments with recent advanced \gnns are in Appendix \ref{apdx:advanced_gnn}. \\
\textit{(2) Fine-Tuning. }The base \gnn model when we fix its encoder and just fine-tune its projection head. The goal is to verify the claim that fine-tuning on new dataset with labels does not necessarily improve performance \citep{gppt2022, gpf2023, allinone2023}.\\
\textit{(3) \gnn Prompting Methods. }Our work is the first attempt for graph prompting without labels and updating the base \gnn's parameters. Thus, we compare with all the SOTA \gnn prompting methods used in the recent benchmark \citep{prog2024}, namely All-In-One \citep{allinone2023} and GPF-Plus \citep{gpf2023}, GraphPrompt \citep{graphprompt2023}, GraphPrompt+ \citep{graphpromptplus2024}, and GPPT \citep{gppt2022}. Also, we have compared our method against additional baselines, using GCN as the BaseModel's architecture, in Tables \ref{tab:few_shot_comparison_graph} and \ref{tab:few_shot_comparison_node} of the Appendix. We use the codebases from the corresponding papers.

We do not allow these methods to update the \gnn's parameters and only their prompting modules are supposed to be learned on the target dataset. This restriction follows the \problemname setting, where the goal is to isolate the effect of the prompting function while keeping the entire source-trained model frozen. We note that this is more restrictive than the original setting of the supervised prompting baselines, which often fine-tune a task-specific prediction head. Therefore, in Appendix~\ref{app:finetuned_decoder_baselines}, we additionally evaluate these baselines under a relaxed setting where their decoders are fine-tuned with labeled target data.

\begin{table*}[t]
%\vspace{-2mm}
    \caption{Node classification results on target datasets for GCN as the base model. Compared to baselines given 25\% of labeled data, \namemodel generally achieves better results without labels.}
    \centering
    \resizebox{\textwidth}{!}{
        \begin{tabular}{L{0.28in}L{0.8in}C{0.3in}*{14}{C{0.32in}}}
            \toprule
            \multirow{2}{6em}{\textbf{Base\\GNN}} & \multirow{2}{6em}{\textbf{Method}} & \multirow{2}{*}{\hspace{-2mm}\textbf{\%Label}} &  \multicolumn{2}{c}{\textbf{Cora}}  &  \multicolumn{2}{c}{\textbf{CiteSeer}}  &  \multicolumn{2}{c}{\textbf{PubMed}} & \multicolumn{2}{c}{\textbf{Flickr}} & \multicolumn{2}{c}{\textbf{Cornell}} & \multicolumn{2}{c}{\textbf{Texas}} & \multicolumn{2}{c}{\textbf{Wisconsin}} \\
            \cmidrule(lr){4-5} \cmidrule(lr){6-7} \cmidrule(lr){8-9} \cmidrule(lr){10-11} \cmidrule(lr){12-13} \cmidrule(lr){14-15} \cmidrule(lr){16-17}
            & & & \textbf{F1} & \textbf{IMP} & \textbf{F1} & \textbf{IMP} & \textbf{F1} & \textbf{IMP} & \textbf{F1} & \textbf{IMP} & \textbf{F1} & \textbf{IMP} & \textbf{F1} & \textbf{IMP} & \textbf{F1} & \textbf{IMP} \\
            \midrule
            \multirow{8}{*}{GCN} & BaseModel & 0 & 53.8\std{2.4} & - & 44.1\std{1.5} & - & 57.1\std{0.8} & - & \underline{16.5}\std{0.4} & - & 19.1\std{6.2} & - & 23.6\std{10.3} & - & 25.2\std{10.2} & - \\
            \cmidrule(lr){2-17}
            & Fine-Tuning & \multirow{6}{*}{25} & 51.7\std{0.5} & -3.9 & 40.0\std{0.3} & -9.3 & 54.3\std{3.4} & -4.9 & 10.4\std{0.1} & -37.0 & 19.1\std{0.0} & 0.0 & \underline{26.4}\std{0.0} & \underline{11.9} & \underline{27.1}\std{0.1} & \underline{7.5} \\
            & GPPT & & 47.8\std{3.5} & -11.2 & 38.4\std{0.5} & -12.9 & 51.6\std{4.8} & -9.6 & 13.5\std{0.5} & -18.2 & 15.1\std{3.0} & -20.9 & 25.6\std{8.6} & 8.5 & 23.4\std{0.1} & -7.1 \\
            & GraphPrompt & & 53.8\std{0.4} & 0.0 & 41.6\std{0.3} & -5.7 & 56.9\std{0.1} & -0.4 & 13.0\std{0.1} & -21.2 & 10.9\std{1.4} & -42.9 & 4.8\std{0.4} & -79.7 & 10.5\std{0.1} & -58.3 \\
            & GraphPrompt+ & & 49.8\std{0.3} & -7.4 & 39.9\std{0.3} & -9.5 & \textbf{62.0}\std{0.4} & \textbf{8.6} & 14.8\std{0.6} & -10.3 & 11.5\std{2.1} & -39.8 & 4.8\std{0.3} & -79.9 & 12.0\std{0.1} & -52.3 \\
            & All-In-One & & 50.5\std{1.1} & -6.1 & 38.3\std{1.0} & -13.2 & 42.1\std{0.9} & -26.3 & 13.8\std{0.3} & -16.4 & 13.0\std{0.8} & -31.9  & 21.7\std{1.6} & -8.1 & 21.4\std{0.1} & -15.1 \\
            & GPF-Plus & & \underline{56.5}\std{0.6} & \underline{5.0} & \underline{45.6}\std{0.6} & \underline{3.4} & 59.1\std{0.4} & 3.5 & 13.3\std{0.3} & -19.4 & \underline{22.0}\std{0.5} & \underline{15.2} & 25.2\std{1.4} & 6.8 & 26.7\std{0.1} & 6.0 \\
            \cmidrule(lr){2-17}
            & \namemodel & \textbf{0} & \textbf{57.3}\std{0.4} & \textbf{6.5} & \textbf{45.7}\std{0.4} & \textbf{3.6} & \underline{61.2}\std{0.3} & \underline{7.2} & \textbf{17.5}\std{0.4} & \textbf{6.1} & \textbf{23.2}\std{0.1} & \textbf{21.5} & \textbf{26.8}\std{0.8} & \textbf{13.6} & \textbf{28.0}\std{0.1} & \textbf{11.1} \\
            \midrule
            \multirow{8}{*}{GAT} & BaseModel & 0 & \underline{47.7}\std{1.3} & - & 41.2\std{2.4} & - & 60.0\std{1.1} & - & 17.0\std{0.2} & - & \underline{18.6}\std{0.2} & - & 28.1\std{0.2} & - & 19.9\std{6.9} & - \\
            \cmidrule(lr){2-17}
            & Fine-Tuning & \multirow{6}{*}{25} & 43.5\std{0.6} & -8.8 & 38.8\std{0.3} & -5.8 & 55.6\std{2.7} & -7.3 & 10.9\std{0.2} & -35.9 & 18.2\std{0.0} & -2.2 & 21.2\std{0.0} & -24.6 & \underline{21.8}\std{0.0} & \underline{9.5} \\
            & GPPT & & 31.5\std{3.9} & -34.0 & 34.3\std{1.8} & -16.7 & 51.7\std{4.6} & -13.8 & 12.9\std{0.1} & -24.1 & 17.2\std{4.5} & -7.5 & 28.2\std{5.5} & 0.4 & 21.5\std{4.1} & 8.0 \\
            & GraphPrompt & & 44.2\std{0.6} & -7.3 & 39.2\std{0.4} & -4.9 & 60.1\std{0.1} & 0.2 & 13.4\std{0.3} & -21.1 & 14.3\std{1.3} & -23.1 & 1.4\std{0.0} & -95.0 & 15.4\std{1.5} & -22.6 \\
            & GraphPrompt+ & & 41.2\std{0.9} & -13.6 & 37.8\std{0.7} & -8.3 & \textbf{64.0}\std{1.1} & \textbf{6.7} & \underline{17.5}\std{0.6} & \underline{2.9} & 13.5\std{2.0} & -27.4 & 1.4\std{0.0} & -95.0 & 17.1\std{2.4} & -14.1 \\
            & All-In-One & & 34.3\std{2.1} & -28.1 & 27.6\std{1.4} & -33.0 & 22.7\std{3.2} & -62.2 & 13.3\std{0.2} & -21.8 & 13.5\std{0.2} & -27.4 & 21.2\std{0.7} & -24.6 & 16.9\std{0.9} & -15.1 \\
            & GPF-Plus & & 47.6\std{1.5} & -0.2 & \underline{42.1}\std{0.6} & \underline{2.2} & 60.1\std{0.3} & 0.2 & 13.8\std{0.2} & -18.8 & 17.9\std{1.1} & -3.8 & \textbf{30.4}\std{0.7} & \textbf{8.2} & 21.7\std{1.2} & 9.0 \\
            \cmidrule(lr){2-17}
            & \namemodel & \textbf{0} & \textbf{48.8}\std{0.9} & \textbf{2.3} & \textbf{42.3}\std{0.5} & \textbf{2.7} & \underline{60.2}\std{0.1} & \underline{0.3} & \textbf{17.6}\std{0.3} & \textbf{3.5} & \textbf{21.8}\std{1.5} & \textbf{17.2} & \underline{29.5}\std{1.0} & \underline{5.0} & \textbf{22.2}\std{1.1} & \textbf{11.6} \\
            \bottomrule
        \end{tabular}}
    \label{tab:nodegcn25}
\end{table*}

\subsection{Results on Graph Classification}
For graph classification, 50\% of graphs are randomly sampled as the source dataset, with higher-homophily graphs having a greater chance of selection; the remaining 50\% form the target dataset. Please see experiments with graph density distribution shift in Table \ref{tab:graphgcndensity} (see the Appendix). The base \gnn is trained on the source dataset. We repeat the experiments with two base \gnns (GCN and GAT) to demonstrate how the models generalize across different architectures. Note that GPPT is limited to node classification, so we exclude it from this experiment.  

Table \ref{tab:graphgcn25} presents graph classification results, where baselines use 25\% labeled data, while our method, \namemodel, uses 0\%. Two key observations highlight \namemodel's contribution: first, it consistently surpasses the BaseModel, validating it as a reliable, non-detrimental prompting method. Second, and \textit{most notably, \namemodel's use of no labels offers broad applicability to diverse unlabeled datasets, marking a step towards graph foundation models.} Interestingly, evaluations with \problemname setting reveal that baselines often fail to improve performance and sometimes make it worse. Most of the graph prompting methods, except for GPF-Plus, perform poorly with both \gnn architectures. Although GPF-Plus has the same prompting function as \namemodel's, it struggles with adapting to distribution shifts. Conversely, \namemodel leverages source data knowledge and generates pseudo-labels from highly confident predictions, learning effectively from samples that closely match the source distribution. This ensures consistent improvement across all cases.
\iffalse
\begin{figure}[ht]
    \centering
    \includegraphics[
    trim=0 0 0 0, clip, 
    width=.6\columnwidth, 
    ]{content/figures/losses.pdf}
    \caption{The effect of regularization objectives on \namemodel with GCN as the base model.} 
    \label{fig:losses}
    \vspace{-2mm}
\end{figure}
\fi
\subsection{Results on Node Classification}

First, we compute PageRank (PR) for all nodes of each dataset. We sample 50\% of nodes from the source dataset based on the normalized PR, so that graphs with higher PR are more likely to be included. The rest are assigned to the target dataset. A 2-hop neighborhood of each node is extracted as a subgraph for task unification to graph classification, and it inherits the main node's label. More experiments with the distribution shift of the type clustering coefficient are provided in Table \ref{tab:nodegcncc} (see the Appendix). 

Node classification results are in Table \ref{tab:nodegcn25}. \namemodel outperforms on all datasets except PubMed (where it is second-best) and, \textit{notably, uses no labeled data, unlike the baselines.} Additionally, the \gnn's performance degrades on target data across all datasets after fine-tuning its projection head (the Fine-Tuning baseline) with 25\% of labels. This verifies that fine-tuning a model on a small-sized labeled dataset may introduce noisy information when the downstream data distribution does not align with the model's learned knowledge \citep{statisticalearning2004, understandml2014, gpt32020}. We also verify that \namemodel maintains high performance on the source domain and does not cause forgetting of learned knowledge, unlike the other baselines; please find the experiments in Appendix \ref{apdx:no_shift}.

An important advantage of an unsupervised method---such as ours---is that it allows the use of large-scale unlabeled datasets. It is important to emphasize that \namemodel achieves the best results on Flickr, the largest dataset, and second-best with large margins over other baselines on PubMed, the second-largest dataset, indicating that \namemodel can perform well on large datasets.

\begin{table*}[!t]
\vspace{-2mm}
    \small
    \caption{\namemodel performance under different strong augmentation rates $p_s$ for the prompted augmented graph. Higher values of $p_s$ provide greater variation in masked feature groups and generally help better capture distribution shifts.}
    \vspace{-2mm}
    \centering
       \scalebox{0.85}{ \begin{tabular}{L{0.2in}*{12}{C{0.3in}}}
            \toprule
            \multirow{2}{6em}{\textbf{$p_s$}} & \multicolumn{2}{c}{\textbf{ENZYMES}} &   \multicolumn{2}{c}{\textbf{PROTEINS}} & \multicolumn{2}{c}{\textbf{DHFR}} & \multicolumn{2}{c}{\textbf{Cora}} & \multicolumn{2}{c}{\textbf{CiteSeer}} & \multicolumn{2}{c}{\textbf{PubMed}} \\
            \cmidrule(lr){2-3} \cmidrule(lr){4-5} \cmidrule(lr){6-7} \cmidrule(lr){8-9} \cmidrule(lr){10-11} \cmidrule(lr){12-13}
            & \textbf{F1} & \textbf{IMP} & \textbf{F1} & \textbf{IMP} & \textbf{F1} & \textbf{IMP} & \textbf{F1} & \textbf{IMP} & \textbf{F1} & \textbf{IMP} & \textbf{F1} & \textbf{IMP} \\
            \midrule
            0.0 & 47.7 & - & 55.7 & - & \underline{76.8} & - & 56.3 & - & 44.9 & - & 57.2 & - \\
            \midrule
            0.1 & \textbf{49.1} & \textbf{2.9} & 55.7 & 0.0 & \textbf{77.0} & \textbf{0.3} & 56.7 & 0.7 & 44.9 & 0.0 & 58.6 & 2.4 \\
            0.2 & \underline{48.9} & \underline{2.5} & \underline{55.9} & \underline{0.4} & \underline{76.8} & \underline{0.0} & 56.8 & 0.9 & 45.0 & 0.2 & 59.7 & 4.4 \\
            0.3 & 48.1 & 0.8 & \textbf{56.0} & \textbf{0.5} & 76.2 & -0.8 & \underline{57.1} & \underline{1.4} & \underline{45.2} & \underline{0.7} & \underline{60.4} & \underline{5.6} \\
            0.4 & 46.9 & -1.7 & 55.6 & -0.2 & 75.7 & -1.4 & \textbf{57.3} & \textbf{1.8} & \textbf{45.7} & \textbf{1.8} & \textbf{61.2} & \textbf{7.0} \\
            \bottomrule
        \end{tabular}}
    \label{tab:augratio}
\end{table*}

% Why can an unsupervised method outperform supervised baselines? 
Although supervised learning usually provides more informative gradients, this is not necessarily an upper bound in the \namemodel setting. Since the source-trained \gnn is frozen and the target data are shifted, adaptation mainly requires aligning target samples with the source-trained model. \namemodel uses confidence-thresholded pseudo-labeling and therefore optimizes the prompt only from target samples on which the source-trained model is already certain, reducing the impact of highly shifted samples. Moreover, \namemodel allows the use of the full unlabeled target set, whereas supervised baselines are limited to training on the labeled subset. The few-shot results in Section~\ref{sec:fewshot} further show that labels improve \namemodel when they are used together with our prompt-based alignment.

Additionally, we evaluate the supervised prompting baselines in their more relaxed setting with fine-tuned decoders and 25\% labeled target data in Appendix~\ref{app:finetuned_decoder_baselines}; \namemodel remains more stable under distribution shift despite using no labels and keeping the full base model frozen.

\subsection{Ablation Study}
\subsubsection{The Effect of Regularization}

\iffalse
\begin{figure}[h!]
  \centering
  \includegraphics[
    trim=30 0 0 0, 
    clip, 
    width=0.8\textwidth
  ]{content/figures/regularization_comparison.pdf}
  \caption{The effect of regularization objectives on \namemodel with GCN as the base model.}
  \label{fig:losses}
\end{figure}
\fi

\begin{wrapfigure}{r}{0.6\textwidth}
   \vspace{-8mm}
    \begin{center}
 \centering
    \includegraphics[width=.6\textwidth]{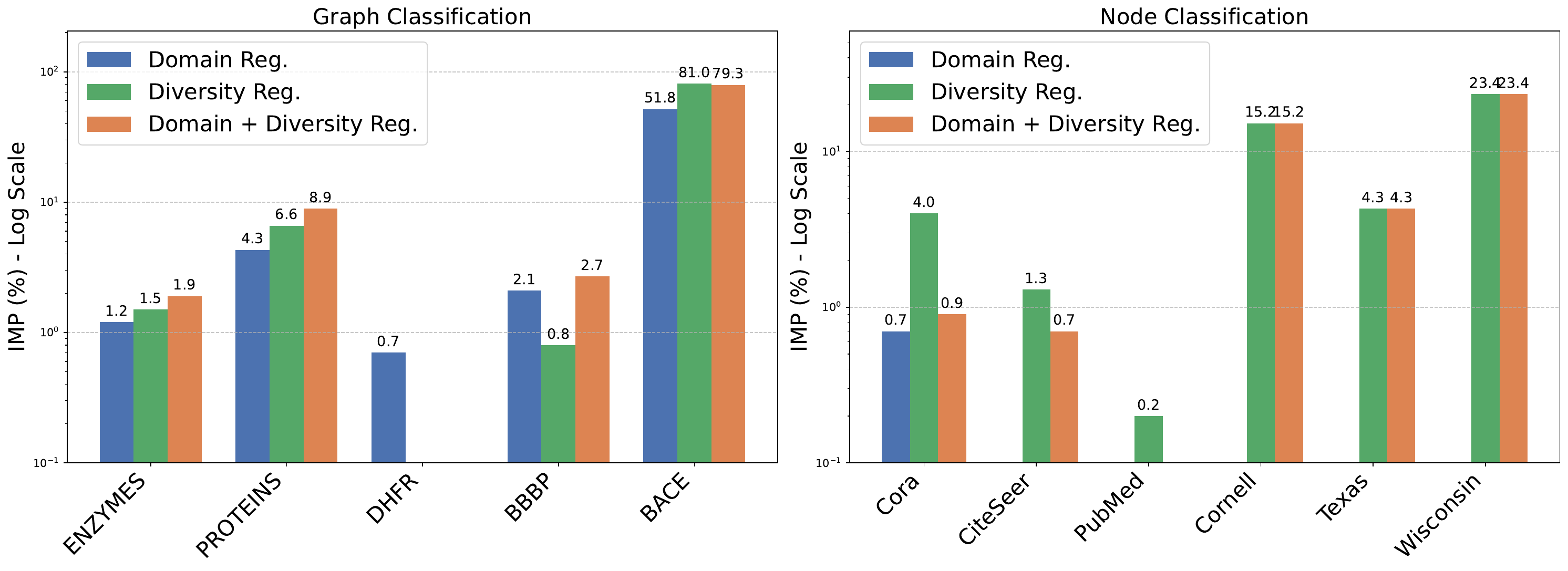}
    \end{center}
  \caption{The effect of regularization objectives on \namemodel with GCN as the base model.}\label{fig:losses}
   \vspace{-2mm}
 \end{wrapfigure}
To evaluate the regularization effect, \namemodel is trained in four scenarios: (1) without regularization (``w/o''), (2) with only domain adaptation regularization (``domain''), (3) with only diversity regularization (``diversity''), and (4) with both regularizations (``domain + diversity''). Settings (2), (3), and (4) are compared to (1), with Figure \ref{fig:losses} showing IMP improvements. 
% If a regularization term fails to enhance the model, IMP is set to zero, meaning it can be neutralized (setting $\lambda_1$ or $\lambda_2$ to zero) in Equation \ref{eq:totalloss}.
During evaluation, if validation performance indicates that a specific regularization term fails to improve the model, its corresponding weight ($\lambda_1$ or $\lambda_2$ in Equation \ref{eq:totalloss}) is set to zero, and the test IMP for that term resolves to zero. Results show that both regularization factors have positive effects across most datasets; however, their combination is not always superior to their individual applications.
 
Domain adaptation regularization is beneficial across all graph classification datasets. We conjecture that the distributions of node classification datasets are more likely to be densely populated, whereas graph datasets often exhibit scattered hollow spaces in the latent space between classes, increasing the likelihood of generating OOD-prompted graphs. Therefore, domain adaptation regularization would be more beneficial. Moreover, a key finding of this experiment is the importance of adding diversity regularization. More specifically, when we have severe class imbalance, for example, in PROTEINS, Cora, and CiteSeer, ``diversity'' empowers the base \gnn significantly, which supports our claims. Appendix \ref{apdx:regeffect} provides more discussion on the effect of these objectives.

\subsubsection{The Effect of Augmentation} 

We mask node features in weakly augmented graphs ($\graph_w$) for pseudo-labeling with probability $p_w$ and in strongly augmented graphs $\graph_s$ for prompting with probability $p_s$. Since $\graph_s$ resembles a distribution shift from $\graph_w$, a higher $p_s$ enables learning from distribution shifts across more feature groups, potentially offering greater robustness. Here, we fix $p_w = 0.1$ and evaluate the impact of varying $p_s$. The IMP for all $p_s$ values is compared to no augmentation $p_s = 0$. The results in Table \ref{tab:augratio} support our intuition that augmenting, confidence pseudo-labeling, and consistency training are advantageous. Higher $p_s$ yields better performance over PROTEINS, Cora, CireSeer, and PubMed because augmentation by feature masking---replacing a group of feature values with 0.0---is semantically more aligned with the discrete and binary features of these datasets; however, for ENZYMES and DHFR with continuous features, this augmentation is not as effective and higher values of $p_s$ may be detrimental and cause learning from noise. Additional experiments on evaluating the effect of augmentation type (e.g. modifying graph structure) are in Appendix \ref{apdx:augmentation_type}. Besides, we show the versatility and generalization of our framework in Appendix \ref{apdx:prompting_function} using All-In-One's prompting function as a choice of function $f$ (see Section \ref{sec:consistency_prompting}). Also, additional diagnostics on pseudo-label accuracy, calibration, and sensitivity to confidence threshold ($\tau$) are provided in Appendix~\ref{app:pseudolabel_reliability}.

\subsubsection{The Effect of Homophily Shift}

To evaluate how performance scales with the increase in covariate shift (or difficulty), we extend our experiment by constructing explicit degrees of shift. Instead of a random 50--50 split (as in the original setting), we sort all samples by homophily (for the graph classification dataset) and PageRank (for node classification datasets). The top 50\% (highest homophily) are used as the source dataset. The remaining 50\% are divided into four target sets with increasing difficulty (amount of covariate shift with respect to source set): (i) Set 1: 50--67.5 (closest to source), (ii) Set 2: 67.5--75, (iii) Set 3: 75--87.5, and (iv) Set 4: 87.5--100 (lowest homophily). Table~\ref{tab:homophily_merged} shows the results. \namemodel maintains strong gains on the easier shifts (Sets 1 \& 2) and continues to deliver improvements even under severe distribution shift (Sets 3 \& 4). In contrast, GPF-Plus is worse and frequently induces negative transfer (negative IMP\%).

\begin{table*}[t]
\centering
\vspace{-2mm}
\small
\caption{Scaling with homophily shift across four target sets. \namemodel is better in most cases.}
\label{tab:homophily_merged}

\scalebox{0.8}{\begin{tabular}{llcccccc}
\toprule
\textbf{Target set} & \textbf{Method} & \textbf{ENZYMES} & \textbf{PROTEINS} & \textbf{DHFR} & \textbf{BACE} & \textbf{Cora} & \textbf{CiteSeer} \\
\midrule

\multirow{2}{*}{\textbf{Set 1}} 
 & GPF-Plus     & -3.6 & 11.5 & -3.2 & 20.7  & -1.8 & 0.2 \\
 & \textbf{UGPrompt} & \textbf{0.9} & \textbf{19.6} & \textbf{-0.6} & \textbf{40.1} & \textbf{0.7} & \textbf{2.0} \\
\midrule

\multirow{2}{*}{\textbf{Set 2}} 
 & GPF-Plus     & -5.4 & 57.6 & -1.9 & 106.1 & 2.4 & 0.2 \\
 & \textbf{UGPrompt} & \textbf{1.0} & \textbf{62.0} & \textbf{2.0} & \textbf{155.6} & \textbf{2.6} & \textbf{0.7} \\
\midrule

\multirow{2}{*}{\textbf{Set 3}} 
 & GPF-Plus     & \textbf{27.1} & 85.7 & \textbf{10.2} & 108.8 & \textbf{2.2} & \textbf{2.1} \\
 & \textbf{UGPrompt} & 3.0 & \textbf{89.2} & 7.3 & \textbf{124.7} & 0.2 & 0.5 \\
\midrule

\multirow{2}{*}{\textbf{Set 4}} 
 & GPF-Plus     & -7.8 & 44.4 & \textbf{3.7} & 97.3 & -2.6 & 1.3 \\
 & \textbf{UGPrompt} & \textbf{1.9} & \textbf{46.4} & 2.7 & \textbf{113.7} & \textbf{3.5} & \textbf{1.9} \\
\bottomrule
% \vspace{-5mm}
\end{tabular}}
\end{table*}

\begin{table*}[!t]
% \vspace{-2mm}
    \small
    \caption{Evaluation of \namemodel’s performance in a few-shot setting using GCN as the base \gnn under homophily and PR distribution shifts. Results indicate considerable gains with labeled data, particularly in node classification tasks.}%, while maintaining competitive performance in fully unsupervised settings.}
    %\vspace{-2mm}
    \centering
       \scalebox{0.8}{ \begin{tabular}{L{0.7in}C{0.4in}*{12}{C{0.3in}}}
            \toprule
            \multirow{2}{6em}{\textbf{Method}} & \multirow{2}{*}{\textbf{\%Label}} & \multicolumn{2}{c}{\textbf{ENZYMES}} &   \multicolumn{2}{c}{\textbf{PROTEINS}} & \multicolumn{2}{c}{\textbf{DHFR}} & \multicolumn{2}{c}{\textbf{Cora}} & \multicolumn{2}{c}{\textbf{CiteSeer}} & \multicolumn{2}{c}{\textbf{PubMed}} \\
            \cmidrule(lr){3-4} \cmidrule(lr){5-6} \cmidrule(lr){7-8} \cmidrule(lr){9-10} \cmidrule(lr){11-12} \cmidrule(lr){13-14}
            & & \textbf{F1} & \textbf{IMP} & \textbf{F1} & \textbf{IMP} & \textbf{F1} & \textbf{IMP} & \textbf{F1} & \textbf{IMP} & \textbf{F1} & \textbf{IMP} & \textbf{F1} & \textbf{IMP} \\
            \midrule
            BaseModel & & 47.7 & - & 51.8 & - & 75.5 & - & 53.8 & - & 44.1 & - & 57.1 & - \\
            \midrule
            \multirow{4}{6em}{\namemodel (ours)} & 25 & \textbf{49.1} & \textbf{2.9} & \textbf{56.2} & \textbf{8.5} & \textbf{77.2} & \textbf{2.3} & \textbf{58.2} & \textbf{8.2} & \textbf{47.0} & \textbf{6.6} & \textbf{63.5} & \textbf{11.2} \\
            & 10 & \underline{49.0} & \underline{2.7} & \underline{56.1} & \underline{8.3} & 77.0 & 2.0 & \underline{57.6} & \underline{7.1} & \underline{46.2} & \underline{4.8} & \underline{62.1} & \underline{8.8} \\
            & 5 & \textbf{49.1} & \textbf{2.9} & \underline{56.1} & \underline{8.3} & \underline{77.1} & \underline{2.1} & 57.5 & 6.9 & 46.0 & 4.3 & \underline{62.1} & \underline{8.8} \\
            & 0 & \textbf{49.1} & \textbf{2.9} & 56.0 & 8.1 & 77.0 & 2.0 & 57.3 & 6.5 & 45.7 & 3.6 & 61.2 & 7.2 \\
            \bottomrule
        \end{tabular}}
        % \vspace{-2mm}
    \label{tab:few_shot}
\end{table*}

\subsection{Few-shot Learning for \namemodel} \label{sec:fewshot}

\namemodel is an unsupervised \gnn prompting method. However, it can potentially utilize labeled data efficiently when it is available. Assuming every batch $\mathcal{B}$ is composed of a set of labeled samples $S_l$ and unlabeled samples $S_u$ s.t. $\mathcal{B}=S_l\cup S_u$, we can replace $\loss_c$ in Equation \ref{eq:consistencyloss} by $\loss_c = \frac{1}{|\mathcal{B}|}(\loss_l + \lambda_3\loss_u)$ in which $\loss_l = \sum_{\graph \in S_l} CE(y^{\graph}, \hat{\mathbf{p}}^{\graph}_{\varphi})$ is the supervised objective term and $\loss_u = \sum_{\graph \in S_u}\mathbb{1}(\max(\tilde{\mathbf{p}}^{\graph}_{\varphi}) \geq \tau)CE(\tilde{y}^{\graph}_{\varphi}, \hat{\mathbf{p}}^{\graph}_{\varphi})$ is the unsupervised term. 

Here we evaluate \namemodel in a few-shot setting. Table \ref{tab:few_shot} presents the results using GCN as the base \gnn under homophily and PR distribution shifts. \namemodel significantly performs better when labels are provided. For node classification datasets, IMP in the 25\% labels setting is notably higher than in the unsupervised case (0\% labels), while improvements on PROTEINS and DHFR are smaller. Meanwhile, performance on ENZYMES remains comparable in the absence of labels, likely due to a significant covariate shift, which makes learning from highly heterophilic data challenging even with labels. The key takeaway emerges by comparing Table \ref{tab:few_shot} with Tables \ref{tab:graphgcn25} and \ref{tab:nodegcn25}, showing that with 25\% of labeled data, \namemodel outperforms all baselines on most datasets except DHFR. Notably, its superiority is also evident in the 0\% label setting.

\subsection{Exploratory Robustness Beyond the UGPP Assumption}

\begin{table}
    %\vspace{-3mm}
    \caption{
    Category-disjoint stress test between source and target datasets, while the projection head remains frozen. Upper panel: BaseModel F1-score on source and target domains. Lower panel: prompting methods' improvements (IMP\%) on the target domain.
    }
    \label{tab:exp1_label_shift}
    \centering{
    \small
        \scalebox{0.8}{\begin{tabular}{L{0.6in}*{3}{C{0.6in}}}
            \toprule
            \textbf{Method} & \textbf{Cornell} & \textbf{Texas} & \textbf{Wisconsin} \\
            \midrule
            \multicolumn{4}{c}{BaseModel Performance (F1-score)} \\
            \midrule
            Source & 46.7 & 79.8 & 60.1 \\
            Target & 15.3 & 13.1 & 13.9 \\
            \midrule
            \multicolumn{4}{c}{Target Domain Improvement (IMP\%)} \\
            \midrule
            GPF-Plus & 30.1 & -1.5 & 5.4 \\
            \textbf{UGPrompt} & \textbf{36.6} & \textbf{9.9} & \textbf{28.1} \\
            \bottomrule
        \end{tabular}
        }}
\end{table}

% To test robustness beyond simulated covariate shift, we design two experiments. We use the Cornell, Texas, and Wisconsin datasets, which share the same feature dimensionality and number of classes. In the first experiment, we evaluate a setting where the source and target datasets have different class distributions. For each dataset, the base GNN is pre-trained on the classes $\{0,1,2\}$ — each dataset has 5 classes. Then the base is frozen, prompted, and evaluated on classes $\{2,3,4\}$ (reindexed to $\{0,1,2\}$).

The formal \problemname setting assumes covariate shift with a shared conditional label distribution. Here, we provide an exploratory stress test beyond this assumption. The goal is to evaluate whether prompting can still improve a frozen model when the target categories are semantically disjoint from those used to train the head. We use the Cornell, Texas, and Wisconsin datasets, which share the same feature dimensionality and number of classes. For each dataset, the base \gnn is trained only on source samples from classes $\{0,1,2\}$, and then the base model is frozen. We then prompt and evaluate it on samples from classes $\{2,3,4\}$. Since the projection head remains frozen with output dimension \(C=3\), the target labels are reindexed to $\{0,1,2\}$ only for computing evaluation metrics, and no classifier weights are updated.

% Table~\ref{tab:exp1_label_shift} summarizes the findings. The BaseModel collapses under this strong semantic/label shift, whereas \namemodel recovers substantial performance across all datasets, outperforming GPF-Plus (the baseline), which occasionally leads to negative transfer.

Table~\ref{tab:exp1_label_shift} summarizes the findings. The BaseModel collapses under this category-disjoint stress test, showing that the original semantic mapping of the frozen head is no longer reliable. Nevertheless, \namemodel recovers substantial performance across all datasets, outperforming GPF-Plus, which occasionally leads to negative transfer.

Our second experiment is designed to evaluate robustness to real-world covariate shift. We consider a more extreme scenario than the simulated covariate shift discussed before. This setting is standard (particularly in transfer learning) and more challenging. Here, we train the base \gnn on a source dataset, such as Cornell, and then prompt and evaluate it on another dataset, such as Texas, which is denoted as Cornell $\rightarrow$ Texas. Results are shown in Table~\ref{tab:exp2_label_shift}. \namemodel consistently delivers positive transfer, whereas GPF-Plus potentially produces negative transfer. This demonstrates \namemodel's robustness even under severe shifts.

\begin{table*}[t]
%\vspace{-4mm}
\centering
%\vspace{-2mm}
\caption{Cross-dataset transfer results. All values are IMP\%. \namemodel consistently delivers positive transfer. C, T, and W denote Cornell, Texas, and Wisconsin, respectively.}
\label{tab:exp2_label_shift}
\small
\scalebox{0.8}{\begin{tabular}{lcccccc}
\toprule
 & C$\rightarrow$T & C$\rightarrow$W & T$\rightarrow$C & T$\rightarrow$W & W$\rightarrow$C & W$\rightarrow$T \\
\midrule
GPF-Plus & -3.4 & 81.1 & -5.0 & 61.9 & 33.0 & 209.8 \\
\textbf{UGPrompt} & 11.1 & 10.4 & 6.2 & 1.2 & 41.5 & 123.5 \\
\bottomrule
\end{tabular}}
%\vspace{-6mm}
\end{table*}

\vspace{-3mm}
\section{Conclusions}
\vspace{-2mm}
In conclusion, we have introduced \namemodel, a novel unsupervised prompting framework for \gnns that overcomes the limitations of existing prompting methods, particularly in scenarios where labeled data is unavailable. \namemodel eliminates the need for updating the base \gnn's parameters on new downstream tasks. \namemodel also enhances the generalization of the base source-trained \gnns without supervision. Experimental results over various datasets validate the effectiveness of \namemodel which outperforms the state-of-the-art prompting methods that rely on labeled data on both graph and node classification tasks in many settings.

\textit{Limitations \& Future Work. } 
% Since we do not involve training the projection head of the pre-trained GNN for the downstream tasks, \namemodel is unable to handle label distribution shift. Consequently, our framework is also unable to employ multi-task/meta learning unless we allow the projection head to be optimized---which necessitates labeled data. Another interesting future direction would be to design a method that selects high-quality pseudolabels in case of a severe covariate distribution shift. 
Since we do not involve training the projection head of the source-trained \gnn for downstream tasks, \namemodel assumes that the source and target domains share the same label space. Thus, it is not designed for a general label distribution shift with new target classes. This limitation is tied to \namemodel's frozen-head requirement. The lack of a unified discrete output space across graph tasks and datasets requires retraining or fine-tuning a task-specific projection head to handle new labels, which falls outside our \problemname setting. Developing universal graph vocabularies or output spaces for open-vocabulary graph prompting is an important future direction. Another interesting future direction would be to develop a method that selects high-quality pseudolabels in the presence of a severe covariate shift.

\newpage
%\subsubsection*{Broader Impact Statement}
%In this optional section, TMLR encourages authors to discuss possible repercussions of their work,
%notably any potential negative impact that a user of this research should be aware of. 
%Authors should consult the TMLR Ethics Guidelines available on the TMLR website
%for guidance on how to approach this subject.

%\subsubsection*{Author Contributions}
%If you'd like to, you may include a section for author contributions as is done
%in many journals. This is optional and at the discretion of the authors. Only add
%this information once your submission is accepted and deanonymized. 

%\subsubsection*{Acknowledgments}
%Use unnumbered third level headings for the acknowledgments. All
%acknowledgments, including those to funding agencies, go at the end of the paper.
%Only add this information once your submission is accepted and deanonymized. 

\bibliography{main}

@inproceedings{
gpf2023,
title={Universal Prompt Tuning for Graph Neural Networks},
author={Taoran Fang and Yunchao Mercer Zhang and Yang Yang and Chunping Wang and Lei CHEN},
booktitle={Neural Information Processing Systems},
year={2023},
}

@inproceedings{shot2020, 
 title={Do We Really Need to Access the Source Data? Source Hypothesis Transfer for Unsupervised Domain Adaptation}, 
 author={Liang, Jian and Hu, Dapeng and Feng, Jiashi}, 
 booktitle={International Conference on Machine Learning},  
 pages={6028--6039},
 year={2020}
}

@inproceedings{gppt2022,
author = {Sun, Mingchen and Zhou, Kaixiong and He, Xin and Wang, Ying and Wang, Xin},
title = {GPPT: Graph Pre-training and Prompt Tuning to Generalize Graph Neural Networks},
year = {2022},
publisher = {Association for Computing Machinery},
doi = {10.1145/3534678.3539249},
booktitle = {Proceedings of the 28th ACM SIGKDD Conference on Knowledge Discovery and Data Mining},
pages = {1717–1727},
}

@inproceedings{allinone2023,
author = {Sun, Xiangguo and Cheng, Hong and Li, Jia and Liu, Bo and Guan, Jihong},
title = {All in One: Multi-Task Prompting for Graph Neural Networks},
year = {2023},
booktitle = {ACM SIGKDD Conference on Knowledge Discovery and Data Mining},
pages = {2120–2131},
}

@inproceedings{
ape2022,
title={Large Language Models are Human-Level Prompt Engineers},
author={Yongchao Zhou and Andrei Ioan Muresanu and Ziwen Han and Keiran Paster and Silviu Pitis and Harris Chan and Jimmy Ba},
booktitle={The Eleventh International Conference on Learning Representations },
year={2023},
url={https://openreview.net/forum?id=92gvk82DE-}
}

@article{prodigy2023,
  title={PRODIGY: Enabling In-context Learning Over Graphs},
  author={Qian Huang and Hongyu Ren and Peng Chen and Gregor Kr\v{z}manc and Daniel Zeng and Percy Liang and Jure Leskovec},
  journal={ArXiv},
  year={2023},
  volume={abs/2305.12600}
}

@inproceedings{doweuda2020, 
 title={Do We Really Need to Access the Source Data? Source Hypothesis Transfer for Unsupervised Domain Adaptation}, 
 author={Liang, Jian and Hu, Dapeng and Feng, Jiashi}, 
 booktitle={International Conference on Machine Learning (ICML)},  
 pages={6028--6039},
 year={2020}
}

@inproceedings{discinfomax2017,
author = {Hu, Weihua and Miyato, Takeru and Tokui, Seiya and Matsumoto, Eiichi and Sugiyama, Masashi},
title = {Learning discrete representations via information maximizing self-augmented training},
year = {2017},
booktitle = {Proceedings of the 34th International Conference on Machine Learning - Volume 70},
pages = {1558–1567},
}

@inproceedings{
acloserlook2021,
title={A Closer Look at Distribution Shifts and Out-of-Distribution Generalization on Graphs},
author={Mucong Ding and Kezhi Kong and Jiuhai Chen and John Kirchenbauer and Micah Goldblum and David Wipf and Furong Huang and Tom Goldstein},
booktitle={NeurIPS 2021 Workshop on Distribution Shifts: Connecting Methods and Applications},
year={2021},
}

@inproceedings{
invariantrep2022,
title={Learning Invariant Graph Representations for Out-of-Distribution Generalization},
author={Haoyang Li and Ziwei Zhang and Xin Wang and Wenwu Zhu},
booktitle={Advances in Neural Information Processing Systems},
editor={Alice H. Oh and Alekh Agarwal and Danielle Belgrave and Kyunghyun Cho},
year={2022},
url={https://openreview.net/forum?id=acKK8MQe2xc}
}

@inproceedings{gnnexplainer2019,
author = {Ying, Zhitao and Bourgeois, Dylan and You, Jiaxuan and Zitnik, Marinka and Leskovec, Jure},
booktitle = {Advances in Neural Information Processing Systems},
title = {GNNExplainer: Generating Explanations for Graph Neural Networks},
volume = {32},
year = {2019}
}

@inproceedings{
dir2022,
title={Discovering Invariant Rationales for Graph Neural Networks},
author={Yingxin Wu and Xiang Wang and An Zhang and Xiangnan He and Tat-Seng Chua},
booktitle={International Conference on Learning Representations},
year={2022},
}

@inproceedings{
good2022,
title={{GOOD}: A Graph Out-of-Distribution Benchmark},
author={Shurui Gui and Xiner Li and Limei Wang and Shuiwang Ji},
booktitle={Thirty-sixth Conference on Neural Information Processing Systems Datasets and Benchmarks Track},
year={2022},
}

@ARTICLE {expsurvey2023,
author = {H. Yuan and H. Yu and S. Gui and S. Ji},
journal = {IEEE Transactions on Pattern Analysis and Machine Intelligence},
title = {Explainability in Graph Neural Networks: A Taxonomic Survey},
year = {2023},
volume = {45},
number = {05},
pages = {5782-5799},
}

@inproceedings{understandattn2019,
 author = {Knyazev, Boris and Taylor, Graham W and Amer, Mohamed},
 booktitle = {Advances in Neural Information Processing Systems},
 title = {Understanding Attention and Generalization in Graph Neural Networks},
 volume = {32},
 year = {2019}
}

@inproceedings{
evalrobust2023,
title={Evaluating Robustness and Uncertainty of Graph Models Under Structural Distributional Shifts},
author={Gleb Bazhenov and Denis Kuznedelev and Andrey Malinin and Artem Babenko and Liudmila Prokhorenkova},
booktitle={Thirty-seventh Conference on Neural Information Processing Systems},
year={2023},
}

@inproceedings{unleash2023,
author = {Sui, Yongduo and Wu, Qitian and Wu, Jiancan and Cui, Qing and Li, Longfei and Zhou, Jun and Wang, Xiang and He, Xiangnan},
booktitle = {Advances in Neural Information Processing Systems},
pages = {18109--18131},
title = {Unleashing the Power of Graph Data Augmentation on Covariate Distribution Shift},
volume = {36},
year = {2023}
}

@article{shiftrobust2021,
title={Shift-robust GNNs: Overcoming the limitations of localized graph training data},
author={Zhu, Qi and Ponomareva, Natalia and Han, Jiawei and Perozzi, Bryan},
journal={Advances in Neural Information Processing Systems},
volume={34},
year={2021}
}

@inproceedings{ogb2020,
author = {Hu, Weihua and Fey, Matthias and Zitnik, Marinka and Dong, Yuxiao and Ren, Hongyu and Liu, Bowen and Catasta, Michele and Leskovec, Jure},
booktitle = {Advances in Neural Information Processing Systems},
pages = {22118--22133},
title = {Open Graph Benchmark: Datasets for Machine Learning on Graphs},
volume = {33},
year = {2020}
}

@inproceedings{beyondhomophily2020,
author = {Zhu, Jiong and Yan, Yujun and Zhao, Lingxiao and Heimann, Mark and Akoglu, Leman and Koutra, Danai},
booktitle = {Advances in Neural Information Processing Systems},
pages = {7793--7804},
title = {Beyond Homophily in Graph Neural Networks: Current Limitations and Effective Designs},
volume = {33},
year = {2020}
}

@inproceedings{linkx2021,
title={Large Scale Learning on Non-Homophilous Graphs: New Benchmarks and Strong Simple Methods},
author={Derek Lim and Felix Matthew Hohne and Xiuyu Li and Sijia Linda Huang and Vaishnavi Gupta and Omkar Prasad Bhalerao and Ser-Nam Lim},
booktitle={Advances in Neural Information Processing Systems},
year={2021},
}

@inproceedings{mpnn2017,
author = {Gilmer, Justin and Schoenholz, Samuel S. and Riley, Patrick F. and Vinyals, Oriol and Dahl, George E.},
title = {Neural message passing for Quantum chemistry},
year = {2017},
booktitle = {Proceedings of the 34th International Conference on Machine Learning - Volume 70},
pages = {1263–1272},
}

@inproceedings{gpt32020,
 author = {Brown, Tom and Mann, Benjamin and Ryder, Nick and Subbiah, Melanie and Kaplan, Jared D and Dhariwal, Prafulla and Neelakantan, Arvind and Shyam, Pranav and Sastry, Girish and Askell, Amanda and Agarwal, Sandhini and Herbert-Voss, Ariel and Krueger, Gretchen and Henighan, Tom and Child, Rewon and Ramesh, Aditya and Ziegler, Daniel and Wu, Jeffrey and Winter, Clemens and Hesse, Chris and Chen, Mark and Sigler, Eric and Litwin, Mateusz and Gray, Scott and Chess, Benjamin and Clark, Jack and Berner, Christopher and McCandlish, Sam and Radford, Alec and Sutskever, Ilya and Amodei, Dario},
 booktitle = {Advances in Neural Information Processing Systems},
 editor = {H. Larochelle and M. Ranzato and R. Hadsell and M.F. Balcan and H. Lin},
 pages = {1877--1901},
 title = {Language Models are Few-Shot Learners},
 volume = {33},
 year = {2020}
}

@article{gpt22019,
  title={Language Models are Unsupervised Multitask Learners},
  author={Radford, Alec and Wu, Jeff and Child, Rewon and Luan, David and Amodei, Dario and Sutskever, Ilya},
  year={2019}
}

@article{proteinfunction2005,
author = {Borgwardt, Karsten M. and Ong, Cheng Soon and Schönauer, Stefan and Vishwanathan, S. V. N. and Smola, Alex J. and Kriegel, Hans-Peter},
title = {Protein function prediction via graph kernels},
journal = {Bioinformatics},
volume = {21},
pages = {i47-i56},
year = {2005}
}

@article{enzymesdb2004,
author = {Schomburg, Ida and Chang, Antje and Ebeling, Christian and Gremse, Marion and Heldt, Christian and Huhn, Gregor and Schomburg, Dietmar},
title = {BRENDA, the enzyme database: updates and major new developments},
journal = {Nucleic Acids Research},
volume = {32},
pages = {D431-D433},
year = {2004}
}

@article{dhfr2003,
author = {Sutherland, Jeffrey J. and O'Brien, Lee A. and Weaver, Donald F.},
title = {Spline-Fitting with a Genetic Algorithm: A Method for Developing Classification Structure-Activity Relationships},
journal = {Journal of Chemical Information and Computer Sciences},
volume = {43},
number = {6},
pages = {1906-1915},
year = {2003},
}

@inproceedings{planetoid2016,
author = {Yang, Zhilin and Cohen, William W. and Salakhutdinov, Ruslan},
title = {Revisiting semi-supervised learning with graph embeddings},
year = {2016},
booktitle = {Proceedings of the 33rd International Conference on International Conference on Machine Learning - Volume 48},
pages = {40–48},
}

@inproceedings{bert2019,
title = "{BERT}: Pre-training of Deep Bidirectional Transformers for Language Understanding",
author = "Devlin, Jacob  and
  Chang, Ming-Wei  and
  Lee, Kenton  and
  Toutanova, Kristina",
booktitle = "Proceedings of the 2019 Conference of the North {A}merican Chapter of the Association for Computational Linguistics: Human Language Technologies, Volume 1 (Long and Short Papers)",
year = "2019",
pages = "4171--4186",
}

@article{infonce2018,
title={Representation Learning with Contrastive Predictive Coding},
author={A{\"a}ron van den Oord and Yazhe Li and Oriol Vinyals},
journal={ArXiv},
year={2018},
volume={abs/1807.03748},
}

@inproceedings{simclr2020,
author = {Chen, Ting and Kornblith, Simon and Norouzi, Mohammad and Hinton, Geoffrey},
title = {A simple framework for contrastive learning of visual representations},
year = {2020},
booktitle = {Proceedings of the 37th International Conference on Machine Learning},
numpages = {11},
}

@inproceedings{transfernlp2019,
title = "Transfer Learning in Natural Language Processing",
author = "Ruder, Sebastian  and
  Peters, Matthew E.  and
  Swayamdipta, Swabha  and
  Wolf, Thomas",
booktitle = "Proceedings of the 2019 Conference of the North {A}merican Chapter of the Association for Computational Linguistics: Tutorials",
year = "2019",
pages = "15--18",
}

@ARTICLE{transfervision2021,
author={Zhuang, Fuzhen and Qi, Zhiyuan and Duan, Keyu and Xi, Dongbo and Zhu, Yongchun and Zhu, Hengshu and Xiong, Hui and He, Qing},
journal={Proceedings of the IEEE}, 
title={A Comprehensive Survey on Transfer Learning}, 
year={2021},
volume={109},
pages={43-76},
}

@inproceedings{graphprompt2023,
author = {Liu, Zemin and Yu, Xingtong and Fang, Yuan and Zhang, Xinming},
title = {GraphPrompt: Unifying Pre-Training and Downstream Tasks for Graph Neural Networks},
year = {2023},
publisher = {Association for Computing Machinery},
booktitle = {Proceedings of the ACM Web Conference 2023},
pages = {417–428},
}

@inproceedings{simgrace20222,
author = {Xia, Jun and Wu, Lirong and Chen, Jintao and Hu, Bozhen and Li, Stan Z.},
title = {SimGRACE: A Simple Framework for Graph Contrastive Learning without Data Augmentation},
year = {2022},
booktitle = {Proceedings of the ACM Web Conference 2022},
pages = {1070–1079},
}

@inproceedings{graphcl2020,
author = {You, Yuning and Chen, Tianlong and Sui, Yongduo and Chen, Ting and Wang, Zhangyang and Shen, Yang},
booktitle = {Advances in Neural Information Processing Systems},
pages = {5812--5823},
title = {Graph Contrastive Learning with Augmentations},
volume = {33},
year = {2020}
}

@article{pseudolabel2013,
author = {Lee, Dong-Hyun},
year = {2013},
title = {Pseudo-Label : The Simple and Efficient Semi-Supervised Learning Method for Deep Neural Networks},
journal = {ICML 2013 Workshop : Challenges in Representation Learning (WREPL)}
}

@inproceedings{pseudoensembel2014,
author = {Bachman, Philip and Alsharif, Ouais and Precup, Doina},
title = {Learning with pseudo-ensembles},
year = {2014},
booktitle = {Proceedings of the 27th International Conference on Neural Information Processing Systems - Volume 2},
pages = {3365–3373},
}

@inproceedings{stochasticregu2016,
author = {Sajjadi, Mehdi and Javanmardi, Mehran and Tasdizen, Tolga},
title = {Regularization with stochastic transformations and perturbations for deep semi-supervised learning},
year = {2016},
booktitle = {Proceedings of the 30th International Conference on Neural Information Processing Systems},
pages = {1171–1179},
}

@inproceedings{temporalensemble2017,
title={Temporal Ensembling for Semi-Supervised Learning},
author={Samuli Laine and Timo Aila},
booktitle={International Conference on Learning Representations},
year={2017},
}

@inproceedings{fixmatch2020,
author = {Sohn, Kihyuk and Berthelot, David and Li, Chun-Liang and Zhang, Zizhao and Carlini, Nicholas and Cubuk, Ekin D. and Kurakin, Alex and Zhang, Han and Raffel, Colin},
title = {FixMatch: simplifying semi-supervised learning with consistency and confidence},
year = {2020},
booktitle = {Proceedings of the 34th International Conference on Neural Information Processing Systems},
}

@inproceedings{flexmatch2021,
author = {Zhang, Bowen and Wang, Yidong and Hou, Wenxin and WU, HAO and Wang, Jindong and Okumura, Manabu and Shinozaki, Takahiro},
booktitle = {Advances in Neural Information Processing Systems},
pages = {18408--18419},
title = {FlexMatch: Boosting Semi-Supervised Learning with Curriculum Pseudo Labeling},
volume = {34},
year = {2021}
}

@inproceedings{freematch2023,
title={FreeMatch: Self-adaptive Thresholding for Semi-supervised Learning},
author={Yidong Wang and Hao Chen and Qiang Heng and Wenxin Hou and Yue Fan and Zhen Wu and Jindong Wang and Marios Savvides and Takahiro Shinozaki and Bhiksha Raj and Bernt Schiele and Xing Xie},
booktitle={The Eleventh International Conference on Learning Representations },
year={2023},
}

@inproceedings{gcn2017,
title={Semi-Supervised Classification with Graph Convolutional Networks},
author={Thomas N. Kipf and Max Welling},
booktitle={International Conference on Learning Representations},
year={2017},
}

@inproceedings{gat2018,
title={Graph Attention Networks},
author={Petar Veličković and Guillem Cucurull and Arantxa Casanova and Adriana Romero and Pietro Liò and Yoshua Bengio},
booktitle={International Conference on Learning Representations},
year={2018},
}

@inproceedings{zerocot2022,
author = {Kojima, Takeshi and Gu, Shixiang (Shane) and Reid, Machel and Matsuo, Yutaka and Iwasawa, Yusuke},
booktitle = {Advances in Neural Information Processing Systems},
pages = {22199--22213},
title = {Large Language Models are Zero-Shot Reasoners},
volume = {35},
year = {2022}
}

@inproceedings{opro2024,
title={Large Language Models as Optimizers},
author={Chengrun Yang and Xuezhi Wang and Yifeng Lu and Hanxiao Liu and Quoc V Le and Denny Zhou and Xinyun Chen},
booktitle={The Twelfth International Conference on Learning Representations},
year={2024},
}

@inproceedings{prefixtuning2021,
title = "Prefix-Tuning: Optimizing Continuous Prompts for Generation",
author = "Li, Xiang Lisa  and
  Liang, Percy",
booktitle = "Proceedings of the 59th Annual Meeting of the Association for Computational Linguistics and the 11th International Joint Conference on Natural Language Processing",
year = "2021",
pages = "4582--4597",
}

@inproceedings{cot2022,
author = {Wei, Jason and Wang, Xuezhi and Schuurmans, Dale and Bosma, Maarten and ichter, brian and Xia, Fei and Chi, Ed and Le, Quoc V and Zhou, Denny},
booktitle = {Advances in Neural Information Processing Systems},
pages = {24824--24837},
publisher = {Curran Associates, Inc.},
title = {Chain-of-Thought Prompting Elicits Reasoning in Large Language Models},
volume = {35},
year = {2022}
}

@misc{gfm2024,
title={Future Directions in the Theory of Graph Machine Learning}, 
author={Christopher Morris and Fabrizio Frasca and Nadav Dym and Haggai Maron and İsmail İlkan Ceylan and Ron Levie and Derek Lim and Michael Bronstein and Martin Grohe and Stefanie Jegelka},
year={2024},
eprint={2402.02287},
archivePrefix={arXiv},
}

@inproceedings{gpt12018,
  title={Improving Language Understanding by Generative Pre-Training},
  author={Alec Radford and Karthik Narasimhan},
  year={2018},
}

@misc{position2024,
title={Position: Graph Foundation Models are Already Here}, 
author={Haitao Mao and Zhikai Chen and Wenzhuo Tang and Jianan Zhao and Yao Ma and Tong Zhao and Neil Shah and Mikhail Galkin and Jiliang Tang},
year={2024},
eprint={2402.02216},
archivePrefix={arXiv},
}

@inproceedings{
talkgraph2024,
title={Talk like a Graph: Encoding Graphs for Large Language Models},
author={Bahare Fatemi and Jonathan Halcrow and Bryan Perozzi},
booktitle={The Twelfth International Conference on Learning Representations},
year={2024},
}

@inproceedings{reframing2022,
title = "Reframing Instructional Prompts to {GPT}k{'}s Language",
author = "Mishra, Swaroop  and
  Khashabi, Daniel  and
  Baral, Chitta  and
  Choi, Yejin  and
  Hajishirzi, Hannaneh",
booktitle = "Findings of the Association for Computational Linguistics: ACL 2022",
year = "2022",
pages = "589--612",
}

@ARTICLE{domainwithoutsource2021,
author={Kim, Youngeun and Cho, Donghyeon and Han, Kyeongtak and Panda, Priyadarshini and Hong, Sungeun},
journal={IEEE Transactions on Artificial Intelligence}, 
title={Domain Adaptation Without Source Data}, 
year={2021},
volume={2},
pages={508-518}
}

@article{selftraining2020,
title={Self-Training With Noisy Student Improves ImageNet Classification},
author={Qizhe Xie and Eduard H. Hovy and Minh-Thang Luong and Quoc V. Le},
journal={Conference on Computer Vision and Pattern Recognition (CVPR)},
year={2020},
pages={10684-10695},
}

@article{pytorch2017,
title={Automatic differentiation in PyTorch},
author={Paszke, Adam and Gross, Sam and Chintala, Soumith and Chanan, Gregory and Yang, Edward and DeVito, Zachary and Lin, Zeming and Desmaison, Alban and Antiga, Luca and Lerer, Adam},
year={2017}
}

@inproceedings{powerofscale2021,
    title = "The Power of Scale for Parameter-Efficient Prompt Tuning",
    author = "Lester, Brian  and
      Al-Rfou, Rami  and
      Constant, Noah",
    editor = "Moens, Marie-Francine  and
      Huang, Xuanjing  and
      Specia, Lucia  and
      Yih, Scott Wen-tau",
    booktitle = "Proceedings of the 2021 Conference on Empirical Methods in Natural Language Processing",
    year = "2021",
    pages = "3045--3059",
}

@article{promptsurvey2023,
author = {Liu, Pengfei and Yuan, Weizhe and Fu, Jinlan and Jiang, Zhengbao and Hayashi, Hiroaki and Neubig, Graham},
title = {Pre-train, Prompt, and Predict: A Systematic Survey of Prompting Methods in Natural Language Processing},
year = {2023},
volume = {55},
number = {9},
issn = {0360-0300},
journal = {ACM Comput. Surv.},
numpages = {35},
}

@book{understandml2014, 
place={Cambridge}, 
title={Understanding Machine Learning: From Theory to Algorithms}, publisher={Cambridge University Press}, 
author={Shalev-Shwartz, Shai and Ben-David, Shai}, 
year={2014}
}

@Inbook{statisticalearning2004,
author="Bousquet, Olivier
and Boucheron, St{\'e}phane
and Lugosi, G{\'a}bor",
editor="Bousquet, Olivier
and von Luxburg, Ulrike
and R{\"a}tsch, Gunnar",
title="Introduction to Statistical Learning Theory",
bookTitle="Advanced Lectures on Machine Learning: ML Summer Schools 2003, Canberra, Australia, February 2 - 14, 2003, T{\"u}bingen, Germany, August 4 - 16, 2003, Revised Lectures",
year="2004",
publisher="Springer Berlin Heidelberg",
pages="169--207",
}

@article{oodgnn2023,
author = {Li, Haoyang and Wang, Xin and Zhang, Ziwei and Zhu, Wenwu},
title = {OOD-GNN: Out-of-Distribution Generalized Graph Neural Network},
year = {2023},
volume = {35},
number = {7},
issn = {1041-4347},
journal = {IEEE Trans. on Knowl. and Data Eng.},
pages = {7328–7340},
numpages = {13}
}

@ARTICLE{gnnood2024,
  author={Fan, Shaohua and Wang, Xiao and Shi, Chuan and Cui, Peng and Wang, Bai},
  journal={IEEE Transactions on Pattern Analysis and Machine Intelligence}, 
  title={Generalizing Graph Neural Networks on Out-of-Distribution Graphs}, 
  year={2024},
  volume={46},
  number={1},
  pages={322-337},
}

@article{generalood2024,
author = {Yang, Jingkang and Zhou, Kaiyang and Li, Yixuan and Liu, Ziwei},
title = {Generalized Out-of-Distribution Detection: A Survey},
year = {2024},
volume = {132},
number = {12},
issn = {0920-5691},
journal = {Int. J. Comput. Vision},
pages = {5635–5662},
numpages = {28},
}

@InProceedings{reviewda2021,
author="Farahani, Abolfazl
and Voghoei, Sahar
and Rasheed, Khaled
and Arabnia, Hamid R.",
editor="Stahlbock, Robert
and Weiss, Gary M.
and Abou-Nasr, Mahmoud
and Yang, Cheng-Ying
and Arabnia, Hamid R.
and Deligiannidis, Leonidas",
title="A Brief Review of Domain Adaptation",
booktitle="Advances in Data Science and Information Engineering",
year="2021",
publisher="Springer International Publishing",
pages="877--894",
}

@InProceedings{uda2019,
author = {You, Kaichao and Long, Mingsheng and Cao, Zhangjie and Wang, Jianmin and Jordan, Michael I.},
title = {Universal Domain Adaptation},
booktitle = {Proceedings of the IEEE/CVF Conference on Computer Vision and Pattern Recognition (CVPR)},
year = {2019}
}

@ARTICLE{surveysda2024,
  author={Li, Jingjing and Yu, Zhiqi and Du, Zhekai and Zhu, Lei and Shen, Heng Tao},
  journal={IEEE Transactions on Pattern Analysis and Machine Intelligence}, 
  title={A Comprehensive Survey on Source-Free Domain Adaptation}, 
  year={2024},
  volume={46},
  number={8},
  pages={5743-5762}
}

@INPROCEEDINGS{generalsda2021,
  author={Yang, Shiqi and Wang, Yaxing and van de Weijer, Joost and Herranz, Luis and Jui, Shangling},
  booktitle={2021 IEEE/CVF International Conference on Computer Vision (ICCV)}, 
  title={Generalized Source-free Domain Adaptation}, 
  year={2021},
  pages={8958-8967},
}

@inproceedings{
prog2024,
title={ProG: A Graph Prompt Learning Benchmark},
author={Chenyi Zi and Haihong Zhao and Xiangguo Sun and Yiqing Lin and Hong Cheng and Jia Li},
booktitle={The Thirty-eight Conference on Neural Information Processing Systems Datasets and Benchmarks Track},
year={2024},
}

@ARTICLE{graphpromptplus2024,
  author={Yu, Xingtong and Liu, Zhenghao and Fang, Yuan and Liu, Zemin and Chen, Sihong and Zhang, Xinming},
  journal={IEEE Transactions on Knowledge and Data Engineering}, 
  title={Generalized Graph Prompt: Toward a Unification of Pre-Training and Downstream Tasks on Graphs}, 
  year={2024},
  volume={36},
  number={11},
  pages={6237-6250}
}

@inproceedings{multigprompt2024,
author = {Yu, Xingtong and Zhou, Chang and Fang, Yuan and Zhang, Xinming},
title = {MultiGPrompt for Multi-Task Pre-Training and Prompting on Graphs},
year = {2024},
isbn = {9798400701719},
publisher = {Association for Computing Machinery},
booktitle = {Proceedings of the ACM Web Conference 2024},
pages = {515–526},
numpages = {12},
}

@inproceedings{
dygprompt2025,
title={Node-Time Conditional Prompt Learning in Dynamic Graphs},
author={Xingtong Yu and Zhenghao Liu and Xinming Zhang and Yuan Fang},
booktitle={The Thirteenth International Conference on Learning Representations},
year={2025},
}

@article{
parameterefficient2024,
title={Parameter-Efficient Fine-Tuning for Large Models: A Comprehensive Survey},
author={Zeyu Han and Chao Gao and Jinyang Liu and Jeff Zhang and Sai Qian Zhang},
journal={Transactions on Machine Learning Research},
issn={2835-8856},
year={2024},
}

@inproceedings{
graphsaint2020,
title={GraphSAINT: Graph Sampling Based Inductive Learning Method},
author={Hanqing Zeng and Hongkuan Zhou and Ajitesh Srivastava and Rajgopal Kannan and Viktor Prasanna},
booktitle={International Conference on Learning Representations},
year={2020},
}

@article{moleculenet2017,
  title={MoleculeNet: a benchmark for molecular machine learning},
  author={Zhenqin Wu and Bharath Ramsundar and Evan N. Feinberg and Joseph Gomes and Caleb Geniesse and Aneesh S. Pappu and Karl Leswing and Vijay S. Pande},
  journal={Chemical Science},
  year={2017},
  volume={9},
  pages={513 - 530},
}

@inproceedings{
geomgcn2020,
title={Geom-GCN: Geometric Graph Convolutional Networks},
author={Hongbin Pei and Bingzhe Wei and Kevin Chen-Chuan Chang and Yu Lei and Bo Yang},
booktitle={International Conference on Learning Representations},
year={2020},
}

@inproceedings{soga2024,
author = {Mao, Haitao and Du, Lun and Zheng, Yujia and Fu, Qiang and Li, Zelin and Chen, Xu and Han, Shi and Zhang, Dongmei},
title = {Source Free Graph Unsupervised Domain Adaptation},
year = {2024},
booktitle = {Proceedings of the 17th ACM International Conference on Web Search and Data Mining},
pages = {520–528},
}

@inproceedings{graphcta2024,
author = {Zhang, Zhen and Liu, Meihan and Wang, Anhui and Chen, Hongyang and Li, Zhao and Bu, Jiajun and He, Bingsheng},
title = {Collaborate to Adapt: Source-Free Graph Domain Adaptation via Bi-directional Adaptation},
year = {2024},
booktitle = {Proceedings of the ACM Web Conference 2024},
pages = {664–675},
}

@article{beyondfinetuning2025,
  author    = {Zheng, Hongling and Shen, Li and Tang, Anke and Luo, Yong and Hu, Han and Du, Bo and Wen, Yonggang and Tao, Dacheng},
  title     = {Learning from models beyond fine-tuning},
  journal   = {Nature Machine Intelligence},
  year      = {2025},
  volume    = {7},
  number    = {1},
  pages     = {6--17},
}

@inproceedings{
gatvII2022,
title={How Attentive are Graph Attention Networks? },
author={Shaked Brody and Uri Alon and Eran Yahav},
booktitle={International Conference on Learning Representations},
year={2022},
}

@inproceedings{
graphgps2022,
title={Recipe for a General, Powerful, Scalable Graph Transformer},
author={Ladislav Rampasek and Mikhail Galkin and Vijay Prakash Dwivedi and Anh Tuan Luu and Guy Wolf and Dominique Beaini},
booktitle={Advances in Neural Information Processing Systems},
year={2022},
}

@inproceedings{pronog2025,
author = {Yu, Xingtong and Zhang, Jie and Fang, Yuan and Jiang, Renhe},
title = {Non-Homophilic Graph Pre-Training and Prompt Learning},
year = {2025},
isbn = {9798400712456},
publisher = {Association for Computing Machinery},
booktitle = {Proceedings of the 31st ACM SIGKDD Conference on Knowledge Discovery and Data Mining V.1},
pages = {1844–1854},
numpages = {11},
location = {Toronto ON, Canada},
series = {KDD '25}
}

@inproceedings{dagprompt2025,
author = {Chen, Qin and Wang, Liang and Zheng, Bo and Song, Guojie},
title = {DAGPrompT: Pushing the Limits of Graph Prompting with a Distribution-aware Graph Prompt Tuning Approach},
year = {2025},
isbn = {9798400712746},
publisher = {Association for Computing Machinery},
booktitle = {Proceedings of the ACM on Web Conference 2025},
pages = {4346–4358},
numpages = {13},
location = {Sydney NSW, Australia},
series = {WWW '25}
}

@inproceedings{gcope2024,
author = {Zhao, Haihong and Chen, Aochuan and Sun, Xiangguo and Cheng, Hong and Li, Jia},
title = {All in One and One for All: A Simple yet Effective Method towards Cross-domain Graph Pretraining},
year = {2024},
isbn = {9798400704901},
publisher = {Association for Computing Machinery},
booktitle = {Proceedings of the 30th ACM SIGKDD Conference on Knowledge Discovery and Data Mining},
pages = {4443–4454},
numpages = {12},
location = {Barcelona, Spain},
series = {KDD '24}
}

@inproceedings{
lora2022,
title={Lo{RA}: Low-Rank Adaptation of Large Language Models},
author={Edward J Hu and yelong shen and Phillip Wallis and Zeyuan Allen-Zhu and Yuanzhi Li and Shean Wang and Lu Wang and Weizhu Chen},
booktitle={International Conference on Learning Representations},
year={2022},
url={https://openreview.net/forum?id=nZeVKeeFYf9}
}
\bibliographystyle{tmlr}

%\appendix

\appendix
%\startcontents[appendices]

%\begin{appendices}

% 3. Print the tracked contents for the 'appendices' group.
% The last argument provides the title.
%\printcontents[appendices]{}{1}{\section*{Appendix Contents}}
\clearpage % Optional: start the actual appendix content on a new page

\section{Appendix}
\subsection{Additional Related Work}
\paragraph{Prompting for LLMs.}
The promising results of early LLMs \citep{gpt12018, bert2019} have inspired prompt-tuning approaches that leverage LLMs' reasoning capabilities with minimal parameter tuning. These modular methods \citep{prefixtuning2021, powerofscale2021} integrate trainable prompting prefixes with LLMs, with their parameters frozen, offering remarkable performance across various tasks while reducing complexity. Besides, \citep{gpt22019} suggested pre-training on a large, diverse corpus (WebText) and demonstrated strong zero-shot performance across various tasks. Following that, \citep{gpt32020} coined the term in-context learning, which refers to the effort to help models, particularly LLMs, generalize to new tasks without any parameter training, only by constructing informative prompts with task descriptions, examples, and instructions. The approaches propose specific language hints to guide the reasoning process, for example, by providing fine-grained and conditionally specific instructions \citep{reframing2022}, or by encouraging the LLM to engage in sequential reasoning with \citep{cot2022} or without \citep{zerocot2022} examples. Additionally, some methods iteratively score, evaluate, and update prompts in refinement loops \citep{ape2022, opro2024}.

\paragraph{Prompting for GNNs.}
A few studies have also adopted prompt tuning for \gnns. Specifically, the main track of these works starts with the ``pre-train, prompt, fine-tune'' paradigm proposed by \citep{gppt2022}; they address the common issue of a discrepancy between training and test objectives that causes performance drop in downstream applications. They design a pre-training task, specifically edge prediction, that can align with the downstream task. Nonetheless, their approach is only applicable to node classification. Later, GraphPrompt \citep{graphprompt2023} and GraphPrompt+ \citep{graphpromptplus2024} propose subgraph similarity detection as a more general \textit{task unification} for different graph tasks to pre-train a \gnn encoder. Their prompting method involves task-specific trainable readout functions. Similarly, PRODIGY \citep{prodigy2023} generates prompts as a combination of example subgraphs, connected to label nodes, and query subgraphs, awaiting connection to label nodes, and pre-trains with a neighborhood-matching task. Unification to subgraph classification is also proposed by GPF-Plus \citep{gpf2023} and ALL-In-One \citep{allinone2023}. However, these methods resemble LLM prompting methods more, in that they prompt input graphs before a frozen \gnn encoder. Specifically, the former adds trainable prompting parameters to the feature matrix of an input graph, and the latter combines a subgraph with a trainable feature matrix and structure with the input. More recently, GCOPE \citep{gcope2024} utilizes graph coordinators to align domain-shifted datasets and remedy negative transfer in cross-domain pre-training. Also, DAGPrompt \citep{dagprompt2025} introduces distribution-aware prompting by utilizing low-rank adaptation for heterophilic graphs. Similarly, to capture diverse node-specific patterns in non-homophilic graphs, PRONOG \citep{pronog2025} introduces a conditional prompting. In addition, Prompting GNNs for dynamic graphs has also been studied lately \citep{dygprompt2025}.

There are two shortcomings with the above \gnn prompting methods. First, all of them require labeled data for training their prompting functions or for testing. Second, they mostly train a new projection head/decoder for the pre-trained \gnn along with the prompting parameters. However, we propose a fully unsupervised prompting method that does not require fine-tuning and achieves promising performance even when competitors have access to all or a subset of the labels.

\paragraph{Consistency Regularization and Pseudo-labeling.}
In the context of Semi-Supervised Learning (SSL), pseudo-labeling \citep{pseudolabel2013, selftraining2020} and consistency regularization \citep{pseudoensembel2014, stochasticregu2016, temporalensemble2017} for training neural network models when labeled data are scarce. The first technique augments the labeled dataset with the model's predictions on the unlabeled dataset, and the second aims to minimize the discrepancy in a model's predictions across random perturbations of the same samples generated by augmentation and dropout layers. In particular, these methods have been applied to domain adaptation to mitigate distribution shifts between source and target datasets \citep{shot2020, domainwithoutsource2021}. A widely studied approach to integrating pseudo-labeling and consistency regularization is to first generate random weakly and strongly augmented instances from the same dataset samples, and then assign pseudo-labels to the weakly augmented samples whenever the model makes confident predictions for them. Models are trained by these pseudo-labels as labels for the strongly augmented samples, along with labeled data if available. To select confident samples, some methods use a fixed certainty threshold \citep{fixmatch2020} while others set dynamic class-wise thresholds \citep{flexmatch2021, freematch2023}. 

Unlike previous work that used consistency regularization and pseudo-labeling, we are not interested in training or fine-tuning a model. Therefore, our novelty lies in interpreting a prompted graph as a strongly augmented instance and using the pseudo-labels from weak augmentation to train the prompting parameters. 

\subsection{Algorithm}
Algorithm \ref{alg:ugprompttraining} presents our method's prompting procedure during training. In line 1, we initialize the base \gnn model with trained parameters tuned on the source and target datasets, the augmentation and prompting functions, and the method's hyperparameters. We initialize the prompting parameter and fix the base \gnn's parameters in lines 3-4. Line 6 shows sampling of a batch of graphs and in line 7 a strong and a weak augmentation is generated from each graph in the batch. Next, the strongly augmented graph is prompted in line 8. This prompted graph, along with the weakly augmented graph, is encoded by \gnns, as in line 9. In line 10, the representation of the prompted and weakly augmented graphs is used to optimize the discriminator. Finally, in lines 9-10, the \gnn's projection head (decoder) decodes the representation, and the diversity, domain adaptation, and consistency objective functions are used to optimize the prompting parameters. Algorithm \ref{alg:ugpromptinference} shows the inference stage of our method. At inference time, a test sample is passed directly to the prompting function without augmentation, and the GNN scores the prompted graph. %\sm{what scores? classifications? be a little more detailed}

\begin{algorithm}[h]
\caption{\namemodel (Training)} \label{alg:ugprompttraining}
\begin{algorithmic}[1]
    \STATE \textbf{Input:} Target unlabeled dataset $\data = \{\graph_i\}_{i=1}^{N_t}$, confidence threshold $\tau$, \gnn model $\varphi = h(.; \theta_h) \circ g(.; \theta_g)$, number of trainable prompting parameters $n_p$, augmentation function $aug(.;p)$, weak augmentation probability $p_w$, and strong augmentation probability $p_s$.
    \STATE \textbf{Output:} Prompting function $f(.; \theta_f)$ with optimized parameters $\theta^\star$.
    \vspace{1em}

    \STATE Initialize prompting parameters $\theta_f$.
    \STATE Freeze the parameters $\theta_g$ and $\theta_h$ of the base \gnn.
    \WHILE{not converged}
        \STATE Sample batch of graphs $\mathcal{B}=\{\graph_i\}_{i=1}^{|\mathcal{B}|} \subset \data$.
        \STATE For every graph $\graph \in \mathcal{B}$ make a weak augmentation $\graph_w \leftarrow aug(\graph, p_w)$ and a strong augmentation $\graph_s \leftarrow aug(\graph, p_s)$.
        \STATE Prompt every strongly augmented graph as $\graph_p \leftarrow f(\graph_s; \theta_f)$.
        \STATE Using encoder $g$, encode every weakly augmented graph as $z^{\graph}_a=g(\graph_w; \theta_g)$ and every prompted graph as $z^{\graph}_p=g(\graph_p; \theta_g)$.
        \STATE Pass all encodings $z^{\graph}_a$ and $z^{\graph}_p$ to domain discriminator and optimize its parameters $\theta_d$ by Equation (\ref{eq:discloss}).
        \STATE Using decoder $h$, compute prediction scores for all encodings $z^{\graph}_a$ and $z^{\graph}_p$ as in Equation (\ref{eq:predscores}).
        \STATE Optimize for the prompting parameters $\theta_f$ by Equation (\ref{eq:totalloss}).
    \ENDWHILE
    \STATE \textbf{return} $f(.; \theta^\star)$.
\end{algorithmic}
\end{algorithm}

\begin{algorithm}[h]
\caption{\namemodel (Inference)} \label{alg:ugpromptinference}
\begin{algorithmic}[1]
    \STATE \textbf{Input:} Target unlabeled dataset $\data = \{\graph_i\}_{i=1}^{N_t}$, confidence threshold $\tau$, freezed pretrained \gnn $\varphi = h(.; \theta_h) \circ g(.; \theta_g)$, optimized prompting function $f(.; \theta_f^\star)$
    \STATE \textbf{Output:} Prediction score set $\text{PRED}_{\data}=\{\varphi(f(\graph_i; \theta_f^\star); \theta_h, \theta_g); \graph_i\in \data\}_{i=1}^{N_t}$
    \vspace{1em}

    \STATE Initialize empty set $\text{PRED}=\{\}$.
    \FOR{$\graph \in \data$}
        \STATE Directly pass $\graph$ to $f$ and make a prompted graph as $\graph_p \leftarrow f(\graph; \theta_f^\star)$.
        \STATE Label the prompted graph using the source-trained \gnn as $\hat{y}^{\graph}_{\varphi}=\argmax \varphi(\graph_p; \theta_h, \theta_g)$
        \STATE Add $\graph$'s label to the to the prediction set as $\text{PRED} \leftarrow \text{PRED} \cup \{\hat{y}^{\graph}_{\varphi}\}$
    \ENDFOR
    \STATE \textbf{return} $\text{PRED}$.
\end{algorithmic}
\end{algorithm}

\subsection{Remark on UGPP definition.} \label{apdx:ugpp}
Firstly, our problem setting differs from out-of-distribution (OOD) generalization methods \citep{oodgnn2023, gnnood2024, generalood2024}. In OOD generalization, given samples of dataset $\data=\{(x_i, y_i)\}_{i=1}^{N}$ drawn from training distribution $P_{train}(X, Y)$, the goal is to train an optimal model $f(.; \theta)$ to have the best generalization to the test samples drawn from the distribution $P_{test}(X, Y)$, where $P_{train}(X, Y) \neq P_{test}(X, Y)$. This differs from our problem setting, as our goal is to propose a prompting method for GNNs that aligns with the in-context learning paradigm of LLMs. As discussed, LLM prompting methods commonly 1) do not retrain or fine-tune the LLM, 2) do not necessarily use labeled data, and 3) do not assume access to the data used for training the LLM. \textsc{Ugpp} directly encourages these properties. We assume the GNN $\varphi(.; \theta_g, \theta_h)$ is first trained on distribution $\prob_X^s\prob_{Y|X}^s$ and $\data_s$ has the same train and test distributions, i.e. $\prob_{X,Y}^{s_{train}} = \prob_{X,Y}^{s_{test}} = \prob_X^s\prob_{Y|X}^s$, while this model is aimed to be used on $\data_t$ with unlabeled training distribution $\prob_{X,Y}^{t_{train}}$ and test distribution $\prob_{X,Y}^{t_{test}}$, such that $\prob_{X}^{t_{train}} = \prob_{X}^{t_{test}} \neq \prob_{X}^{s}$, but $\prob_{Y|X}^{t_{train}} = \prob_{Y|X}^{t_{test}} = \prob_{Y|X}^{s}$ and $\prob_{Y|X}^{t_{train}}$ is not available. So far, this assumption of \textsc{Ugpp} makes it close to the unsupervised domain adaptation (UDA) problem \citep{uda2019, reviewda2021} and satisfies unsupervised learning on $\data_t$, the second property of LLM in-context learning methods discussed above. For the third property, we assume the source dataset $\data_s$ is unobservable after training, which is known as source-free domain adaptation (SFDA) \citep{generalsda2021, surveysda2024}. For the first property, we assume all the trained GNN's parameters ($\theta_g, \theta_h$) are fixed and we do not fine-tune them on dataset $\data_t$, analogous to LLM in-context learning. So, our work also differs from SFDA methods, which allow fine-tuning all or a portion of the parameters on $\data_t$. In summary, our problem setting differs from UDA, SFDA, and OOD generalization, and it encourages the evaluation and design of generalized prompting methods for GNNs.

\subsection{Additional Details on Experimental Setup}
\label{apdx:add_exp_setup}
\subsubsection{Datasets Statistics}\label{apdx:datasets}
In this work, we use datasets with different tasks and types of features. For graph classification, we evaluate on bioinformatics and molecular datasets, specifically ENZYMES \citep{enzymesdb2004} for multi-class classification and DHFR \citep{dhfr2003} for binary classification (both with continuous features), PROTEINS \citep{proteinfunction2005} with discrete features for binary classification, also on BBBP and BACE \citep{moleculenet2017}. For node classification, we use common citation networks Cora, CiteSeer, PubMed \citep{planetoid2016}, and Flickr \citep{graphsaint2020} with discrete features of online images for multi-class classification. We also experimented on Cornell, Texas, and Wisconsin \citep{geomgcn2020} as semantic web datasets with discrete features. Table \ref{tab:datasets} shows the statistics of these datasets. 

% \subsection{Remark on Excluding LLM-based baselines} \review{LLM-based approaches do not align with our setup. We focus on \gnn prompt tuning, where a prompting module enhances a fixed pre-trained \gnn’s performance on a target dataset. In contrast, all known LLM-based methods for graph tasks train an LLM \citep{canlarge2023, llaga2024} or an LLM-\gnn integrated \citep{oneforall2024, graphprompter2024} model from scratch and do not improve a fixed pre-trained \gnn.

% In addition, LLM-based methods have the following limitations \citep{surveyllmgnnkdd2024}: 1) They are restricted to Text Attributed Graphs (TAGs), while ENZYMES, PROTEINS, DHFR, and Flickr in our experiments are not TAGs. 2) Most LLM-based methods, like LLaGA, focus on node-/edge-level tasks by inputting small neighborhoods around nodes into the LLM. However, graph classification samples cannot be easily converted into textual sequences. 3) Current LLM-based methods, to our knowledge, ignore non-textual node attributes and only consider node labels.}
% \sm{we don't need this }

\subsubsection{Details on the Distribution Shift}\label{apdx:dist_shift}
Introducing distribution shift to graphs is challenging. First, the position of nodes does not matter in graphs, therefore, introducing distribution shift by geometric transformations is not an option. Second, it is hard to find invariant features among all graphs for universal manipulations to inject shift---like color manipulations in image domain since color channels are common features. Moreover, random perturbation to node features cannot be seen as a distribution shift because it may lead to noisy datasets rather than a distribution shift with some patterns. We study different distribution shifts in the graph domain to make a comprehensive evaluation for the setting of \problemname. Here is a brief review of these distribution shifts. 

\textbf{Node-level. } Previous works study two main categories of distribution shift for nodes: 1- based on added random noise to node features \citep{understandattn2019} 2- based on structural properties such node degrees \citep{good2022, expsurvey2023}, clustering coefficient \citep{evalrobust2023}, Page Rank (PR) \citep{evalrobust2023}, and Personalized Page Rank (PPR) \citep{shiftrobust2021, evalrobust2023}. Because applying random noise to node features cannot necessarily represent distribution shifts, we use structural properties in our work. For that, we choose PR for main experiments as a popularity-based property since it implies challenging distribution shift \citep{shiftrobust2021, evalrobust2023}. We also study the clustering coefficient as a density-based property and provide results.

\textbf{Graph-level. } Apart from datasets with inherently shifted distributions \citep{moleculenet2017, gnnexplainer2019, ogb2020, acloserlook2021, dir2022, good2022}, number of nodes \citep{unleash2023, oodgnn2023}, average node degrees \citep{invariantrep2022} and other graph properties can be utilized for introducing distribution shift for graph-level tasks. However, for main experiments, we choose edge homophily ratio \citep{beyondhomophily2020} for generating graph datasets with distribution shift because \gnns are intrinsically affected by this property for information aggregation \citep{mpnn2017, beyondhomophily2020, geomgcn2020, linkx2021}, and also our experiments on common graph classification datasets show higher variance of homophily ratio among graphs of these datasets. Additionally, we evaluate our method against graph density as another distribution shift as well. 

\begin{table*}[t]
    \caption{Statistic of the datasets used for experiments.}    \small
    \centering
       \resizebox{\textwidth}{!}{ \begin{tabular}{L{0.6in}*{2}{C{0.4in}}*{5}{C{0.6in}}*{2}{C{0.65in}}}
            \toprule
            \textbf{Dataset} & \textbf{\#Classes} & \textbf{\#Graphs} & \textbf{\#Nodes} & \textbf{\#Edges} & \textbf{\#Features} & \textbf{Avg. \#Nodes} & \textbf{Avg. \#Edges} & \textbf{Continuous Feature} & \textbf{Discrete Feature} \\
            \midrule
            ENZYMES & 6 & 600 & - & - & 21 & 32.63 & 62.14 & \ding{51} & \ding{51} \\
            DHFR & 2 & 756 & - & - & 56 & 42.43 & 44.54 & \ding{51} & \ding{51} \\
            PROTEINS & 2 & 1113 & - & - & 4 & 39.06 & 72.82 & \ding{55} & \ding{51} \\
            BACE & 2 & 1513 & - & - & 9 & 34.1 & 73.7 & \ding{55} & \ding{51} \\
            BBBP & 2 & 2050 & - & - & 9 & 23.9 & 51.6 & \ding{55} & \ding{51} \\
            Cornell & 5 & 1 & 183 & 298 & 1703 & - & - & \ding{55} & \ding{51} \\
            Texas & 5 & 1 & 183 & 325 & 1703 & - & - & \ding{55} & \ding{51} \\
            Wisconsin & 5 & 1 & 251 & 515 & 1703 & - & - & \ding{55} & \ding{51} \\
            Cora & 7 & 1 & 2708 & 10556 & 1433 & - & - & \ding{55} & \ding{51} \\
            CiteSeer & 6 & 1 & 3327 & 9104 & 3703 & - & - & \ding{55} & \ding{51} \\
            PubMed & 3 & 1 & 19717 & 88648 & 500 & - & - & \ding{55} & \ding{51} \\
            Flickr & 7 & 1 & 89250 & 899756 & 500 & - & - & \ding{55} & \ding{51} \\
            \bottomrule
        \end{tabular}}
    \label{tab:datasets}
    % 	\vskip -0.1in
    % \vspace{-3mm}
\end{table*}

\subsubsection{Experimental Details}\label{apdx:exp_details}

\textbf{Task Unification. }A common practice for prompt tuning and in-context learning methods in graph domains is unification of different tasks \citep{gppt2022, allinone2023, prodigy2023}. However, defining a unified task in the graph domain is more challenging than in text (e.g., using LLMs), for two main reasons. First, proposing a unified task requires an enormous amount of data, labeled or unlabeled, while, in contrast to text gathering, these large graph datasets are not feasible. Second, there is a wide variety of downstream graph tasks. Nonetheless, we unify the tasks to graph classification \citep{allinone2023}, considering the message-passing intrinsic of \gnn \citep{mpnn2017}. 

To reduce node classification to graph classification, we select the induced subgraph of each ego node within its $k$-hop neighborhood and assign the label of the ego node to this subgraph. One can also reduce edge-level tasks by selecting the $k$-hop neighborhoods around the nodes lying on the endpoints of each edge \citep{allinone2023}.

\paragraph{Implementation Details.}
We have implemented our experiments in Pytorch \citep{pytorch2017} and used a single GPU core NVIDIA GeForce RTX 3090. To make results reliable, we run each experiment with 10 different random initializations of seeds before dataset creation and with 5 trials of model parameter initialization for every seed, which sums up to a total of 50 runs for every experiment. 

We split all datasets based on properties to make a distribution shift to a 50\% source dataset and a 50\% target dataset. Since we split source and target datasets in the beginning in favor of making distribution shifts, we randomly make our own train, validation, and test splits for every trial, although the node classification datasets have original splits. Therefore, the train, validation, and test split is set to 0.6, 0.1, 0.3 for graph classification datasets and to 0.3, 0.1, 0.6 for node classification datasets---to reflect more on their original splits as they have many more test examples compared to train ones. Besides, for node classification, we find the induced graphs within nodes selected from datasets after the source-target split.

We tune hyper-parameters based on the average F1 score on validation sets as follows. For we select learning rate from $\{0.01, 0.001\}$, batch size from $\{16, 32, 64\}$, number of epochs from $\{30, 50, 60\}$, loss function weights for domain adaptation and diversity ($\lambda_1, \lambda_2$) from $\{0.25, 0.5, 0.75, 1.0, 1.25, 1.5\}$, the $L_2$ regularization factor $(\lambda_3)$ from $\{0.1, 0.2\}$, the augmentation probability $p_u$ from $\{0.1, 0.2, 0.3, 0.4, 0.5\}$ and for $p_w$ from $\{0.05, 0.1, 0.2\}$, the number of trainable prompting parameter vectors $n_p$ from $\{10, 20, 30, 50, \expect_{N_G}\}$ where $\expect_{N_G}$ is the average number of nodes in graphs for the graph datasets. For the certainty threshold $\tau$, we either chose a fixed threshold following FixMatch \citep{fixmatch2020} or a dynamic class-wise threshold following FlexMatch \citep{flexmatch2021}, then we select the threshold from $\{0.1, 0.3, 0.5, 0.7\}$. The final selection of all hyper-parameters for the GCN as base \gnn and for main distribution shifts (edge homophily and PR) is provided in the codes provided by the link before.

\subsection{Additional Experimental Results}\label{apdx:results}

\subsubsection{Comparison with More baselines}

To validate whether \namemodel with no label constraint can offer practical advantages over few-shot methods, in this section, we evaluate our method against more recent state-of-the-art models, namely DAGPrompt, PRONOG, and GCOPE. As evidenced in Tables \ref{tab:few_shot_comparison_graph} and \ref{tab:few_shot_comparison_node}, \namemodel generally outperforms the few-shot baselines in new experiments. 

Specifically, \namemodel consistently surpasses all baselines on graph classification and achieves competitive (first or second best) results on node classification. The robustness of zero-label acts as a powerful regularizer, forcing distribution alignment for positive transfer, which contrasts sharply with few-shot methods that often suffer from catastrophic negative transfer and overfitting to biased samples under distribution shifts. In addition, \namemodel has notably less complex (lower number of trainable parameters) than the baselines. Therefore, heavier models like DAGPrompt (which has separate trainable low-rank matrices for each input and GNN encoder layer) fail catastrophically on smaller graph classification datasets---despite having superior performance on large and heterophilic node classification tasks---and undesirably show negative improvement as it requires a larger number of samples.

\begin{table*}[!t]
    \vspace{-2mm}
    \caption{Comparison against methods specialized for few-shot learning on graph classification under distribution shift. \namemodel (with 0\% labels) consistently outperforms few-shot competitors (with 25\% labels), which often suffer catastrophic negative transfer.}
    \centering
    \resizebox{0.8\textwidth}{!}{
        \begin{tabular}{L{1.2in}C{0.5in}*{5}{C{0.7in}}}
            \toprule
            \textbf{Method} & \textbf{\%Label} & \textbf{ENZYMES} & \textbf{PROTEINS} & \textbf{DHFR} & \textbf{BBBP} & \textbf{BACE} \\
            \midrule
            DAGPrompt & \multirow{3}{*}{25} & -47.6 & -36.4 & -9.3 & -3.1 & -7.0 \\
            PRONOG & & \underline{-26.8} & -26.6 & \underline{-2.1} & \underline{0.0} & \underline{9.8} \\
            GCOPE & & -28.5 & \underline{-15.8} & -6.2 & -3.5 & -15.2 \\
            \cmidrule(lr){1-7}
            \textbf{\namemodel (ours)} & \textbf{0} & \textbf{2.9} & \textbf{8.1} & \textbf{2.0} & \textbf{1.1} & \textbf{10.4} \\
            \bottomrule
        \end{tabular}}
    \label{tab:few_shot_comparison_graph}
\end{table*}

\begin{table*}[!t]
    \vspace{-2mm}
    \caption{Comparison against methods specialized for few-shot learning on node classification under distribution shift. Few-shot methods achieve the highest gain on most datasets, but \namemodel is the best overall performer without relying on any labeled data.}
    \centering
    \resizebox{0.9\textwidth}{!}{
        \begin{tabular}{L{1.4in}C{0.5in}*{7}{C{0.5in}}}
            \toprule
            \textbf{Method} & \textbf{\%Label} & \textbf{Cora} & \textbf{CiteSeer} & \textbf{PubMed} & \textbf{Flickr} & \textbf{Cornell} & \textbf{Texas} & \textbf{Wisconsin} \\
            \midrule
            DAGPrompt & \multirow{3}{*}{25} & \underline{3.2} & \underline{1.0} & \textbf{16.3} & \textbf{16.4} & \textbf{25.1} & \textbf{37.7} & \textbf{15.1} \\
            PRONOG & & -7.1 & -3.9 & -5.3 & -26.7 & -3.7 & \underline{34.3} & 4.4 \\
            GCOPE & & 0.4 & -4.5 & -2.4 & -30.9 & -14.1 & -6.8 & -3.9 \\
            \cmidrule(lr){1-9}
            \textbf{\namemodel (ours)} & \textbf{0} & \textbf{6.5} & \textbf{3.6} & \underline{7.2} & \underline{6.1} & \underline{21.5} & 13.6 & \underline{11.1} \\
            \bottomrule
        \end{tabular}}
    \label{tab:few_shot_comparison_node}
\end{table*}

\subsubsection{Effectiveness across Different Distribution Shifts.} \label{apdx:exp_dist_shift}
The problem setup \problemname validates how well prompting methods can relieve the performance drop of a \gnn facing distribution shift. Therefore, we aim to show the generalization of \namemodel across different kinds of distribution shifts. Previously, we evaluated our method on distribution shift based on edge homophily ratio and Page Rank (PR). Here we add shifts based on graph density for graph classification and clustering coefficient for node classification.

\textit{Graph Density Distribution Shift. } Table \ref{tab:graphgcndensity} illustrates \namemodel generally attains the best results on the graph datasets with consistent positive performance gain, while the competitors mostly have negative performance gain on DHFR and PROTEINS in all cases even when they take advantage of 50\% of labeled data of the target distribution. Besides, we have the second-best performance on ENZYMES after Fine-tuning in 50\% label setting while beating this baseline in 25\% label setting.

\textit{Node Clustering Coefficient Density Distribution Shift. } The results in Table \ref{tab:nodegcncc} reflect on the superiority of \namemodel over all competitors on CiteSeer and PubMed. However, GPF-Plus outperforms our method on Cora when it has access to 50\% of labeled samples while is beaten by \namemodel when 25\% of labels are available. Notably, we excel over all of the baselines on PubMed with a high margin and with no labels. 

\begin{table*}[!t]
    \vspace{-2mm}
    \caption{Graph classification results for GCN as the base model on target datasets having distribution shift based on graph density. Overall, the best results are attained by \namemodel without any labels of the target dataset compared to the case where baselines have access to 50\% of labels.}
    \centering
    \scalebox{0.8}{
        \begin{tabular}{L{1.2in}*{7}{C{0.4in}}}
            \toprule
            \multirow{2}{6em}{\textbf{Method}} & \multirow{2}{*}{\hspace{-2mm}\textbf{\%Label}} &  \multicolumn{2}{c}{\textbf{ENZYMES}}  &  \multicolumn{2}{c}{\textbf{PROTEINS}}  &  \multicolumn{2}{c}{\textbf{DHFR}} \\
            \cmidrule(lr){3-4} \cmidrule(lr){5-6} \cmidrule(lr){7-8}
            & & \textbf{F1} & \textbf{IMP} & \textbf{F1} & \textbf{IMP} & \textbf{F1} & \textbf{IMP} \\
            \midrule
            BaseModel & 0 & 39.1\std{4.1} & 0.0 & 56.1\std{4.0} & 0.0 & 76.3\std{3.8} & 0.0 \\
            \midrule
            Fine-Tuning & \multirow{4}{*}{50} & \textbf{40.7}\std{0.7} & \textbf{4.1} & 39.5\std{1.6} & -29.6 & 76.6\std{0.2} & 0.4 \\
            GraphPrompt & & 33.6\std{1.4} & -14.1 & 55.7\std{0.9} & -0.7 & 58.3\std{3.1} & -23.6 \\
            All-In-One & & 30.0\std{3.2} & -23.3 & 37.3\std{9.5} & -33.5 & \underline{77.2}\std{0.7} & \underline{1.2} \\
            GPF-Plus & & 39.2\std{1.4} & 0.3 & \underline{57.6}\std{1.2} & \underline{2.7} & 74.2\std{0.9} & -2.8 \\
            \midrule
            Fine-Tuning & \multirow{4}{*}{25} & 38.0\std{1.1} & -2.8 & 37.0\std{3.6} & -34.0 & 73.6\std{0.5} & -3.5 \\
            GraphPrompt & & 29.9\std{1.4} & -23.5 & 53.3\std{1.8} & -5.0 & 57.8\std{3.1} & -24.2 \\
            All-In-One & & 27.3\std{3.8} & -30.2 & 34.2\std{10.7} & -39.0 & 77.1\std{0.7} & 1.0 \\
            GPF-Plus & & 39.1\std{1.3} & 0.0 & 57.2\std{1.7} & 2.0 & 74.2\std{0.9} & -2.8 \\
            \midrule
            \namemodel (ours) & \textbf{0} & \underline{40.0}\std{1.0} & \underline{2.3} & \textbf{58.3}\std{0.8} & \textbf{3.9} & \textbf{78.0}\std{0.8} & \textbf{2.2} \\
            \bottomrule
        \end{tabular}}
    \label{tab:graphgcndensity}
    % 	\vskip -0.1in
\end{table*}

\begin{table*}[!t]
    \vspace{-2mm}
    \caption{Node classification results for GCN as the base model on target datasets having distribution shift based on node clustering coefficient. When 50\% of target dataset labels are visible to baseline, the unsupervised \namemodel achieves the second-best results while it performs the best overall datasets in cases where 25\% of labels are available for competitors.}
    \centering
    \scalebox{0.8}{
        \begin{tabular}{L{1.2in}*{7}{C{0.4in}}}
            \toprule
            \multirow{2}{6em}{\textbf{Method}} & \multirow{2}{*}{\hspace{-2mm}\textbf{\%Label}} &  \multicolumn{2}{c}{\textbf{Cora}}  &  \multicolumn{2}{c}{\textbf{CiteSeer}}  &  \multicolumn{2}{c}{\textbf{PubMed}} \\
            \cmidrule(lr){3-4} \cmidrule(lr){5-6} \cmidrule(lr){7-8}
            & & \textbf{F1} & \textbf{IMP} & \textbf{F1} & \textbf{IMP} & \textbf{F1} & \textbf{IMP} \\
            \midrule
            BaseModel & 0 & 59.0\std{3.1} & 0.0 & 44.1\std{1.6} & \underline{0.0} & 60.0\std{0.8} & 0.0 \\
            \midrule
            Fine-Tuning & \multirow{5}{*}{50} & 60.2\std{0.9} & 2.0 & 38.7\std{0.5} & -12.2 & 53.5\std{2.3} & -10.8 \\
            GraphPrompt & & 59.9\std{0.3} & 1.5 & \underline{44.8}\std{0.3} & -1.6 & \underline{61.3}\std{0.1} & \underline{2.2} \\
            GPPT & & 47.9\std{6.8} & -18.8 & 39.0\std{2.2} & -11.6 & 55.1\std{3.7} & -8.2 \\
            All-In-One & & 53.7\std{1.2} & -9.0 & 39.4\std{1.0} & -10.7 & 47.1\std{0.7} & -21.5 \\
            GPF-Plus & & \textbf{61.4}\std{0.6} & \textbf{4.1} & 41.7\std{0.6} & -5.6 & \underline{61.3}\std{1.0} & \underline{2.2} \\
            \midrule
            Fine-Tuning & \multirow{5}{*}{25} & 56.5\std{0.6} & -4.2 & 39.9\std{0.4} & -9.5 & 48.7\std{3.9} & -18.8 \\
            GraphPrompt & & 58.0\std{0.4} & -1.7 & 43.7\std{0.3} & -0.9 & \underline{61.3}\std{0.1} & -2.2 \\
            GPPT & & 46.4\std{4.6} & -21.4 & 38.2\std{2.9} & -13.4 & 54.5\std{3.6} & -9.2 \\
            All-In-One & & 53.7\std{1.0} & -9.0 & 38.3\std{0.9} & -13.2 & 48.9\std{0.8} & -18.5 \\
            GPF-Plus & & 59.8\std{0.5} & 1.4 & 40.3\std{0.6} & -8.6 & 60.9\std{0.8} & 1.5 \\
            \midrule
            \namemodel (ours) & \textbf{0} & \underline{60.5}\std{0.3} & \underline{2.5} & \textbf{45.1}\std{0.4} & \textbf{2.3} & \textbf{64.7}\std{0.3} & \textbf{7.8} \\
            \bottomrule
        \end{tabular}}
    \label{tab:nodegcncc}
\end{table*}

% \iffalse
% \subsubsection{Few-shot Learning}

% \namemodel was originally proposed as an unsupervised \gnn prompting method, with previous studies demonstrating its effectiveness. However, it also has the potential to efficiently utilize labeled data when available. This experiment evaluates \namemodel’s performance in a few-shot setting. Table \ref{tab:few_shot} presents the results using GCN as the base \gnn under homophily and PR distribution shifts.

% The key finding from Table \ref{tab:few_shot} is the significant improvement in \namemodel’s performance when labels are provided. Specifically, for node classification datasets, the IMP value in the 50\% labels setting is, on average, twice that of the unsupervised case (0\% labels), while PROTEINS and DHFR show smaller improvements. Conversely, performance on ENZYMES remains comparable or better in the absence of labels, likely due to a significant covariate shift between source and target datasets, which makes learning from highly heterophilous data challenging.

% Overall, the main takeaway from this experiment comes from comparing Table \ref{tab:few_shot} with Tables \ref{tab:graphgcn} and \ref{tab:nodegcn}. This comparison reveals that when 25\% or 50\% of labeled data is fairly provided to all methods, \namemodel outperforms every baseline across all datasets except for DHFR. Notably, previous results also demonstrated \namemodel’s superior performance in most cases, even under a fully unsupervised setting.
% \fi

\subsubsection{Generalization to Advanced GNN Backbones}\label{apdx:advanced_gnn}

A core principle of a prompting method is that it should be agnostic to the specific GNN backbone, demonstrating effectiveness across a range of architectures. To validate this, we extend our experiments beyond the foundational GCN and GAT models by incorporating two powerful, recent \gnns. We specifically chose: 1) GATv2 \citep{gatvII2022}, a model that uses dynamic attention to better handle graphs with varying levels of homophily and heterophily. 2) GraphGPS \citep{graphgps2022}, a graph transformer that uses a global attention mechanism to address common \gnn challenges like oversmoothing and oversquashing.

The results of these new experiments are presented in Table \ref{tab:graphgps_gatv2} for graph and node classification tasks, respectively. The findings clearly demonstrate that our method, \namemodel, consistently outperforms all baselines when applied to these advanced GNN backbones. This is particularly noteworthy as the baselines retain the significant advantage of access to labeled data from the target domain, whereas our method operates in a completely unsupervised manner. These results confirm that the effectiveness of our prompting framework is independent of a specific \gnn architecture and that it generalizes robustly to more powerful and modern backbones.

\begin{table*}[!t]
    \vspace{-2mm}
    \caption{Graph classification results for GraphGPS and GATv2 as the base model on target datasets having distribution shift based on node clustering coefficient. \namemodel achieves almost the best across all datasets without labels, while 25\% of labels are available for competitors.}
    \centering
    \resizebox{\textwidth}{!}{
        \begin{tabular}{L{0.7in}L{1.2in}C{0.35in}*{10}{C{0.3in}}}
            \toprule
            \multirow{2}{6em}{\textbf{BaseGNN}} & \multirow{2}{*}{\hspace{-2mm}\textbf{Method}} & \multirow{2}{*}{\hspace{-2mm}\textbf{\%Label}} & \multicolumn{2}{c}{\textbf{BBBP}}  &  \multicolumn{2}{c}{\textbf{BACE}}  &  \multicolumn{2}{c}{\textbf{Cornell}}  &  \multicolumn{2}{c}{\textbf{Texas}}  &  \multicolumn{2}{c}{\textbf{Wisconsin}} \\
            \cmidrule(lr){4-5} \cmidrule(lr){6-7} \cmidrule(lr){8-9} \cmidrule(lr){10-11} \cmidrule(lr){12-13}
            & & & \textbf{F1} & \textbf{IMP} & \textbf{F1} & \textbf{IMP} & \textbf{F1} & \textbf{IMP} & \textbf{F1} & \textbf{IMP} & \textbf{F1} & \textbf{IMP} \\
            \midrule
            \multirow{6}{*}{GraphGPS} & BaseModel & 0 & 87.0 & - & 73.0 & - & 22.8 & - & 24.4 & - & 20.8 & - \\
            \cmidrule(lr){2-13}
            & Fine-Tuning & \multirow{5}{*}{25} & \underline{87.8} & \underline{0.8} & 70.0 & -4.1 & \underline{24.8} & \underline{8.8} & \underline{27.0} & \underline{10.7} & 19.6 & -5.8 \\
            & GraphPrompt+ & & 77.8 & -10.6 & 51.5 & -29.5 & 12.7 & -44.3 & 22.8 & -6.6 & 5.9 & -71.6 \\
            & All-In-One & & 72.1 & -17.1 & 39.5 & -46.9 & 19.6 & -14.0 & 0.3 & -98.8 & 22.3 & 7.2 \\
            & GPF-Plus & & 87.7 & \underline{0.8} & \underline{73.6} & \underline{0.8} & 22.9 & 0.4 & 26.9 & 10.2 & \underline{22.5} & \underline{8.2} \\
            \cmidrule(lr){2-13}
            & \namemodel (ours) & \textbf{0} & \textbf{88.6} & \textbf{1.8} & \textbf{75.0} & \textbf{2.7} & \textbf{26.6} & \textbf{16.7} & \textbf{28.2} & \textbf{15.6} & \textbf{23.1} & \textbf{11.1} \\
            \midrule
            \multirow{6}{*}{GATv2} & BaseModel & 0 & 88.8 & - & 63.8 & - & 16.0 & - & 22.0 & - & 24.7 & - \\
            \cmidrule(lr){2-13}
            & Fine-Tuning & \multirow{5}{*}{25} & \underline{89.4} & \underline{0.7} & 67.8 & 6.3 & 17.8 & 11.3 & \underline{22.3} & \underline{1.4} & 23.5 & -4.9 \\
            & GraphPrompt+ & & 83.6 & -5.9 & 64.3 & 0.8 & 10.3 & -35.6 & 2.0 & -90.9 & 8.2 & -66.8 \\
            & All-In-One & & 88.5 & -0.3 & 47.4 & -25.7 & \underline{18.6} & \underline{16.3} & 16.6 & -24.5 & 16.9 & -31.6 \\
            & GPF-Plus & & 89.3 & 0.6 & \textbf{68.6} & \textbf{7.5} & 18.2 & 13.8 & 21.8 & -0.9 & \underline{25.4} & \underline{2.8} \\
            \cmidrule(lr){2-13}
            & \namemodel (ours) & \textbf{0} & \textbf{89.6} & \textbf{0.9} & \underline{67.9} & \underline{6.4} & \textbf{20.2} & \textbf{26.3} & \textbf{23.0} & \textbf{4.5} & \textbf{26.2} & \textbf{6.1} \\
            \bottomrule
        \end{tabular}}
    \label{tab:graphgps_gatv2}
\end{table*}

\subsubsection{Comparison with Source-Free Domain Adaptation Methods}\label{apdx:sfda_comparison}

% While our problem setup is distinct from SFDA methods, the shared goal of adapting a model without source data motivates a direct empirical comparison. 
Since \problemname is a restrictive prompt-based setting within the broader SFDA landscape, the shared goal of adapting a model without source data motivates a direct empirical comparison.
To this end, we evaluate \namemodel against two recent, state-of-the-art graph SFDA methods: SOGA\citep{soga2024} and GraphCTA\citep{graphcta2024}. Since these methods are designed for node classification, we conduct the evaluation on our node classification datasets.

To create a fair comparison within our prompting-focused problem setup, we adapt these baselines. SFDA methods typically fine-tune the entire model; instead, we align them with the ``lightweight fine-tuning'' paradigm common to other baselines by freezing the source-trained GNN's encoder and only allowing the decoder (prediction head) to be trained on the target data. This contrasts with our method, \namemodel, where both the encoder and decoder remain fully frozen.

The results of this comparison are presented in Table \ref{tab:sfda_eval}. \namemodel consistently and significantly outperforms both adapted SFDA methods across all datasets. This outcome is powerful, as it demonstrates that our parameter-efficient approach of training only a prompt is more effective for adaptation than the more common fine-tuning strategy for the prediction head. In addition, our method is significantly lighter since we only train a small set of trainable prompting vectors, but these methods are originally proposed to fine-tune all parameters of the pretrained model. 

\begin{table*}[!t]
    \footnotesize
    \caption{Comparison again unsupervised SFDA methods. \namemodel outperforms both competitors while it has significantly less number of trainable parameters.}
    \centering
    \scalebox{0.8}{
        \begin{tabular}{L{1.0in}C{0.4in}*{4}{C{0.3in}C{0.3in}}}
            \toprule
            \multirow{2}{*}{\textbf{Method}} & \multirow{2}{*}{\textbf{Label}} & \multicolumn{2}{c}{\textbf{Cora}} & \multicolumn{2}{c}{\textbf{CiteSeer}} & \multicolumn{2}{c}{\textbf{PubMed}} & \multicolumn{2}{c}{\textbf{Flickr}} \\
            \cmidrule(lr){3-4} \cmidrule(lr){5-6} \cmidrule(lr){7-8} \cmidrule(lr){9-10}
            & & \textbf{F1} & \textbf{IMP} & \textbf{F1} & \textbf{IMP} & \textbf{F1} & \textbf{IMP} & \textbf{F1} & \textbf{IMP} \\
            \midrule
            Base Model & - & \underline{53.8} & - & \underline{44.1} & - & \underline{57.1} & - & \underline{16.5} & - \\
            \midrule
            SOGA & \multirow{2}{*}{0\%} & 53.1 & -1.3 & 43.7 & -0.9 & 54.9 & -3.9 & 14.5 & -12.1 \\
            GraphCTA & & 49.2 & -8.6 & 38.0 & -13.8 & 12.1 & -78.8 & OOM & OOM \\
            \midrule
            \namemodel (ours) & 0\% & \textbf{57.3} & \textbf{6.5} & \textbf{45.7} & \textbf{3.6} & \textbf{61.2} & \textbf{7.2} & \textbf{17.5} & \textbf{6.1} \\
            \bottomrule
        \end{tabular}
    }
    \label{tab:sfda_eval}
\end{table*}

\subsection{Comparison with Decoder Fine-tuned Prompting Baselines}
\label{app:finetuned_decoder_baselines}

In the main experiments, we restrict all prompting baselines to the frozen-\gnn setting required by \problemname. This allows us to isolate the effect of the prompting function without conflating it with fine-tuning the source-trained encoder or prediction head. However, several supervised GNN prompting methods were originally designed to fine-tune a task-specific decoder. To provide a more comprehensive comparison, we further evaluate the supervised baselines in this relaxed setting, where their decoders are fine-tuned using 25\% labeled target data.

Table~\ref{tab:decoder_finetune_baselines} reports the performance improvement over the frozen BaseModel. Even with access to labeled target data and decoder fine-tuning, the supervised baselines often suffer from negative transfer under distribution shift. In contrast, \namemodel keeps both the encoder and decoder frozen, uses no labels, and still achieves stable positive improvements in most cases. These results further support that the gains of \namemodel come from effective unsupervised prompting rather than from additional fine-tuning.

\begin{table*}[!t]
    \vspace{-2mm}
    \small
    \caption{Comparison of IMP\% against supervised prompting baselines with fine-tuned decoders using 25\% labeled target data. GPPT is not applicable (NA) to graph classification. \namemodel generally outperforms the baselines without labels or decoder fine-tuning.}
    \centering
    \scalebox{0.8}{
        \begin{tabular}{L{0.9in}*{9}{C{0.6in}}}
            \toprule
            \textbf{Method} & \textbf{Texas} & \textbf{Cornell} & \textbf{Wisconsin} & \textbf{Cora} & \textbf{CiteSeer} & \textbf{PubMed} & \textbf{ENZYMES} & \textbf{PROTEINS} & \textbf{DHFR} \\
            \midrule
            GPPT & 9.2\std{32.8} & -20.9\std{15.5} & -8.1\std{14.7} & -13.6\std{5.0} & -12.6\std{3.9} & -9.8\std{2.4} & NA & NA & NA \\
            GraphPrompt & 23.1\std{39.3} & -5.2\std{32.7} & -11.9\std{21.0} & -0.8\std{6.8} & -3.5\std{4.0} & -0.5\std{1.9} & -21.2\std{9.2} & -1.9\std{21.7} & -6.1\std{9.5} \\
            GraphPrompt+ & 27.5\std{35.2} & 12.0\std{34.6} & -0.6\std{15.0} & -9.0\std{8.1} & -9.5\std{3.6} & 8.7\std{2.1} & -50.0\std{7.0} & -13.7\std{23.9} & -14.7\std{6.8} \\
            All-In-One & -11.8\std{20.5} & -34.5\std{14.3} & -36.5\std{12.8} & -10.6\std{18.4} & -19.8\std{9.5} & -10.0\std{11.9} & -24.2\std{9.1} & -25.6\std{26.7} & -0.4\std{5.1} \\
            GPF-Plus & 24.9\std{32.7} & 5.8\std{30.4} & 4.7\std{25.3} & -1.7\std{5.3} & -3.0\std{3.9} & 10.2\std{2.3} & -17.3\std{12.0} & -17.7\std{20.3} & 2.2\std{4.9} \\
            \midrule
            \namemodel & 13.6\std{23.5} & 23.6\std{16.2} & 12.7\std{10.5} & 2.5\std{4.8} & 1.3\std{3.5} & 6.1\std{2.6} & 0.8\std{9.6} & 6.9\std{16.2} & 1.4\std{5.5} \\
            \bottomrule
        \end{tabular}}
    \label{tab:decoder_finetune_baselines}
\end{table*}

\begin{figure}[!b]
    \centering
    \includegraphics[width=0.95\textwidth]{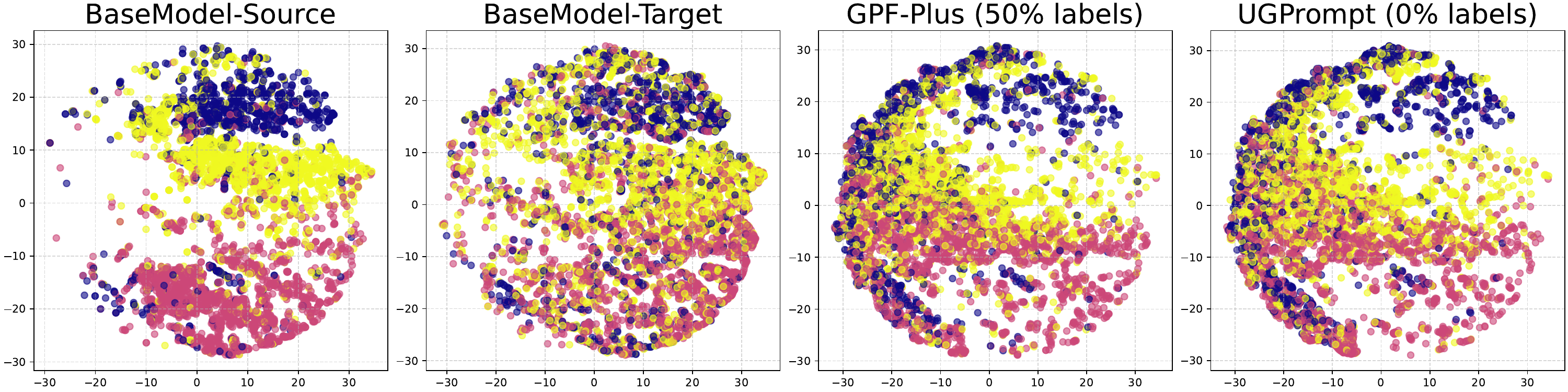}
    \caption{Distribution of embeddings generated by the base \gnn encoder on the PubMed dataset under PR distribution shift with GCN as the base \gnn. Embeddings for BaseModel on source and target test data, without prompting, and for \namemodel and GPF-Plus when target test graphs are prompted, are presented. This highlights \namemodel’s ability to mitigate distribution shifts without labeled data, producing well-separated representations that are similar to the source distribution.}
    \label{fig:dists}
\end{figure}

\subsubsection{Embedding distribution analysis under distribution shift} \label{apdx:shift_visualization}
Figure \ref{fig:dists} illustrates the distribution of embeddings produced by the base \gnn's encoder for the PubMed dataset. The setting is for PR distribution shift with GCN as the base \gnn. It compares the embeddings of the BaseModels on source (BaseModel-source) and target (BaseModel-Target) test data, with graphs not prompted in either case. Additionally, it shows the embeddings for \namemodel and GPF-Plus when prompted with graphs from the target test data, while \namemodel is fully unsupervised and GPF-Plus consumes 50\% of the labeled data. 

This figure provides empirical evidence supporting our claims. First, \namemodel, which employs consistency regularization without labeled data, achieves equal or better performance in mitigating distribution shift compared to GPF-Plus, although the latter has access to labeled data. Second, when using confident pseudo-labels and keeping the \gnn's parameters fixed after source-training, the prompting function aligns with the knowledge learned from the source data. This approach preserves samples within densely homophilous regions while pushing uncertain samples (those in overlapping regions) away, replicating the source data distribution as seen in BaseModel-Source. As a result, the method produces well-separated representations, enabling the projection head to effectively discriminate between classes using the same source-trained weights.

\subsubsection{Effect of regularization methods}\label{apdx:regeffect}

We study the effect of domain adaptation and diversity regularization methods introduced in Section \ref{sec:prompt_regularization}. Minimizing the entropy of the score's expected value (as a diversity regularization term) has also been used in prior representation learning studies \citep{discinfomax2017, doweuda2020}. This regularization term relieves the harmful effect of class imbalance. Meanwhile, the domain adaptation regularization term is more beneficial for smaller datasets, since inferring their distribution from a small number of samples is difficult, increasing the likelihood of generating OOD prompted graphs. To showcase the effect of these objectives, we introduce two common methods of measuring class imbalance: the Imbalanced Ratio (IR) and Normalized Entropy (NE). A higher IR and a lower NE show more class imbalance in the datasets. Denoting the frequency of each class of a dataset as $f_i$, while we have $C$ classes, IR and NE are defined as below:

\begin{align}
    IR = \frac{f_{max}}{f_{min}}; \qquad f_{min}=\underset{i}{\argmin}\ f_i, \quad f_{max}=\underset{i}{\argmax}\ f_i\\\
    NE = \frac{H}{log(C)}; \qquad H=-\sum_{i=1}^{C}p_ilog(p_i), \quad p_i=\frac{f_i}{\sum_{i=1}^Cf_i}
\end{align}

Tables \ref{tab:ds_reg_stats} shows the total number of samples in the target datasets (50\% of the actual dataset sizes because we split the datasets into half source and half target), the statistics of their corresponding test sets, and the IMP\% of \namemodel on these datasets (we fix GCN as the base GNN).

\begin{table*}[t]
    \caption{Statistic of the datasets used for experiments.}
    \centering
       \scalebox{0.8}{ \begin{tabular}{L{0.6in}*{3}{C{0.5in}}*{2}{C{1.2in}}}
            \toprule
            \textbf{Dataset} & \textbf{\#Samples} & \textbf{NE} & \textbf{IR} & \textbf{IMP\% of Diversity Reg.} & \textbf{IMP\% of Domain Reg.} \\
            \midrule
            ENZYMES & 300 & 0.997 & 1.26 & 1.5 & 1.2 \\
            PROTEINS & 556 & 0.949 & 1.72 & 6.6 & 4.3 \\
            DHFR & 378 & 0.980 & 1.39 & 0.0 & 0.7 \\
            Cora & 1354 & 0.941 & 4.72 & 4.0 & 0.7 \\
            CiteSeer & 1663 & 0.965 & 1.94 & 1.3 & 0.0 \\
            PubMed & 9858 & 0.966 & 1.91 & 0.2 & 0.0 \\
            \bottomrule
        \end{tabular}}
    \label{tab:ds_reg_stats}
    % 	\vskip -0.1in
    % \vspace{-3mm}
\end{table*}

When looking at both graph and node classification results, the general trends support our claims that the regularization methods have a positive impact. The first main trend observed in each task group is that diversity regularization is more effective as dataset imbalance increases; specifically, on PROTEINS and Cora, which are the most imbalanced datasets for graph and node classification respectively, diversity regularization shows the highest IMP\%. Furthermore, when considering dataset size, domain regularization generally demonstrates its greatest effect on smaller datasets. For example, Cora, the node classification dataset with the fewest samples, benefits the most from domain regularization. An exception regarding domain regularization's effect might be seen when comparing PROTEINS with DHFR and ENZYMES; this could be because PROTEINS has discrete features, whereas ENZYMES and DHFR also include continuous features, potentially reducing the impact of domain regularization.

\begin{table*}[!t]
    \caption{Ablation study on the contribution of each objective function ($\lambda_1$ and $\lambda_2$ are respectively the weights of diversity and domain adaptation objectives as in Equation \ref{eq:totalloss}) while using GCN as the base model. Both objectives show positive effect on \namemodel's performance.}
    \vspace{-2mm}
    \centering
    \scalebox{0.8}{
        \begin{tabular}{L{0.6in}C{1.2in}*{7}{C{0.4in}}}
            \toprule
            \textbf{Dataset} & Base ($\lambda_1$, $\lambda_2$=0.0) & \multicolumn{7}{c}{} \\
            \midrule
            \multirow{4}{*}{\textbf{Cora}} & \multirow{4}{*}{\textbf{F1}=\textbf{56.3}} & $\lambda_1$ & 0.25 & 0.5 & 0.75 & 1.0 & 1.25 & 1.5 \\
            & & \textbf{IMP\%} & 1.4 & 1.6 & 2.0 & 2.0 & 2.0 & 2.0 \\
            \cmidrule(lr){3-9}
            & & $\lambda_2$ & 0.25 & 0.5 & 0.75 & 1.0 & 1.25 & 1.5 \\
            & & \textbf{IMP\%} & 1.2 & 1.1 & 1.1 & 0.9 & 0.9 & 1.1 \\
            \midrule
            \multirow{4}{*}{\textbf{PROTEINS}} & \multirow{4}{*}{\textbf{F1}=\textbf{51.0}} & $\lambda_1$ & 0.25 & 0.5 & 0.75 & 1.0 & 1.25 & 1.5 \\
            & & \textbf{IMP\%} & 3.1 & 5.9 & 7.1 & 7.5 & 7.5 & 7.5 \\
            \cmidrule(lr){3-9}
            & & $\lambda_2$ & 0.25 & 0.5 & 0.75 & 1.0 & 1.25 & 1.5 \\
            & & \textbf{IMP\%} & 4.1 & 5.1 & 6.3 & 6.3 & 6.5 & 6.9 \\
            \bottomrule
        \end{tabular}}
    \label{tab:lambdaeffect}
\end{table*}

To better show the effects of each regularization method, we also conduct a more comprehensive ablation study on Cora and PROTEINS, which serve as representatives of datasets for node and graph classification. Table \ref{tab:lambdaeffect} shows how increasing the weight of each regularization term, as in Equation \ref{eq:totalloss}, improves the performance on both datasets, especially on PROTEINS.

\subsubsection{Effectiveness under no distribution shift}\label{apdx:no_shift}

A key concern for any adaptation method is whether adapting to a new target distribution causes ``catastrophic forgetting,'' degrading performance on the original source domain. To measure this, we adapted models on the target data (with a distribution shift) and then evaluated their performance on both the original source test set and the target test set. The results are presented in Table \ref{tab:source_target}, with columns marked (S) for source and (T) for target evaluation. The findings show that \namemodel not only adapts effectively to the target distribution but also maintains higher performance on the source data compared to traditional lightweight fine-tuning, demonstrating its robustness against catastrophic forgetting.

To further understand our method's behavior, we also investigated its effect in a no-shift setting, where the model is prompted with a target domain that shares the same distribution as the source domain. As shown in Table \ref{tab:no_shift_eval}, we measured the F1 score improvement (IMP) over the base model in this scenario. The results demonstrate that \namemodel consistently improves performance even without a distribution shift. This benefit is more evident than that of competing baselines, which show smaller gains and occasionally fail to improve performance at all. This indicates that our prompting approach serves as a general performance enhancer, not just a tool for mitigating distribution shifts.

\begin{table*}[!t]
    \footnotesize
    \caption{Performance comparison of different methods across various source (S) and target (T) datasets.}
    \centering
    % \scalebox{0.85}{
    \resizebox{\textwidth}{!}{
        \begin{tabular}{L{0.65in}C{0.3in}*{6}{C{0.25in}C{0.25in}}}
            \toprule
            \multirow{2}{*}{\textbf{Method}} & \multirow{2}{*}{\textbf{\%Label}} & \multicolumn{2}{c}{\textbf{Cora}} & \multicolumn{2}{c}{\textbf{CiteSeer}} & \multicolumn{2}{c}{\textbf{PubMed}} & \multicolumn{2}{c}{\textbf{ENZYMES}} & \multicolumn{2}{c}{\textbf{PROTEINS}} & \multicolumn{2}{c}{\textbf{DHFR}} \\
            \cmidrule(lr){3-4} \cmidrule(lr){5-6} \cmidrule(lr){7-8} \cmidrule(lr){9-10} \cmidrule(lr){11-12} \cmidrule(lr){13-14}
            & & \textbf{F1(S)} & \textbf{F1(T)} & \textbf{F1(S)} & \textbf{F1(T)} & \textbf{F1(S)} & \textbf{F1(T)} & \textbf{F1(S)} & \textbf{F1(T)} & \textbf{F1(S)} & \textbf{F1(T)} & \textbf{F1(S)} & \textbf{F1(T)} \\
            \midrule
            Base Model & - & \underline{75.5} & \underline{53.8} & \underline{63.1} & \underline{44.1} & \textbf{81.9} & \underline{57.1} & \underline{49.5} & \underline{47.7} & \underline{60.5} & \underline{51.8} & 78.2 & 75.5 \\
            \midrule
            Fine-Tuning & 25 & 69.8 & 51.7 & 57.2 & 40.0 & 74.2 & 54.3 & 43.4 & 46.4 & 56.2 & 47.5 & \textbf{79.4} & \underline{76.8} \\
            \midrule
            \namemodel (ours) & \textbf{0} & \textbf{75.7} & \textbf{57.3} & \textbf{63.3} & \textbf{45.7} & \underline{81.0} & \textbf{61.2} & \textbf{49.6} & \textbf{49.1} & \textbf{63.3} & \textbf{56.0} & \underline{78.5} & \textbf{77.0} \\
            \bottomrule
        \end{tabular}
    }
    \label{tab:source_target}
\end{table*}

\begin{table*}[!t]
    \small
    \caption{Performance comparison of \namemodel against the competitor under no distribution shift. An overall assessment across all datasets shows that our unsupervised method performs superiorly.}
    \centering
    \resizebox{\textwidth}{!}{
        \begin{tabular}{L{1.0in}*{6}{C{0.3in}C{0.3in}}C{0.4in}}
            \toprule
            \multirow{2}{*}{\textbf{Method}} & \multirow{2}{*}{\textbf{Label}} & \multicolumn{2}{c}{\textbf{Cora}} & \multicolumn{2}{c}{\textbf{CiteSeer}} & \multicolumn{2}{c}{\textbf{PubMed}} & \multicolumn{2}{c}{\textbf{ENZYMES}} & \multicolumn{2}{c}{\textbf{PROTEINS}} & \multicolumn{2}{c}{\textbf{DHFR}} \\
            \cmidrule(lr){3-4} \cmidrule(lr){5-6} \cmidrule(lr){7-8} \cmidrule(lr){9-10} \cmidrule(lr){11-12} \cmidrule(lr){13-14}
            & & \textbf{F1} & \textbf{IMP} & \textbf{F1} & \textbf{IMP} & \textbf{F1} & \textbf{IMP} & \textbf{F1} & \textbf{IMP} & \textbf{F1} & \textbf{IMP} & \textbf{F1} & \textbf{IMP}\\
            \midrule
            Base Model & & 64.9 & - & 50.5 & - & 73.2 & - & 54.5 & - & \underline{51.8} & - & 75.1 & - \\
            \midrule
            GraphPrompt+ & \multirow{2}{*}{25} & \textbf{72.6} & \textbf{11.9} & 48.8 & -3.4 & \textbf{76.4} & \textbf{4.4} & 57.1 & 4.8 & 35.2 & -32.2 & 55.4 & -26.3 \\
            GPF-Plus & & \underline{66.5} & \underline{2.5} & \underline{51.2} & \underline{1.4} & 73.6 & 0.5 & \underline{57.9} & \underline{6.2} & 49.5 & -4.4 & \underline{76.1} & \underline{1.3} \\
            \midrule
            \namemodel (ours) & \textbf{0} & 65.9 & 1.5 & \textbf{51.4} & \textbf{1.8} & \underline{75.8} & \underline{3.6} & \textbf{58.7} & \textbf{7.7} & \textbf{52.2} & \textbf{0.8} & \textbf{76.2} & \textbf{1.5} \\
            \bottomrule
        \end{tabular}
    }
    \label{tab:no_shift_eval}
\end{table*}

\subsubsection{Effect of the number of trainable prompting vectors}
As the final ablation study, we evaluate the effect of increasing the number of trainable prompting vectors. For this experiment, we also fix GCN as the base GNN and report the results in Table \ref{tab:tokeneffect}. This empirical evaluation clarifies that increasing the number of does not have a significant impact on \namemodel's performance. This conclusion is indeed favorable, meaning that our method can achieve desirable results even with a considerably low number of trainable parameters. 

\begin{table*}[!t]
    \caption{The effect of the number of trainable prompting vectors ($n_t$) while using GCN as the base model. A higher number of trainable vectors yields only a marginal improvement, and the model performs favorably with fewer trainable vectors.}
    \vspace{-2mm}
    \centering
    \scalebox{0.8}{
        \begin{tabular}{L{0.6in}*{6}{C{0.6in}}}
            \toprule
            \textbf{Dataset} & $n_t=10$ & $n_t=20$ & $n_t=30$ & $n_t=40$ & $n_t=50$ & $n_t=60$ \\
            \midrule
            \textbf{Cora} & 57.3 & 57.2 & 57.3 & 57.4 & 57.4 & 57.5 \\
            \textbf{PROTEINS} & 55.5 & 56.0 & 56.1 & 55.8 & 56.7 & 56.2 \\
            \bottomrule
        \end{tabular}}
    \label{tab:tokeneffect}
\end{table*}

\subsubsection{Analysis of Computational Cost}\label{apdx:computation_time}

A practical consideration for any adaptation method is its computational cost. The unsupervised nature of \namemodel, which relies on data augmentation and multiple regularization components, introduces manageable overhead during training. This is an expected trade-off for the significant advantage of operating without labeled data. Table \ref{tab:time_comparison} shows the seconds of the average full dataset training epoch time and test time across node classification datasets. Our training times are marginally higher than supervised prompting baselines, but scale reasonably on large graphs like PubMed and Flickr.

However, the more critical metric for real-world deployment is inference efficiency. Once the prompt is trained, the adaptation process is complete. At test time, the expensive training components, such as data augmentation and the discriminator, are no longer required. The inference step simply involves a forward pass through the frozen GNN with the learned prompt, resulting in a computational complexity of $O(NLd^2 + L|E|d + Ndn_t)$---as discussed in Section \ref{sec:objective_complexity_analysis}---compared to $O(NLd^2 + L|E|d)$ of a regular \gnn models and $n_t$ is number of trainable prompting vectors. Since, in all our experiments, 
 even for large graphs $n_t\leq 50$, the overhead can be neglected. Our empirical results confirm this efficiency, showing that \namemodel's test time is consistently below the average of competing prompting methods, making it a lightweight and practical solution for deployment.

 \begin{table*}[!t]
    \small
    \caption{Comparison of \namemodel with other prompting methods based on the average full dataset training and test (inference) time measures in seconds. \namemodel has marginally higher training time and a test time below the average of all methods.}
    \centering
    \resizebox{\textwidth}{!}{
        \begin{tabular}{L{1.0in}*{4}{C{0.8in}C{0.8in}}}
            \toprule
            \multirow{2}{*}{\textbf{Method}} & \multicolumn{2}{c}{\textbf{Cora}} & \multicolumn{2}{c}{\textbf{CiteSeer}} & \multicolumn{2}{c}{\textbf{PubMed}} & \multicolumn{2}{c}{\textbf{Flickr}} \\
            \cmidrule(lr){2-3} \cmidrule(lr){4-5} \cmidrule(lr){6-7} \cmidrule(lr){8-9}
            & Test time (s), 813 nodes & Train time (s), 406 nodes & Test time (s), 984 nodes & Train time (s), 492 nodes & Test time (s), 5917 nodes & Train time (s), 2957 nodes & Test time (s), 22313 nodes & Train time (s), 17850 nodes \\
            \midrule
            Fine-Tuning & 0.075 & 0.077 & 0.096 & 0.052 & 1.081 & 0.636 & 1.833 & 1.585 \\
            All-In-One & 1.638 & 0.746 & 0.452 & 0.26 & 7.016 & 3.933 & 10.782 & 9.856 \\
            GraphPrompt+ & 0.715 & 0.802 & 0.166 & 0.197 & 4.206 & 4.970 & 2.243 & 4.094 \\
            GPF-Plus & 0.939 & 0.766 & 0.351 & 0.275 & 5.542 & 3.754 & 7.494 & 9.479 \\
            UGPrompt \textbf{(ours)} & 0.925 & 1.181 & 0.315 & 0.458 & 5.56 & 4.946 & 6.730 & 13.904 \\
            \midrule
            Average across Prompting Methods & 1.054 & 0.874 & 0.321 & 0.298 & 5.581 & 4.401 & 6.812 & 9.333 \\
            \bottomrule
        \end{tabular}
    }
    \label{tab:time_comparison}
\end{table*}

\subsubsection{Effect of changing the prompting function} \label{apdx:prompting_function}
An advantage of \namemodel, as discussed in Section \ref{sec:consistency_prompting}, is its versatility, as it serves as a general framework agnostic to the base \gnn and the prompting function. This means our unsupervised framework can potentially enhance a prefix prompting function such as \citep{gpf2023, allinone2023}. To support this claim, we design an experiment where we substitute our experimental prompting function (which is similar to GPF-Plus) with All-In-One's prompting function and present the results in Table \ref{tab:allinone}. These results show that All-In-One’s prompting performs better when integrated into our framework. Notably, this improvement occurs without the use of labeled data.

\paragraph{Remark on prompt tuning and other PEFT methods.}
Other parameter-efficient fine-tuning (PEFT) methods for adaptation, such as LoRA, may also be considered for source-trained \gnn adaptation. However, input-level prompt tuning is more aligned with the \problemname setting for two reasons. First, it treats the source-trained \gnn as a frozen, potentially inaccessible model, since the prompt only modifies the input graph representation and does not require accessing or changing its internal layers. In contrast, methods such as LoRA inject trainable parameters into specific layers and therefore require architectural access to the model. Second, our consistency-based objective compares a weakly augmented graph with a strongly augmented and prompted graph, both of which are processed by the same frozen \gnn. This keeps adaptation isolated in the prompting function. Moreover, standard \gnns are usually much smaller than LLMs, so layer-wise adapters may add a non-negligible number of parameters compared to our small set of trainable prompting vectors, which is important when adaptation relies only on unlabeled target data. We leave a systematic study of other PEFT methods under \problemname as future work.

\begin{table*}[!t]
    \caption{Evaluation of \namemodel with All-In-One's prompting function using GCN as the base model. The results show that All-In-One's prompting function performs better when wrapped in our unsupervised framework, demonstrating that \namemodel is a versatile prompting framework that can enhance prefix prompting methods.}
    \vspace{-2mm}
    %\footnotesize
    \centering
    \scalebox{0.8}{
        \begin{tabular}{L{1.1in}*{7}{C{0.4in}}}
            \toprule
            \multirow{2}{6em}{\textbf{Method}} & \multirow{2}{*}{\hspace{-2mm}\textbf{\%Label}} &  \multicolumn{2}{c}{\textbf{ENZYMES}}  &  \multicolumn{2}{c}{\textbf{PROTEINS}}  &  \multicolumn{2}{c}{\textbf{DHFR}} \\
            \cmidrule(lr){3-4} \cmidrule(lr){5-6} \cmidrule(lr){7-8}
            & & \textbf{F1} & \textbf{IMP} & \textbf{F1} & \textbf{IMP} & \textbf{F1} & \textbf{IMP} \\
            \midrule
            BaseModel & 0 & 47.7\std{5.7} & - & \textbf{51.8}\std{7.0} & - & 75.5\std{6.3} & - \\
            \midrule
            \multirow{2}{6em}{All-In-One} & 50 & \underline{48.7}\std{1.0} & \underline{2.1} & 45.8\std{10.4} & \underline{-11.6} & \underline{79.2}\std{0.6} & \underline{4.8} \\
            & 25 & 45.8\std{1.9} & -4.0 & 38.1\std{13.4} & -26.4 & 79.1\std{0.6} & 4.6 \\
            \midrule
            \namemodel (ours) & \textbf{0} & \textbf{48.9}\std{0.9} & \textbf{2.5} & \underline{50.8}\std{2.5} & \textbf{-1.9} & \textbf{79.3}\std{2.4} & \textbf{5.0} \\
            \bottomrule
        \end{tabular}}
    \label{tab:allinone}
    % 	\vskip -0.1in
\end{table*}

\subsubsection{Effect of types of augmentation}\label{apdx:augmentation_type}

Augmentation is a key component of our framework. As discussed in Section \ref{sec:consistency_prompting}, the type of augmentation should align with the prompting function. For instance, if the prompting function applies feature modifications, as in our main experimental prompting function, feature augmentation is expected to be more beneficial than structural augmentation (e.g., adding or removing edges), and vice versa.

As an ablation study on the type of augmentation, Table \ref{tab:augmentation_type} shows how different types of augmentation impact \namemodel's performance. Here, our prompting function transforms the feature matrix, the feature augmentation masks features, and structural augmentation drops edges. Results meet our expectations that feature augmentation improves performance, as it better aligns with the prompting function. Since none of the existing \gnn prompting functions can be categorized solely as structural prompting (without changing the node representations), we leave experimenting with such a prompting method for future work. 

\begin{table*}[!t]
    \small
    \caption{Augmentation function's effect on \namemodel performance using GPF-Plus's feature prompting. Feature augmentation aligns more with feature prompting and generally achieves better results.}
    \centering
       \resizebox{\textwidth}{!}{ \begin{tabular}{L{0.6in}C{0.7 in}*{12}{C{0.3in}}}
            \toprule
            \multirow{2}{*}{\textbf{Method}} & \multirow{2}{10em}{\textbf{Aug. Type}} & \multicolumn{2}{c}{\textbf{ENZYMES}} &   \multicolumn{2}{c}{\textbf{PROTEINS}} & \multicolumn{2}{c}{\textbf{DHFR}} & \multicolumn{2}{c}{\textbf{Cora}} & \multicolumn{2}{c}{\textbf{CiteSeer}} & \multicolumn{2}{c}{\textbf{PubMed}} \\
            \cmidrule(lr){3-4} \cmidrule(lr){5-6} \cmidrule(lr){7-8} \cmidrule(lr){9-10} \cmidrule(lr){11-12} \cmidrule(lr){13-14}
            & & \textbf{F1} & \textbf{IMP} & \textbf{F1} & \textbf{IMP} & \textbf{F1} & \textbf{IMP} & \textbf{F1} & \textbf{IMP} & \textbf{F1} & \textbf{IMP} & \textbf{F1} & \textbf{IMP} \\
            \midrule
            BaseModel & & 47.7 & - & 51.8 & - & 75.5 & - & 53.8 & - & 44.1 & - & 57.1 & - \\
            \midrule
            \multirow{2}{6em}{\namemodel (ours)} & Feature & \textbf{49.1} & \textbf{2.9} & \textbf{56.0} & \textbf{8.1} & \textbf{77.0} & \textbf{2.0} & \textbf{57.3} & \textbf{6.5} & \textbf{45.6} & \textbf{2.9} & \textbf{61.0} & \textbf{6.8} \\
            & Structural & 48.2 & 1.0 & 54.5 & 5.2 & 75.2 & -0.4 & 57.0 & 5.9 & 44.8 & 1.6 & 55.7 & -2.5 \\
            \bottomrule
        \end{tabular}}
    \label{tab:augmentation_type}
\end{table*}

\subsection{Source Alignment of Confidence-filtered Target Samples}
\label{app:mmd_confident_samples}

Although the source data are not used during \namemodel training, we perform a diagnostic analysis to understand the role of confidence filtering. Therefore, source compatibility is induced indirectly through the frozen source-trained \gnn and the confidence threshold used in $L_c$.

To verify this effect, we measure the Maximum Mean Discrepancy (MMD) between the latent embeddings of the source data and two subsets of weakly augmented target samples: confident samples with $\max(\tilde{\mathbf{p}}^{\graph}_{\varphi}) \ge \tau$, and unconfident samples with $\max(\tilde{\mathbf{p}}^{\graph}_{\varphi}) < \tau$. Table~\ref{tab:mmd_confident_samples} shows that confident target samples are generally closer to the source distribution than unconfident ones. This supports our interpretation that confidence filtering selects the portion of the target distribution that is more compatible with the source-trained model.

Using the adversarial regularizer in $L_{\mathrm{adv}}$ helps align the prompted graphs with their non-prompted weak augmentations on the target domain. According to the results, although this objective does not directly align the prompted representations with the unobserved source distribution, it prevents them from drifting too far from the target-side anchors, which are closer to the source samples.

\begin{table*}[!t]
    \centering
    \caption{Source alignment of confidence-filtered target samples. We report MMD between source embeddings and confident/unconfident weakly augmented target samples. A lower MMD indicates closer alignment with the source distribution.}
    \label{tab:mmd_confident_samples}
    \scalebox{0.85}{
        \begin{tabular}{l*{9}{c}}
            \toprule
            \textbf{Dataset} & \textbf{CiteSeer} & \textbf{Cornell} & \textbf{DHFR} & \textbf{Cora} & \textbf{Wisconsin} & \textbf{ENZYMES} & \textbf{PubMed} & \textbf{PROTEINS} & \textbf{Texas} \\
            \midrule
            Conf. Set & 0.026 & 0.118 & 0.117 & 0.055 & 0.126 & 0.072 & 0.024 & 0.394 & 0.099 \\
            UNConf. Set & 0.119 & 0.257 & 0.381 & 0.172 & 0.299 & 0.075 & 0.726 & 0.262 & 0.214 \\
            \bottomrule
        \end{tabular}}
    \label{tab:mmd_alignment}
\end{table*}

\subsection{Pseudo-label Reliability and Sensitivity to Confidence Threshold}
\label{app:pseudolabel_reliability}

Pseudo-labeling under covariate shift may suffer from confirmation bias when high confidence predictions are incorrect. To evaluate this risk, we report pseudo-label accuracy, calibration before and after prompting, and sensitivity to the confidence threshold $\tau$ in Table~\ref{tab:pseudolabel_reliability}. For binary classification datasets, we have $\tau \in \{0.6, 0.7, 0.8, 0.9\}$ and for multi-class classification we use $\tau \in \{0.2, 0.4, 0.6, 0.8\}$. First, increasing $\tau$ consistently improves pseudo-label accuracy, indicating that the threshold successfully filters out noisy predictions caused by distribution shift. Second, comparing the expected calibration error (ECE) before and after prompting reveals that \namemodel generally improves model calibration rather than amplifying overconfidence. Together, our regularized prompting framework extracts reliable signals and mitigates confirmation bias. We will include this analysis in the revised appendix.

\begin{table*}[!t]
    \centering
    \caption{
    Sensitivity of pseudo-label accuracy and calibration before and after prompting to the confidence threshold $\tau$. \namemodel shows that increasing the threshold generally improves pseudo-label precision.
    }
    % \vspace{2mm}
    \resizebox{0.8\textwidth}{!}{
        \begin{tabular}{L{0.6in}C{0.7in}C{1.1in}C{1.1in}C{1.1in}}
            \toprule
            \textbf{Dataset} & \textbf{Threshold ($\tau$)} & \textbf{BaseModel Pseudo ACC} $\uparrow$ & \textbf{\namemodel ECE} $\downarrow$ & \textbf{BaseModel ECE} $\downarrow$ \\
            \midrule
            \multirow{4}{*}{Cornell} 
            & 0.2 & 0.289 & 0.141 & \multirow{4}{*}{0.138} \\
            & 0.4 & 0.302 & 0.144 & \\
            & 0.6 & 0.314 & \textbf{0.135} & \\
            & 0.8 & \textbf{0.357} & 0.137 & \\
            \midrule
            \multirow{4}{*}{Wisconsin} 
            & 0.2 & 0.413 & 0.154 & \multirow{4}{*}{0.145} \\
            & 0.4 & 0.417 & 0.151 & \\
            & 0.6 & 0.422 & \textbf{0.145} & \\
            & 0.8 & \textbf{0.439} & \textbf{0.145} & \\
            \midrule
            \multirow{4}{*}{Texas} 
            & 0.2 & 0.379 & 0.203 & \multirow{4}{*}{0.213} \\
            & 0.4 & 0.411 & 0.192 & \\
            & 0.6 & \textbf{0.412} & 0.162 & \\
            & 0.8 & 0.359 & \textbf{0.161} & \\
            \midrule
            \multirow{4}{*}{ENZYMES} 
            & 0.2 & 0.538 & 0.189 & \multirow{4}{*}{0.185} \\
            & 0.4 & 0.560 & 0.188 & \\
            & 0.6 & 0.579 & 0.189 & \\
            & 0.8 & \textbf{0.613} & \textbf{0.185} & \\
            \midrule
            \multirow{4}{*}{DHFR} 
            & 0.6 & 0.735 & \textbf{0.098} & \multirow{4}{*}{0.105} \\
            & 0.7 & 0.735 & \textbf{0.098} & \\
            & 0.8 & 0.764 & 0.101 & \\
            & 0.9 & \textbf{0.825} & 0.101 & \\
            \midrule
            \multirow{4}{*}{PROTEINS} 
            & 0.6 & 0.698 & \textbf{0.082} & \multirow{4}{*}{0.092} \\
            & 0.7 & 0.698 & \textbf{0.082} & \\
            & 0.8 & 0.724 & 0.083 & \\
            & 0.9 & \textbf{0.750} & 0.083 & \\
            % \midrule
            % \multirow{4}{*}{Cora} 
            % & 0.6 & 0.591 & 0.294 & \multirow{4}{*}{0.282} \\
            % & 0.7 & 0.600 & 0.294 & \\
            % & 0.8 & 0.608 & 0.294 & \\
            % & 0.9 & \textbf{0.620} & 0.294 & \\
            % \midrule
            % \multirow{4}{*}{CiteSeer} 
            % & 0.6 & 0.498 & 0.154 & \multirow{4}{*}{0.153} \\
            % & 0.7 & 0.508 & 0.154 & \\
            % & 0.8 & 0.518 & 0.154 & \\
            % & 0.9 & \textbf{0.533} & \textbf{0.153} & \\
            % \midrule
            % \multirow{4}{*}{PubMed} 
            % & 0.6 & 0.598 & 0.154 & \multirow{4}{*}{0.035} \\
            % & 0.7 & 0.604 & 0.154 & \\
            % & 0.8 & 0.608 & 0.153 & \\
            % & 0.9 & \textbf{0.614} & \textbf{0.152} & \\
            \bottomrule
        \end{tabular}}
    \label{tab:pseudolabel_reliability}
\end{table*}

\subsubsection{Unsupervised Prompting vs Fully Supervised Prompting} \label{apdx:fully_supervised}
\begin{figure*}[h]
    \centering
    \begin{minipage}{0.725\columnwidth}
        \centering
        \includegraphics[width=1.0\columnwidth]{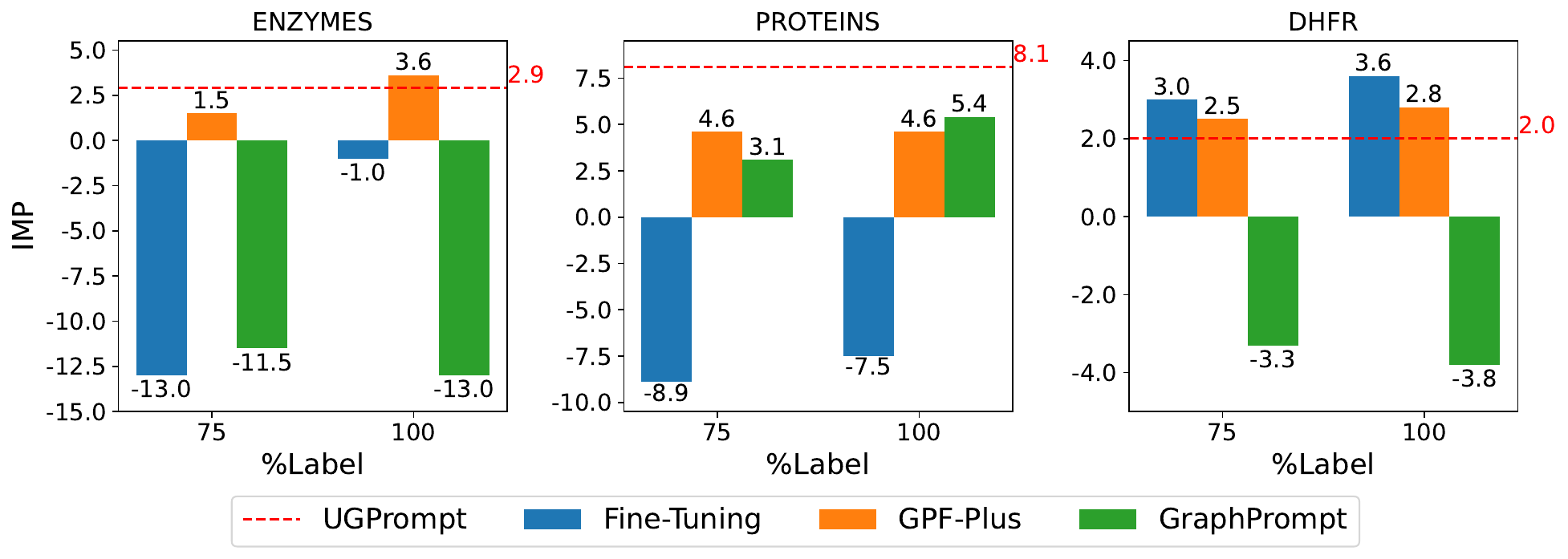}
        \subcaption{Graph classification under edge homophily distribution shift for GCN as base model.}
    \end{minipage}
    
    \vspace{0.18cm} % space between images
    
    \begin{minipage}{0.725\columnwidth}
        \centering
        \includegraphics[width=1.0\columnwidth]{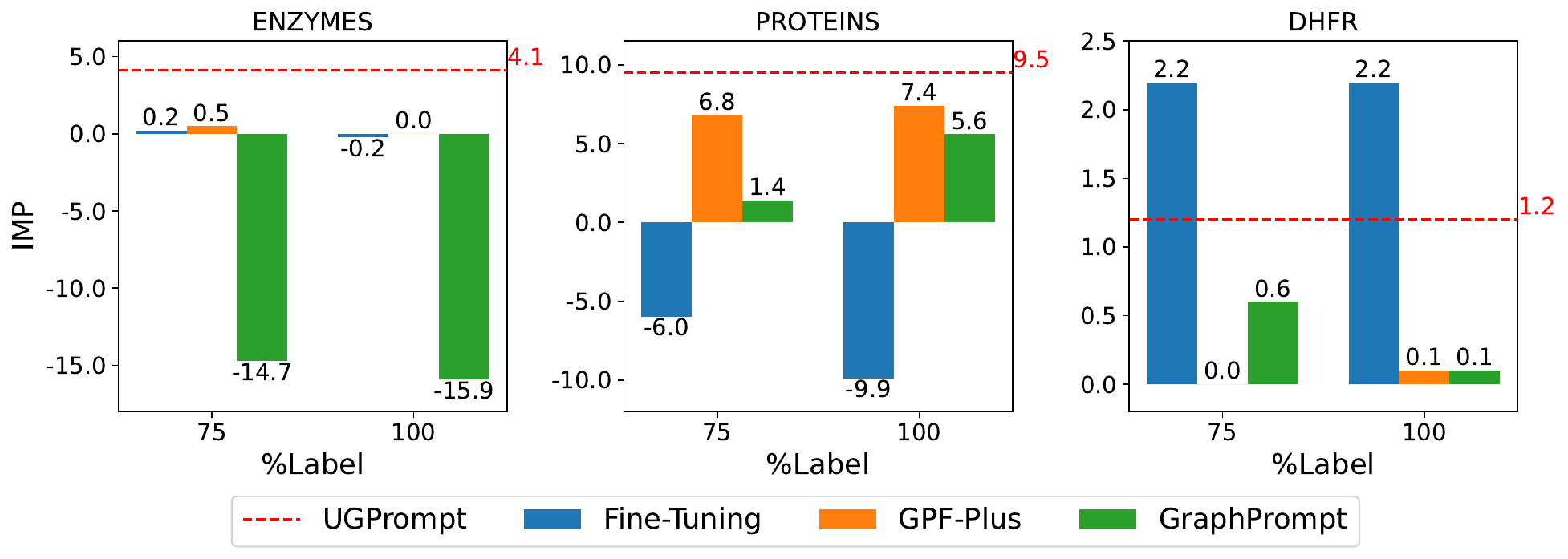}
        \subcaption{Graph classification under edge homophily distribution shift for GAT as base model.}
    \end{minipage}

    \vspace{0.18cm} % space between images

    \begin{minipage}{0.725\columnwidth}
        \centering
        \includegraphics[width=1.0\columnwidth]{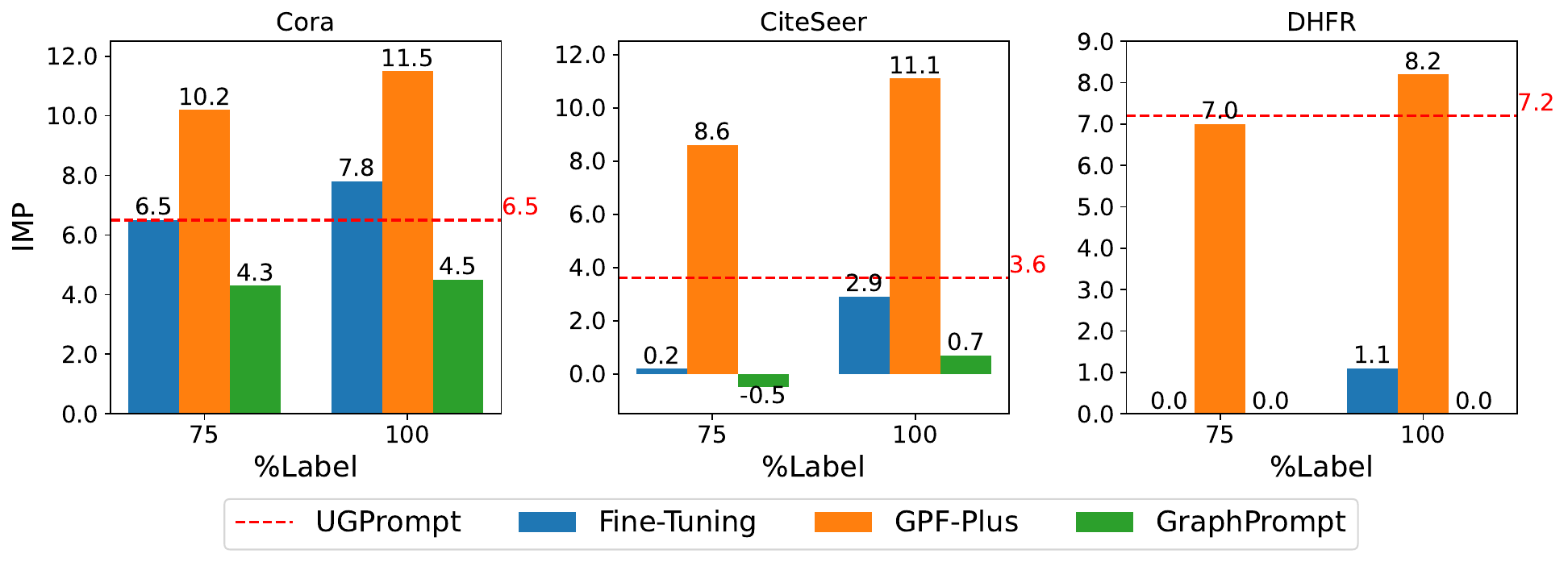}
        \subcaption{Node classification under PR distribution shift for GCN as base model.}
    \end{minipage}
    
    \vspace{0.18cm} % space between images
    
    \begin{minipage}{0.725\columnwidth}
        \centering
        \includegraphics[width=1.0\columnwidth]{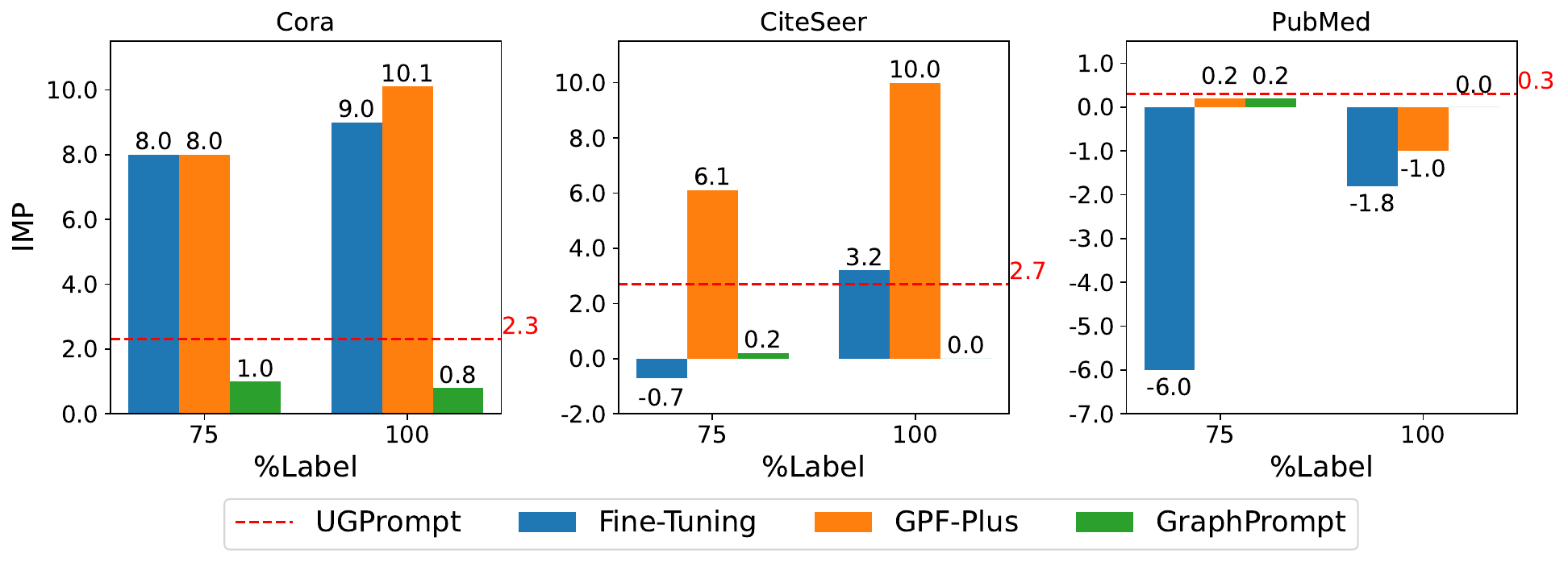}
        \subcaption{Node classification under PR distribution shift for GAT as base model.}
    \end{minipage}

    \caption{Performance gains for GCN and GAT base models on graph and node classification tasks in the presence of edge homophily (a, b) and node page rank (c, d) distribution shifts; where the competitor prompting methods utilize 100\% and 75\% labeled data of target distributions. \namemodel without using any labels always improves over the base model and achieves the best results on graph classification with GAT, while ranking second-best in other cases.}
    \label{fig:mainshifts_label}
\end{figure*}

\begin{figure*}[ht]
    \centering
    \begin{minipage}{0.725\columnwidth}
        \centering
        \includegraphics[width=1.0\columnwidth]{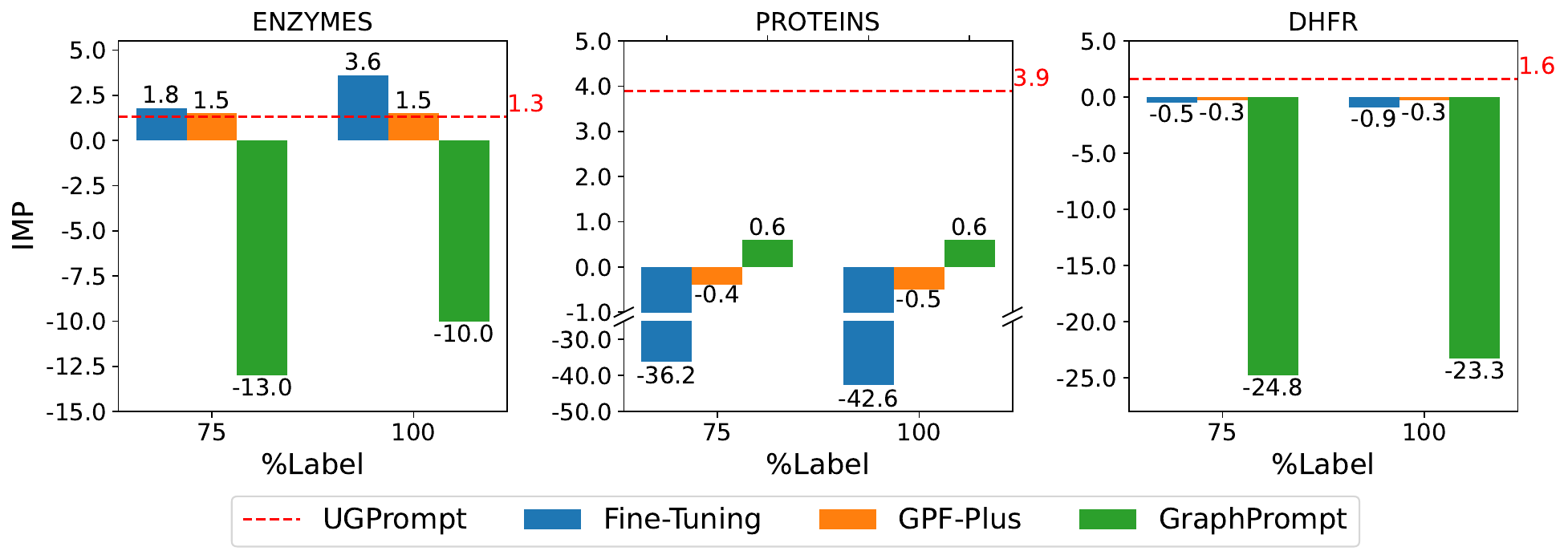}
        \subcaption{Graph classification under graph density distribution shift for GCN as base model.}
    \end{minipage}
    
    \vspace{0.5cm} % space between images
    
    \begin{minipage}{0.725\columnwidth}
        \centering
        \includegraphics[width=1.0\columnwidth]{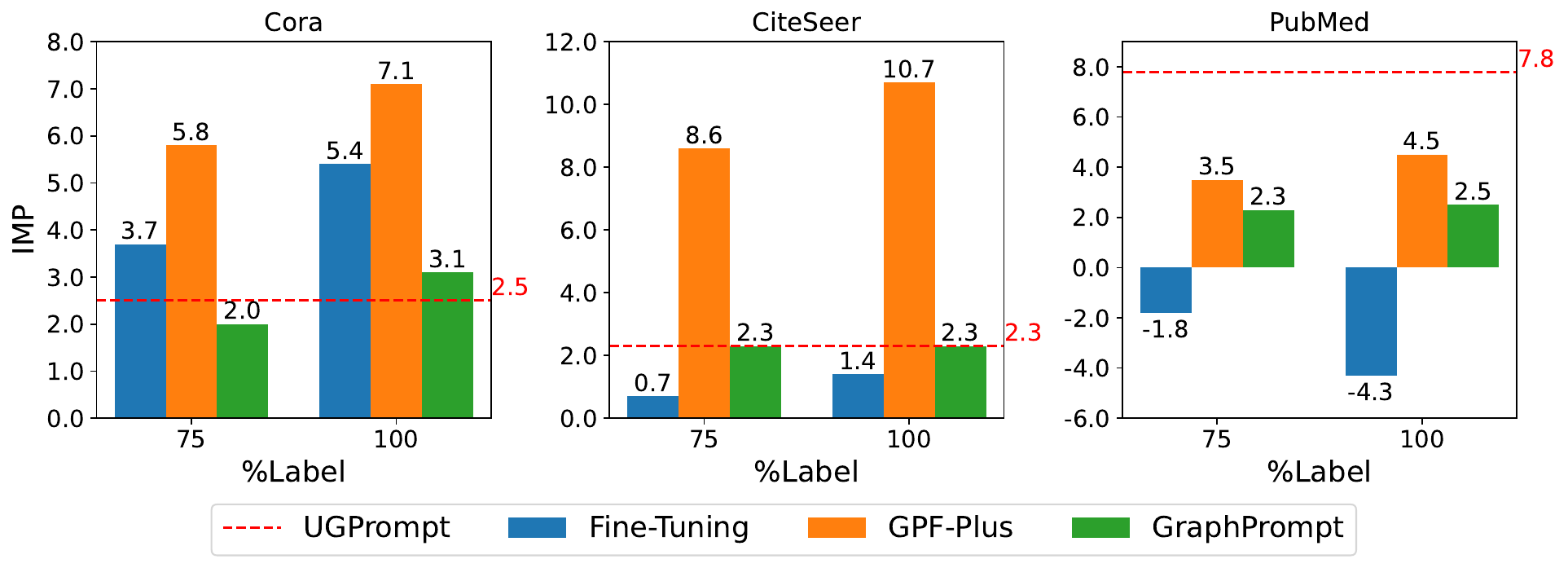}
        \subcaption{Node classification under clustering coefficient distribution shift for GCN as base model.}
    \end{minipage}

    \caption{Performance gains for GCN as the base model on graph and node classification tasks in the presence of graph density (a) and node clustering coefficient (b) distribution shifts when the competitor prompting methods utilize 100\% and 75\% labeled data of target distributions. The trend is similar to other distribution shifts (Figure \ref{fig:mainshifts_label}) where \namemodel generally attains the best results on graph classification and the second-best on node classification.}
    \label{fig:othershifts_label}
\end{figure*}

Since \namemodel achieves significant improvements across different experiments, we are interested in evaluating competitors while allowing them to access 50\%, 75\%, and 100\% of the labeled data from the target distributions. We show the results of these settings in Tables \ref{tab:graphgcn50} and \ref{tab:nodegcn50}, and Figures \ref{fig:mainshifts_label} and \ref{fig:othershifts_label}. All results are reported for both GCN and GAT under distribution shifts induced by edge homophily for graph classification and PR node classification.

Graph classification results for both GCN and GAT base models in Table \ref{tab:graphgcn50} illustrate the superior performance of \namemodel as an unsupervised method compared to the baselines in most cases, even when they have access to 50\% of the labeled samples for training on the target datasets. On the node classification datasets, \namemodel achieves the second-best performance as shown in Table \ref{tab:nodegcn50}. Specifically, for larger node classification datasets, it is noteworthy that we beat all baselines on Flickr and only underperform GraphPrompt+ on PubMed. 

\begin{table*}[t]
    \caption{Graph classification results on target datasets for GCN as the base model. Our unsupervised method mostly achieves the best results even when the competitors use 50\% of the labeled data.}
    \centering
    \scalebox{0.8}{
        \begin{tabular}{L{0.8in}L{1.1in}C{0.3in}*{6}{C{0.4in}}}
            \toprule
            \multirow{2}{6em}{\textbf{Base GNN}} & \multirow{2}{6em}{\textbf{Method}} & \multirow{2}{*}{\hspace{-2mm}\textbf{\%Label}} &  \multicolumn{2}{c}{\textbf{ENZYMES}}  &  \multicolumn{2}{c}{\textbf{PROTEINS}}  &  \multicolumn{2}{c}{\textbf{DHFR}} \\
            \cmidrule(lr){4-5} \cmidrule(lr){6-7} \cmidrule(lr){8-9}
            & & & \textbf{F1} & \textbf{IMP} & \textbf{F1} & \textbf{IMP} & \textbf{F1} & \textbf{IMP} \\
            \midrule
            \multirow{7}{*}{GCN} & BaseModel & 0 & 47.7\std{5.7} & - & 51.8\std{7.0} & - & 75.5\std{6.3} & - \\
            \cmidrule(lr){2-9}
            & Fine-Tuning & \multirow{5}{*}{50} & 45.0\std{2.0} & -5.7 & 46.9\std{1.6} & -9.5 & \underline{77.6}\std{0.2} & \underline{2.8} \\
            & GraphPrompt & & 40.2\std{1.5} & -15.7 & \underline{54.1}\std{1.0} & \underline{4.4} & 73.1\std{0.9} & -3.2 \\
            & GraphPrompt+ & & 29.7\std{1.7} & -37.7 & 49.7\std{1.0} & -4.1 & 65.5\std{1.3} & -13.2 \\
            & All-In-One & & \underline{48.7}\std{1.0} & \underline{2.1} & 45.8\std{10.4} & -11.6 & \textbf{79.2}\std{0.6} & \textbf{4.8} \\
            & GPF-Plus & & 48.6\std{0.9} & 1.9 & 53.8\std{2.4} & 3.9 & 77.4\std{0.3} & 2.5 \\
            \cmidrule(lr){2-9}
            & \namemodel (ours) & \textbf{0} & \textbf{49.1}\std{0.6} & \textbf{2.9} & \textbf{56.0}\std{1.5} & \textbf{8.1} & 77.0\std{2.4} & 2.0 \\
            \midrule
            \multirow{7}{*}{GAT} & BaseModel & 0 & \underline{44.1}\std{6.4} & - & 51.5\std{8.1} & - & 77.3\std{3.5} & - \\
            \cmidrule(lr){2-9}
            & Fine-Tuning & \multirow{5}{*}{50} & 43.3\std{1.2} & -1.8 & 49.2\std{0.5} & -4.5 & \underline{78.0}\std{0.2} & \underline{0.9} \\
            & GraphPrompt & & 35.3\std{1.9} & -20.0 & 50.3\std{0.9} & -2.3 & 77.1\std{0.9} & -0.3 \\
            & GraphPrompt+ & & 29.2\std{6.2} & -33.8 & 52.1\std{6.3} & 1.2 & 60.5\std{12.7} & -21.7 \\
            & All-In-One & & 40.3\std{2.5} & -8.6 & 41.8\std{11.2} & -18.8 & 77.3\std{1.2} & 0.0 \\
            & GPF-Plus & & 43.4\std{2.0} & -1.6 & \underline{54.8}\std{2.7} & \underline{6.4} & 77.2\std{0.8} & -0.6 \\
            \cmidrule(lr){2-9}
            & \namemodel (ours) & \textbf{0} & \textbf{45.9}\std{2.2} & \textbf{4.1} & \textbf{56.4}\std{2.0} & \textbf{9.5} & \textbf{78.2}\std{0.9} & \textbf{1.2} \\
            \bottomrule
        \end{tabular}}
    \label{tab:graphgcn50}
\end{table*}

\begin{table*}[t]
    \caption{Node classification results on target datasets for GCN as the base model. Comparing all baselines with access to 50\% of labeled data, \namemodel achieves the second-best results without labels.}
    \centering
    \resizebox{\textwidth}{!}{
        \begin{tabular}{L{0.8in}L{1.1in}C{0.3in}*{8}{C{0.4in}}}
            \toprule
            \multirow{2}{6em}{\textbf{Base GNN}} & \multirow{2}{6em}{\textbf{Method}} & \multirow{2}{*}{\hspace{-2mm}\textbf{\%Label}} &  \multicolumn{2}{c}{\textbf{Cora}} & \multicolumn{2}{c}{\textbf{CiteSeer}} & \multicolumn{2}{c}{\textbf{PubMed}} & \multicolumn{2}{c}{\textbf{Flickr}} \\
            \cmidrule(lr){4-5} \cmidrule(lr){6-7} \cmidrule(lr){8-9} \cmidrule(lr){10-11}
            & & & \textbf{F1} & \textbf{IMP} & \textbf{F1} & \textbf{IMP} & \textbf{F1} & \textbf{IMP} & \textbf{F1} & \textbf{IMP} \\
            \midrule
            \multirow{8}{*}{GCN} & BaseModel & 0 & 53.8\std{2.4} & - & 44.1\std{1.5} & - & 57.1\std{0.8} & - & \underline{16.5}\std{0.4} & - \\
            \cmidrule(lr){2-11}
            & Fine-Tuning & \multirow{6}{*}{50} & 54.5\std{1.2} & 1.3 & 43.5\std{0.4} & -1.4 & 56.3\std{2.1} & -1.4 & 10.3\std{0.4} & -37.6 \\
            & GPPT & & 50.5\std{3.1} & -6.1 & 40.6\std{1.2} & -7.9 & 51.8\std{3.7} & -9.3 & 13.6\std{0.5} & -17.6 \\
            & GraphPrompt & & 55.8\std{0.3} & 3.7 & 43.8\std{0.3} & -0.7 & 57.1\std{0.1} & 0.0 & 13.1\std{0.2} & -20.6\\
            & GraphPrompt+ & & \underline{57.3}\std{0.6} & \underline{6.5} & 42.9\std{0.2} & -2.7 & \textbf{64.9}\std{0.2} & \textbf{13.7} & 14.9\std{0.7} & -9.7 \\
            & All-In-One & & 50.3\std{1.2} & -6.5 & 39.3\std{1.0} & -10.9 & 39.8\std{1.2} & -30.3 & 13.5\std{0.3} & -18.2 \\
            & GPF-Plus & & \textbf{58.2}\std{0.6} & \textbf{8.2} & \textbf{46.8}\std{0.7} & \textbf{6.1} & 60.3\std{0.5} & 5.6 & 13.1\std{0.1} & -20.6 \\
            \cmidrule(lr){2-11}
            & \namemodel (ours) & \textbf{0} & \underline{57.3}\std{0.4} & \underline{6.5} & \underline{45.7}\std{0.4} & \underline{3.6} & \underline{61.2}\std{0.3} & \underline{7.2} & \textbf{17.5}\std{0.1} & \textbf{6.1} \\
            \midrule
            \multirow{8}{*}{GCN} & BaseModel & 0 & 47.7\std{1.3} & - & 41.2\std{2.4} & - & 60.0\std{1.1} & - & \underline{17.0}\std{0.2} & - \\
            \cmidrule(lr){2-11}
            & Fine-Tuning & \multirow{6}{*}{50} & 47.1\std{1.7} & -1.3 & 39.9\std{0.5} & -3.2 & 56.5\std{2.2} & -5.8 & 10.8\std{0.3} & -36.5 \\
            & GPPT & & 32.8\std{3.8} & -31.2 & 35.6\std{1.2} & -13.6 & 51.8\std{5.5} & -13.7 & 13.1\std{0.3} & -22.9 \\
            & GraphPrompt & & 48.2\std{0.5} & 1.0 & 40.7\std{0.3} & -1.2 & 60.1\std{0.1} & 0.2 & 13.5\std{0.1} & -20.6 \\
            & GraphPrompt+ & & 48.2\std{0.5} & 1.0 & 42.0\std{0.3} & 1.9 & \textbf{67.1}\std{0.1} & \textbf{11.8} & \textbf{17.5}\std{0.6} & \textbf{2.9} \\
            & All-In-One & & 34.6\std{4.1} & -27.5 & 33.0\std{1.2} & -19.9 & 25.4\std{5.2} & -57.7 & 12.5\std{0.5} & -26.5 \\
            & GPF-Plus & & \textbf{49.6}\std{1.4} & \textbf{4.0} & \textbf{43.1}\std{0.6} & \textbf{4.6} & 60.1\std{0.5} & 0.2 & 13.2\std{0.2} & -22.4\\
            \cmidrule(lr){2-11}
            & \namemodel (ours) & \textbf{0} & \underline{48.8}\std{0.9} & \underline{2.3} & \underline{42.3}\std{0.5} & \underline{2.7} & \underline{60.2}\std{0.1} & \underline{0.3} & \textbf{17.5}\std{0.1} & \textbf{2.9} \\
            \bottomrule
        \end{tabular}}
    \label{tab:nodegcn50}
\end{table*}

Observing Figure \ref{fig:mainshifts_label}, when we use GAT as the base model, \namemodel outperforms GPF-Plus and GraphPrompt on graph classification datasets even when they utilize fully labeled target datasets. \textit{This clearly shows the effectiveness of our proposed method, \namemodel.} Additionally, we achieve closely competitive results with GCN as the base model in 100\% label setting overall. Besides looking at the node classification results for both GNNs, we generally obtain the second-best improvement in 75\% label setting and perform favorably with 100\% compared to the best baseline. 
 
Next, we evaluate our method in distribution shifts of graph density and clustering coefficient under the 75\% and 100\% label settings. Here, we fix GCN as the base GNN. Results are in Figure \ref{fig:othershifts_label}. Similar to the previous distributions, for graph density and clustering coefficient, \namemodel achieves competitive or better performance on graph classification datasets and ranks second after GPF-Plus on node classification, even though our method does not see any labels.  
 
Finally, an interesting finding of these experiments is that competitors can occasionally cause a performance drop compared to the base models, which is unexpected and undesirable. On the other hand, \namemodel, although it is an unsupervised method, does not negatively affect any dataset, any \gnn architecture, or any type of distribution shift. Also, all the above results lead to the same conclusion: \namemodel can enhance base GNNs that encounter different distribution shifts across various downstream tasks. Since \namemodel achieves promising results in a fully unsupervised manner, it offers new avenues for leveraging large unlabeled datasets to improve the generalization of \gnns.

\end{document}